\documentclass[10pt,twocolumn,letterpaper]{article}
\usepackage[pagenumbers]{cvpr}
\newenvironment{nolinenumbers}{}{}
\usepackage{graphicx}
\usepackage{amsmath,amssymb}
\usepackage{amsthm}
\usepackage{fix-cm}
\usepackage{tikz}

\usepackage{booktabs,multirow,colortbl,tabularx}
\usepackage[above,below]{placeins}
\definecolor{TableHeader}{HTML}{E9EFF5}
\definecolor{TableGroup}{HTML}{F0F3F6}
\definecolor{TableAlt}{HTML}{F7F9FB}
\definecolor{TableOurs}{HTML}{E6F1FA}
\definecolor{TableOursHead}{HTML}{D3E6F6}
\usepackage{pifont}

\usepackage{microtype}
\usepackage{enumitem}
\usepackage{algorithm}
\usepackage{algorithmic}
\renewcommand{\algorithmicrequire}{\textbf{Input:}}
\renewcommand{\algorithmicensure}{\textbf{Output:}}
\definecolor{cvprblue}{rgb}{0.21,0.49,0.74}
\usepackage[breaklinks,colorlinks,allcolors=cvprblue]{hyperref}
\def\confName{CVPR}
\def\confYear{2027}

\newcommand{\argmax}{\operatorname*{arg\,max}}

\title{DriveTTO: Understanding Search and Selection in Test-Time Trajectory Optimization for End-to-End Driving}
\hypersetup{%
  pdftitle={DriveTTO: Understanding Search and Selection in Test-Time Trajectory Optimization for End-to-End Driving},
  pdfauthor={Xinyao Rao, Zecheng Tang, Zheyuan Lai, Liuxuan Sun, Minjun Yu, Xun Cao, Wei Li},
  pdfsubject={Training-free trajectory optimization for autonomous driving},
  bookmarksnumbered=true
}

\usepackage{stfloats}
\usepackage{flushend}
\selectfont
\makeatletter
\def\cvprsection{\@startsection{section}{1}{\z@}{-10pt}{7pt}{\large\bf}}
\def\cvprsubsection{\@startsection{subsection}{2}{\z@}{-8pt}{5pt}{\elvbf}}
\def\cvprsubsubsection{\@startsection{subsubsection}{3}{\z@}{-6pt}{4pt}{\tenbf}}
\renewcommand\paragraph{\@startsection{paragraph}{4}{\z@}{0pt}{-1em}{\normalfont\normalsize\bfseries}}
\makeatother

\definecolor{ArxivPurple}{HTML}{783875}
\definecolor{ArxivRule}{HTML}{B897B6}
\colorlet{ArxivCard}{ArxivPurple!9!white}
\newcommand{\arxivReportDate}{November 21, 2026}

\newcommand{\arxivHeadRule}{{\color{ArxivRule}\hrule height .5pt}}
\newcommand{\arxivHeaderText}[1]{%
  {\normalfont\sffamily\bfseries\small\color{ArxivPurple}#1}%
}
\newcommand{\arxivCoverHead}{%
  \vbox{%
    \hbox to \textwidth{%
      \includegraphics[height=26pt]{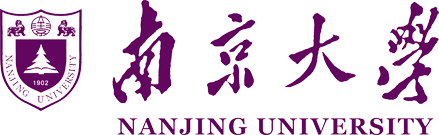}%
      \hspace{16pt}%
      \includegraphics[height=26pt]{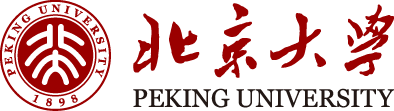}%
      \hspace{16pt}%
      \raisebox{6.75pt}{\includegraphics[height=12.5pt]{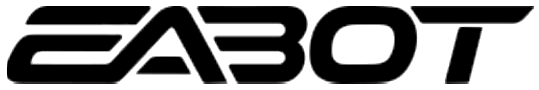}}%
      \hfil
      \raisebox{2pt}{\arxivHeaderText{\arxivReportDate}}%
    }%
    \kern6pt
    \arxivHeadRule
  }%
}
\newcommand{\arxivRunningHead}{%
  \vbox{%
    \hbox to \textwidth{%
      \arxivHeaderText{DriveTTO}\hfil
      \arxivHeaderText{\thepage}%
    }%
    \kern8pt
    \arxivHeadRule
  }%
}

\makeatletter
\newcommand{\ps@arxiv}{%
  \ps@plain
  \let\@oddhead\arxivRunningHead
  \let\@evenhead\arxivRunningHead
  \let\@oddfoot\@empty
  \let\@evenfoot\@empty
}
\newcommand{\ps@arxivcover}{%
  \ps@arxiv
  \let\@oddhead\arxivCoverHead
  \let\@evenhead\arxivCoverHead
  \def\@oddfoot{\hfil{\normalfont\color{ArxivPurple}\thepage}\hfil}%
  \let\@evenfoot\@oddfoot
}
\makeatother
\begin{document}
\pagestyle{arxiv}
\raggedbottom
\onecolumn
\thispagestyle{arxivcover}
\begingroup
\setlength{\parindent}{0pt}
\setlength{\parskip}{0pt}
\definecolor{ArxivLink}{HTML}{365C70}
\hypersetup{urlcolor=ArxivLink}
\newsavebox{\arxivProjectIcon}
\sbox{\arxivProjectIcon}{%
  \tikz[x=1pt,y=1pt,baseline=0pt,draw=ArxivLink,line width=.55pt]{%
    \useasboundingbox (0,0) rectangle (10,9);
    \draw (5,4.5) circle[radius=4pt];
    \draw (5,4.5) ellipse[x radius=1.8pt,y radius=4pt];
    \draw (1,4.5) -- (9,4.5);
    \draw (1.5,2.7) -- (8.5,2.7);
    \draw (1.5,6.3) -- (8.5,6.3);
  }%
}
\newcommand{\arxivIcon}[1]{%
  \raisebox{\dimexpr.5ex-.5\height\relax}{\makebox[11pt][c]{#1}}%
}
\vspace*{5pt}
\noindent
\begin{tikzpicture}
  \node[fill=ArxivCard,draw=none,rounded corners=9pt,
        inner xsep=14pt,inner ysep=12pt,outer sep=0pt] {%
    \begin{minipage}{\dimexpr\textwidth-28pt\relax}
      \begin{center}
        {\fontsize{16}{18.5}\selectfont\bfseries
          DriveTTO: Understanding Search and Selection\\
          in Test-Time Trajectory Optimization for End-to-End Driving\par}
        \vspace{8pt}
        {\fontsize{10.3}{12.5}\selectfont
          Xinyao Rao$^{1}$\hspace{0.65em}%
          Zecheng Tang$^{2}$\hspace{0.65em}%
          Zheyuan Lai$^{3}$\hspace{0.65em}%
          Liuxuan Sun$^{1}$\hspace{0.65em}%
          Minjun Yu$^{3}$\hspace{0.65em}%
          Xun Cao$^{1}$\hspace{0.65em}%
          Wei Li$^{1,3,*}$\par}
        \vspace{3pt}
        {\fontsize{9.5}{11.5}\selectfont
          $^{1}$Nanjing University\qquad $^{2}$Peking University\qquad $^{3}$EABOT.AI\par}
      \end{center}
      \vspace{1pt}
      {\fontsize{10}{12}\selectfont\normalfont
        \renewenvironment{abstract}{\noindent\textbf{Abstract.}\enspace\ignorespaces}{\par}%
        \begin{abstract}
In difficult driving scenes, a learned scorer can select a poor trajectory from a fixed pool even when a better candidate is available. Nearby vehicles and road boundaries can also leave only a narrow range of high-scoring maneuvers. These constraints vary with the scene and are difficult to cover exhaustively during training. We introduce DriveTTO, a training-free method that searches around native proposals and checks each proposed refinement against a retained trajectory. A frozen scorer evaluates completed paths at the endpoints they actually reach, guiding resampling in progressively smaller neighborhoods. From a shared reference, one branch draws on planner proposals and another perturbs vehicle controls. DriveTTO blends their poses at matching times and accepts a branch or fusion update only when both frozen scorers predict a gain. On Navtest, DriveTTO recovers nonzero predictive driver model scores (PDMS) in 51.35\% of native zero-score cases on average. Accounting for new failures, the zero-score rate falls from 1.63\% to 1.13\%. Benchmark-specific configurations achieve 95.54 PDMS and 92.30 extended PDMS (EPDMS) on Navtest, 57.10 EPDMS on Navhard, and 40.1 HD-Score on HUGSIM under observed commands. With a single frozen scorer, DriveTTO-S improves all six evaluated planners on both Navtest and Navhard.
\end{abstract}%
      }
      \vspace{8pt}
      {\fontsize{9.3}{11.5}\selectfont\urlstyle{same}
        \begin{tabular}{@{}c@{\hspace{0.55em}}l@{\hspace{0.7em}}l@{}}
          \arxivIcon{\includegraphics[width=10pt]{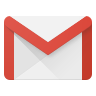}}
            & \textbf{Email:} & \href{mailto:yoloredal@gmail.com}{yoloredal@gmail.com}\\
          \arxivIcon{\includegraphics[width=10pt]{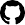}}
            & \textbf{Code:} & \url{https://github.com/Rtwotwo/DriveTTO}\\
          \arxivIcon{\usebox{\arxivProjectIcon}}
            & \textbf{Project:} & \url{https://rtwotwo.github.io/DriveTTO}
        \end{tabular}\par
      }
      \vspace{4pt}
      {\fontsize{8.7}{10.5}\selectfont $^{*}$Corresponding author.\par}
    \end{minipage}%
  };
\end{tikzpicture}\par
\vspace{10pt}
\begin{minipage}{\textwidth}
\begin{nolinenumbers}
\centering
\resizebox{\linewidth}{\height}{%
  \includegraphics[width=.94\linewidth]{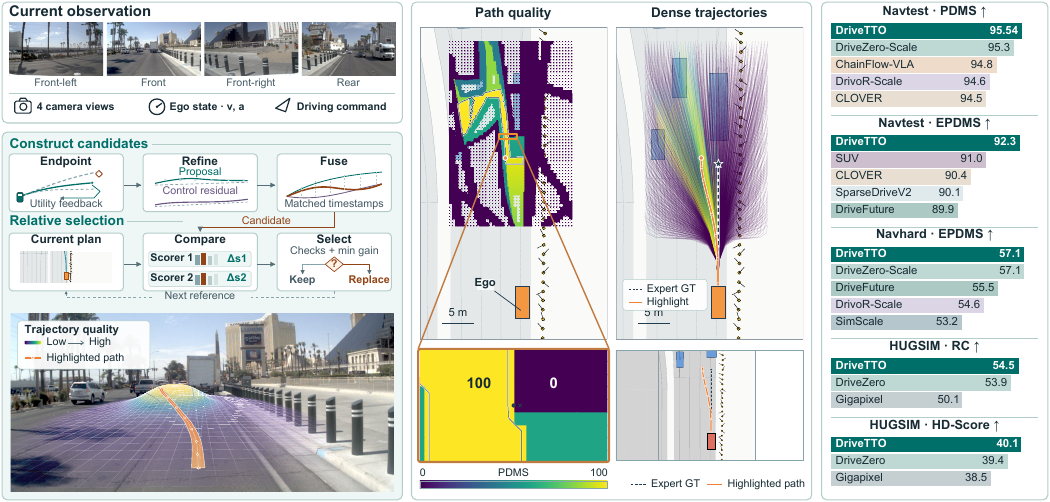}%
}%
\captionsetup{type=figure}
\caption{\textbf{DriveTTO performs training-free trajectory optimization.} \emph{Left.} Endpoint search, local refinement, and temporal fusion construct trajectory candidates. Frozen scorers evaluate updates relative to the current reference, which is retained when no replacement qualifies. \emph{Center.} In the same Navtrain scene, two endpoint requests 0.05\,m apart yield paths with offline PDMS of 100 and 0. \emph{Right.} DriveTTO matches or exceeds the baselines shown on all three benchmarks.}
\label{fig:intro}
\end{nolinenumbers}
\end{minipage}
\par\vspace{2pt}
\endgroup
\twocolumn
\flushbottom
\section{Introduction}
\label{sec:introduction}
\looseness=-1
Reliable trajectory selection is essential for autonomous driving, where planning errors can cause collisions or road departures. End-to-end planners map sensor observations to future motion~\cite{uniad2022,weng2024paradrive}. Recent approaches generate multiple candidate trajectories~\cite{diffusiondrive2024,wang2026chainflow} and use learned scorers to select an output~\cite{kirby2026registers,xu2026toad}. In difficult scenes, scoring errors can cause a single decision over the native pool to return a suboptimal or even zero-scoring plan, despite the presence of better alternatives.

\looseness=-1
Existing selection and search strategies face two interacting limitations. Reranking cannot recover a trajectory absent from the candidate pool, while broader sampling remains dependent on imperfect scores. Across six planners, our diagnostic pool expansion improves the best available offline score yet lowers the quality of single-scorer selections (Table~\ref{tab:opportunity}). Meanwhile, endpoint sweeps reveal that surrounding vehicles and road boundaries can confine high-scoring trajectories to narrow regions (Figure~\ref{fig:intro}, center). In one stopping scene, a high-scoring slice spans only 0.30\,m, and a 0.05\,m lateral request change switches PDMS from 100 to zero (Figure~\ref{fig:dense_stop_boundary}). Such precise, interaction-dependent constraints are difficult to anticipate exhaustively during training. They motivate localized exploration coupled with checks on whether a proposed update should replace the current plan.

\looseness=-1
We introduce DriveTTO, a training-free framework that couples trajectory search with relative selection (Figure~\ref{fig:intro}, left). Native proposals initialize a supported search region and supply path geometry. Endpoint search uses feedback from completed trajectories to guide resampling within contracting local neighborhoods. Bounded geometric refinement establishes a shared reference for proposal and control branches, which exploit learned motion patterns and explore local corrections beyond the bank. To constrain unreliable updates, scorers from two frozen planners compare candidates with the reference in matched scoring contexts. Time-varying fusion allows the branch outputs to contribute differently along the planning horizon. Branch and fusion replacements require both scorers to predict improvement, with the current reference retained otherwise. Experiments demonstrate native failure recovery and gains across driving benchmarks and frozen planners. Our contributions are as follows.
\begin{itemize}[leftmargin=*,itemsep=.5pt,topsep=2pt,parsep=0pt]
\interlinepenalty=0
  \item \looseness=-1 \textbf{Training-free trajectory optimization.} DriveTTO integrates proposal-supported endpoint search, bounded local refinement, and temporal fusion under frozen scoring models, enabling scene-specific correction without parameter updates.
  \item \textbf{Recovery of native planning failures.} DriveTTO recovers an average of 51.35\% of native zero-PDMS cases to nonzero scores on Navtest. Including newly introduced failures, the overall zero-score rate decreases from 1.63\% to 1.13\%.
  \item \textbf{Benchmark performance and planner transfer.} Benchmark configurations achieve 95.54 PDMS and 92.30 EPDMS on Navtest, 57.10 EPDMS on Navhard, and 40.1 HD-Score on HUGSIM under observed commands. The single-scorer variant, DriveTTO-S, improves all six evaluated native planners on both Navtest and Navhard.
\end{itemize}
\section{Related Work}
\label{sec:related}
\paragraph{End-to-end driving.}
\looseness=-1
End-to-end planners learn trajectories from sensor observations~\cite{uniad2022,weng2024paradrive}. DiffusionDrive models multiple trajectory modes~\cite{diffusiondrive2024}, while world-action and vision-language models incorporate predictive or semantic structure~\cite{world4drive2025,wang2026chainflow}. Drive-JEPA combines video representation learning with simulator-guided multimodal trajectory distillation~\cite{wang2026drivejepa}, and DrivoR learns driving-quality components for selection~\cite{kirby2026registers}. DriveTTO complements these learned capabilities by constructing additional trajectories and selecting updates relative to a retained reference, with planner and scorer weights frozen.

\paragraph{Test-time trajectory optimization.}
\looseness=-1
Parallel scaling combines learned computation streams in language models~\cite{chen2025parallel}. In planning, cross-entropy methods sample action sequences and refit distributions to elite candidates~\cite{chua2018pets,pinneri2020icem}. ToAD uses frozen scorers for control-space search, regularizes deviations from the native plan, and compares the final candidate with that plan~\cite{xu2026toad}. DriveTTO combines endpoint completion, proposal and control refinement, and temporal fusion, comparing updates with a retained trajectory throughout the search.

\paragraph{Learned scoring and selection.}
\looseness=-1
Optimizing against imperfect reward models can improve proxy scores without improving target quality. Ensembles can mitigate such overoptimization through conservative aggregation, including worst-case objectives that use the lowest predicted reward across models~\cite{coste2024ensembles}. DriveTTO screens and ranks updates using reference-relative gains computed in shared scoring contexts. Fixed-pool controls separate ranking from screening, while failure accounting measures recovered and newly introduced failures alongside mean quality.
\section{Methodology}
\label{sec:method}
\begingroup

\subsection{Framework overview}
\label{sec:contract}
A useful plan can lie outside the native pool, and a scorer can overlook it even after search finds it. The top row of Figure~\ref{fig:framework} shows how DriveTTO addresses these two problems. Endpoint search and local refinement construct alternatives. Relative selection tests whether a candidate improves on the retained plan. The planners and their scorers remain frozen.

Given a scene observation $o$, DriveTTO takes a planner's native trajectory $n$, proposal banks $\mathcal P_1(o),\mathcal P_2(o)$, and frozen scorers $S_1,S_2$. Each trajectory $\tau=\{(p_t,\psi_t)\}_{t=1}^T$ specifies position and heading on a common time grid.

\looseness=-1
In the Navtest PDMS configuration (Full), $S_1$ selects the endpoint path $f$. Under $S_2$, blending gives $b_0$ and geometric refinement produces the shared center $b_G$. The proposal and control branches start from $b_G$ and return paths $a$ and $c$. Fusion combines their poses at matching times. Both scorers compare each resulting path with the retained branch output $e$. Table~\ref{tab:configuration_map} gives the schedules used on other benchmarks.

DriveTTO-S reuses one frozen scorer and the target planner's bank in both roles. Identical candidates, references, banks, and slot order make the two gains coincide, reducing dual-positive acceptance to one scorer's preference.

\begin{figure*}[t]
\centering
\includegraphics[trim=0bp 1bp 0bp 12bp,clip,width=\textwidth]{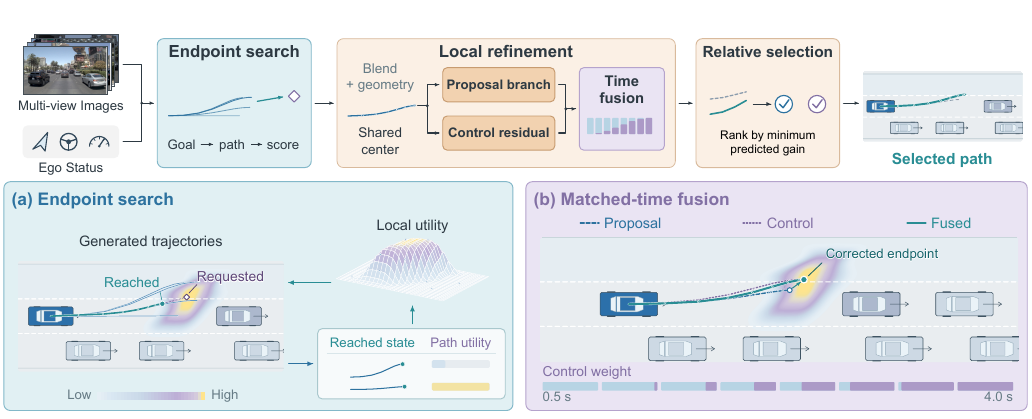}
\caption{\textbf{DriveTTO overview.} Across the top, endpoint search supplies a path for local refinement. The proposal and control branches share a starting trajectory. A fused path replaces the retained branch output only if both scorers predict a gain. The smaller gain ranks eligible blends. In (a), the utility of a completed path is recorded at its reached endpoint and guides the next requests. In (b), the control weight varies along the horizon as the two paths are blended at matching times. Paths and utilities are schematic.}
\label{fig:framework}
\end{figure*}

\begin{algorithm}[t]
\caption{DriveTTO trajectory optimization (Full)}
\label{alg:drivetto}
\small
\renewcommand{\algorithmicrequire}{\textbf{Input}}
\renewcommand{\algorithmicensure}{\textbf{Output}}
\begin{algorithmic}[1]
\REQUIRE Observation $o$, native plan $n$, banks $\mathcal P_1,\mathcal P_2$, scorers $S_1,S_2$
\item[]\textit{DriveTTO-S uses $S_1=S_2$ and $\mathcal P_1=\mathcal P_2$.}
\ENSURE Selected trajectory $\hat\tau$
\item[]\textit{Endpoint search}
\STATE Initialize the endpoint reference using $n$ and groups from $\mathcal P_1$
\STATE Score completions with $S_1$ and store $(\hat g_i,u_i)$ by Eq.~\eqref{eq:achieved_state}
\STATE $f\gets$ verified winner under $S_1$ or the retained reference
\item[]\textit{Local refinement}
\STATE $d_2\gets\argmax_{d\in\mathcal P_2}S_2(d;o,\mathcal P_2)$
\STATE $b_0\gets$ highest-scoring $\mathcal I_\alpha(f,d_2)$ under $S_2$ with $\alpha\leq.5$
\STATE $b_G\gets$ geometric refinement of $b_0$ with checks by $S_2$
\STATE $a\gets$ proposal branch output scored by $S_1,S_2$ against $b_G$
\STATE $c\gets$ control branch output scored by $S_1,S_2$ against $b_G$
\item[]\textit{Temporal fusion and relative selection}
\STATE $e\gets c$ if both scorers give $c$ a larger gain than $a$, else $e\gets a$
\STATE $\mathcal F\gets\{\tau^w:w\in\mathcal W\}$ from $a,c$ by Eq.~\eqref{eq:fusion}
\STATE Use $S_1,S_2$ to score each blend's gain over $e$
\STATE Rank blends with two positive gains by the smaller gain
\RETURN Highest-ranked blend, or $e$ if none qualifies
\end{algorithmic}
\end{algorithm}

\paragraph{Comparing with the current reference.}
\label{sec:verify}
\looseness=-1
A learned score can depend on the proposal bank in which a trajectory is evaluated. Comparing scores obtained from different banks may therefore mix a trajectory preference with a change in scoring context. We evaluate each candidate together with its reference in fixed slots of the same bank and use their score difference as the predicted gain. For stages involving multiple scorer--context pairs, the minimum gain prevents a favorable prediction from compensating for an unfavorable prediction in another pair.

For candidate $c$ and reference $b$, let $\mathcal B_{m,k}(b,c)$ denote their common bank. For $S_m$, a native context draws the surrounding proposals from $\mathcal P_m$ and a shared context draws from both banks. The scorer--context pairs $\mathcal I_r$ at a minimum-gain ranking stage $r$ give
\begin{equation}
\begin{aligned}
\Delta_{m,k}(c;b)&=S_m(c;o,\mathcal B_{m,k})-S_m(b;o,\mathcal B_{m,k}),\\
q_r(c;b_r)&=\min_{(m,k)\in\mathcal I_r}\Delta_{m,k}(c;b_r).
\end{aligned}
\label{eq:paired_difference}
\end{equation}
Pairing cancels score offsets shared by the two paths within that bank. We write $\Delta_m$ for the gain predicted by $S_m$ in its native context.

\looseness=-1
Screening keeps $b_r$ eligible. Other retained candidates enter $\mathcal C_r^{\rm valid}$ if they meet the stage's motion, score, and interpolation-strength requirements. Ranking selects $c_r^\star=\argmax_{c\in\mathcal C_r^{\rm valid}}q_r(c;b_r)$, assigning $b_r$ zero gain and priority on a tie at zero. The margin $\delta_r$ and source check $W_r$ determine the next reference,
\begin{equation}
b_{r+1}=\begin{cases}
c_r^\star,&q_r(c_r^\star;b_r)\geq\delta_r\ \land\ W_r(c_r^\star)=1,\\
b_r,&\text{otherwise}.
\end{cases}
\label{eq:accept}
\end{equation}
Without a source veto, $W_r=1$. Acceptance updates the reference. Rejection retains $b_r$, including any earlier accepted error.

\subsection{Endpoint search}
\label{sec:realize_endpoint}
\looseness=-1
Narrow high-scoring regions motivate local exploration around native proposals rather than unrestricted perturbation of complete paths. Endpoint search uses their terminal states to initialize a supported search region and their intermediate geometry to complete new requests. Completion may miss the requested state, so feedback is stored at the endpoint actually reached (Figure~\ref{fig:framework}(a)). This associates utility with the trajectory that was evaluated.

A requested terminal state $g_i=(x_T,y_T,\psi_T,v_T)$ and a native parent $d_i$ give
\begin{equation}
\tau_i=\Phi(g_i,d_i),\quad \hat g_i=H(\tau_i),\quad
u_i=U(\tau_i;o)-\rho_i.
\label{eq:achieved_state}
\end{equation}
\looseness=-1
Completion $\Phi$ uses the parent's intermediate geometry to target the requested position and recomputes heading and speed. $H$ extracts the achieved state. Requested heading and speed enter the search distance and realization cost $\rho_i$. To compute $U$, $S_1$ scores completed paths alongside $n$ in a bank drawn from $\mathcal P_1$. We add each path's gain over $n$ to the original score of $n$.

\looseness=-1
We add $(\hat g_i,u_i)$ to a scene-specific archive after each valid completion. Over two rounds, resampling contracts local neighborhoods around supported endpoints while the archive accumulates feedback. Utility at a requested state is estimated from its eight nearest achieved states, using Gaussian weights on position, wrapped heading, and speed. To choose $f$, Algorithm~\ref{alg:drivetto} compares the direct completions and, when available, re-decoded paths with the reference from the native path groups. A candidate must clear the score margin in both the first and last non-anchor slots of the $S_1$ comparison bank and pass the rank and component checks. A source veto on the winner keeps the reference (Appendix~\ref{app:endpoint}).

\subsection{Local refinement}
Endpoint search leaves local geometry and timing to refine. Bounded interpolation incorporates a native alternative, and geometric search explores changes beyond that blend. The retained path anchors proposal and control branches, which exploit learned motion patterns and explore corrections beyond the banks.

\paragraph{Bounded interpolation.}
\label{sec:generators}
\looseness=-1
To bring a native alternative into the endpoint path, we blend $f$ with $d_2$, the proposal in $\mathcal P_2$ that $S_2$ ranks highest. At each time step $\mathcal I_\alpha(f,d_2)$ interpolates positions linearly and headings circularly. $S_2$ scores the five strengths $\alpha\in\{0,.25,.5,.75,1\}$ together. Its highest-scoring arm with $\alpha\leq.5$ becomes $b_0$, limiting the shift from $f$ toward $d_2$.

\paragraph{Geometric refinement.}
\looseness=-1
Geometry lets the endpoint and timing vary beyond the initial interpolation. Two rounds of elite sampling explore longitudinal and lateral endpoint offsets and time scales around the retained plan. $S_2$ ranks the paths in a five-path context and checks the winner against the entering reference in its native context. A final interpolation of strength at most .75 must also pass the native check (Appendix~\ref{app:geometry}). The accepted path, or the entering plan on rejection, becomes $b_G$. Both branches generate candidates around $b_G$ and measure gains against it, so their outputs can be compared before fusion (Figure~\ref{fig:framework}, shared center).
\label{sec:parallel}

\paragraph{Proposal branch.}
\looseness=-1
This branch tests whether moving $b_G$ toward a path in either proposal bank improves its predicted score. Each scorer ranks both banks. Up to four drafts per ranking, separated by a mean positional distance above .25\,m, are each blended with $b_G$ at five strengths. The arms with $\alpha\leq.5$ compete alongside the endpoint and initial plans. Both native-context gains must be positive for a replacement. Ranking by the smaller gain returns a path $a$ and its gain pair $g^a$ relative to $b_G$. If no replacement qualifies, $a=b_G$.

\paragraph{Control branch.}
\label{sec:control_refinement}
\looseness=-1
The proposal banks need not contain the local adjustment required by the current scene. The control residual in Figure~\ref{fig:framework} therefore adds candidates by perturbing acceleration and yaw rate. Directly replacing the stored trajectory with a rollout of reconstructed controls can introduce a reconstruction offset even when the perturbation is zero. To isolate the intended change, we add the difference between perturbed and nominal rollouts to the stored reference trajectory. This construction preserves the reference exactly at zero perturbation, while a common scale bounds the positional and angular changes.

Let $u_0$ denote controls reconstructed from $b_G$, $R$ the deterministic rollout, and $Bz$ a piecewise-linear perturbation with four knots per channel. The candidate is
\begin{equation}
\begin{aligned}
\Delta R(z)&=R(u_0+Bz)\ominus R(u_0),\\
c(z)&=b_G\oplus\kappa(z)\Delta R(z).
\end{aligned}
\label{eq:transport}
\end{equation}
Here $\ominus$ subtracts positions and wraps heading differences, and $\oplus$ applies the pose residual. The common scale $\kappa$ limits changes to .8\,m and .06\,rad (Appendix~\ref{app:transport}).

We draw 32 antithetic samples $z\in[-2,2]^8$ and select four elites under
\begin{equation}
L(z)=\min_{m\in\{1,2\}}\Delta_m(c(z);b_G)-10^{-3}\|z\|_2^2/8.
\label{eq:maximin_control_objective}
\end{equation}
The samples and a candidate at half the elite mean compete with $b_G$ under dual-positive admission and unpenalized minimum-gain ranking. The winner forms five interpolation arms with $b_G$. Eligible arms require two positive native-context gains and strength at most .75. Their minimum gain over the zero arm in the joint bank determines $c$, with native-context gains $g^c$ relative to $b_G$ (Appendix~\ref{app:transport}).

\begin{table*}[!t]
\caption{\textbf{Test-time optimization across frozen planners.} Full-split results compare Native, reproduced ToAD~\cite{xu2026toad}, and DriveTTO-S with shared anchors, top-64 banks, and DrivoR scorers. Navhard cells report S1/S2. Blue denotes component gains over Native, bold marks group-best aggregates, and C denotes comfort. Appendix~\ref{app:transfer_protocol} gives protocols and budgets.}
\label{tab:transfer}
\centering
\fontsize{7.5}{9.5}\selectfont
\setlength{\tabcolsep}{1pt}
\renewcommand{\arraystretch}{1.04}
\begin{tabularx}{\textwidth}{@{}l*{9}{>{\hsize=1.24\hsize\linewidth=\hsize\raggedleft\arraybackslash}X}r*{5}{>{\hsize=.568\hsize\linewidth=\hsize\raggedleft\arraybackslash}X}r@{}}
\toprule
\rowcolor{TableGroup}
& \multicolumn{10}{c}{\textbf{Navhard} (Navsim-v2)} & \multicolumn{6}{c}{\textbf{Navtest} (Navsim-v1)}\\
\cmidrule(lr){2-11}\cmidrule(l){12-17}
\rowcolor{TableHeader}
Planner / variant & NC & DAC & DDC & TL & EP & TTC & LK & HC & EC & EPDMS & NC & DAC & TTC & C & EP & PDMS\\
\midrule
ZTRS~\cite{ztrs2025} & 96.9/89.1 & 97.3/87.1 & 99.9/94.8 & 100/98.1 & 73.1/73.1 & 96.2/86.9 & 94.7/56.6 & 97.8/96.3 & 59.6/62.7 & 46.17 & 97.8 & 99.4 & 90.1 & 100 & 80.7 & 87.13\\
\quad + ToAD & \textcolor{cvprblue}{98.9}/\textcolor{cvprblue}{91.1} & \textcolor{cvprblue}{97.6}/\textcolor{cvprblue}{91.3} & 99.7/94.6 & 100/\textcolor{cvprblue}{98.4} & 71.1/70.6 & \textcolor{cvprblue}{98.4}/\textcolor{cvprblue}{88.7} & 92.4/\textcolor{cvprblue}{60.5} & 97.6/\textcolor{cvprblue}{98.5} & 53.8/52.8 & 49.69 & \textcolor{cvprblue}{98.4} & \textcolor{cvprblue}{99.6} & \textcolor{cvprblue}{92.5} & 100 & \textcolor{cvprblue}{81.2} & 88.58\\
\rowcolor{TableOurs}
\quad + \textbf{DriveTTO-S} & \textcolor{cvprblue}{98.9}/\textcolor{cvprblue}{91.9} & \textcolor{cvprblue}{98.4}/\textcolor{cvprblue}{93.5} & \textcolor{cvprblue}{100}/\textcolor{cvprblue}{95.6} & 100/\textcolor{cvprblue}{99.1} & 70.2/69.0 & \textcolor{cvprblue}{98.9}/\textcolor{cvprblue}{90.6} & 93.1/\textcolor{cvprblue}{60.5} & 97.6/\textcolor{cvprblue}{97.3} & 48.4/55.5 & \textbf{51.49} & \textcolor{cvprblue}{98.9} & \textcolor{cvprblue}{99.6} & \textcolor{cvprblue}{94.4} & 100 & \textcolor{cvprblue}{85.1} & \textbf{91.02}\\
\addlinespace[2.5pt]
GTRS~\cite{gtrs2025} & 97.6/86.8 & 96.4/84.6 & 99.2/91.3 & 99.3/98.8 & 80.2/82.7 & 96.9/84.4 & 94.9/51.9 & 97.8/95.4 & 60.4/56.1 & 43.87 & 98.9 & 99.0 & 95.7 & 100 & 84.5 & 90.93\\
\quad + ToAD & \textcolor{cvprblue}{97.8}/\textcolor{cvprblue}{91.7} & \textcolor{cvprblue}{96.9}/\textcolor{cvprblue}{89.8} & \textcolor{cvprblue}{99.4}/\textcolor{cvprblue}{94.0} & \textcolor{cvprblue}{100}/98.7 & 77.1/78.3 & \textcolor{cvprblue}{98.0}/\textcolor{cvprblue}{88.9} & 92.2/\textcolor{cvprblue}{54.2} & 97.8/\textcolor{cvprblue}{96.8} & 55.1/49.3 & 49.32 & \textcolor{cvprblue}{99.3} & \textcolor{cvprblue}{99.5} & \textcolor{cvprblue}{96.8} & 100 & \textcolor{cvprblue}{85.4} & 92.16\\
\rowcolor{TableOurs}
\quad + \textbf{DriveTTO-S} & \textcolor{cvprblue}{98.7}/\textcolor{cvprblue}{92.0} & \textcolor{cvprblue}{97.1}/\textcolor{cvprblue}{90.8} & \textcolor{cvprblue}{99.7}/\textcolor{cvprblue}{94.3} & \textcolor{cvprblue}{100}/\textcolor{cvprblue}{99.1} & 76.1/76.2 & \textcolor{cvprblue}{98.4}/\textcolor{cvprblue}{90.2} & \textcolor{cvprblue}{95.6}/\textcolor{cvprblue}{52.0} & 97.8/95.1 & 47.1/45.4 & \textbf{49.43} & \textcolor{cvprblue}{99.3} & \textcolor{cvprblue}{99.6} & \textcolor{cvprblue}{96.9} & 100 & \textcolor{cvprblue}{88.4} & \textbf{93.52}\\
\addlinespace[2.5pt]
Hydra-MDP~\cite{li2024hydramdp} & 98.4/85.2 & 95.1/82.5 & 99.6/89.6 & 99.8/98.5 & 82.7/85.2 & 98.0/82.4 & 95.6/50.6 & 97.8/95.6 & 66.7/48.8 & 41.17 & 98.5 & 98.8 & 94.9 & 100 & 85.5 & 90.83\\
\quad + ToAD & 98.0/\textcolor{cvprblue}{89.9} & \textcolor{cvprblue}{96.9}/\textcolor{cvprblue}{88.0} & 99.3/\textcolor{cvprblue}{93.8} & \textcolor{cvprblue}{100}/\textcolor{cvprblue}{98.6} & 78.9/80.2 & 97.8/\textcolor{cvprblue}{87.3} & 93.1/\textcolor{cvprblue}{51.2} & 97.8/\textcolor{cvprblue}{97.6} & 56.0/48.2 & \textbf{48.08} & \textcolor{cvprblue}{99.0} & \textcolor{cvprblue}{99.4} & \textcolor{cvprblue}{96.2} & 100 & \textcolor{cvprblue}{86.5} & 92.28\\
\rowcolor{TableOurs}
\quad + \textbf{DriveTTO-S} & \textcolor{cvprblue}{98.6}/\textcolor{cvprblue}{90.2} & \textcolor{cvprblue}{97.6}/\textcolor{cvprblue}{88.5} & 99.6/\textcolor{cvprblue}{93.3} & \textcolor{cvprblue}{100}/98.5 & 78.6/78.6 & \textcolor{cvprblue}{98.2}/\textcolor{cvprblue}{87.7} & 94.9/\textcolor{cvprblue}{52.3} & 97.8/\textcolor{cvprblue}{97.3} & 54.7/48.5 & 47.63 & \textcolor{cvprblue}{99.1} & \textcolor{cvprblue}{99.4} & \textcolor{cvprblue}{96.6} & 100 & \textcolor{cvprblue}{88.9} & \textbf{93.46}\\
\addlinespace[2.5pt]
iPad~\cite{ipad2025} & 96.4/82.9 & 96.2/83.1 & 96.4/85.5 & 99.1/97.4 & 82.9/85.2 & 96.4/80.0 & 93.6/50.7 & 97.8/93.5 & 45.8/38.9 & 36.84 & 98.6 & 98.4 & 94.9 & 99.8 & 88.1 & 91.79\\
\quad + ToAD & \textcolor{cvprblue}{97.9}/\textcolor{cvprblue}{91.9} & \textcolor{cvprblue}{98.4}/\textcolor{cvprblue}{91.7} & \textcolor{cvprblue}{99.3}/\textcolor{cvprblue}{95.0} & \textcolor{cvprblue}{99.8}/\textcolor{cvprblue}{98.6} & 76.4/76.8 & \textcolor{cvprblue}{98.0}/\textcolor{cvprblue}{90.8} & 92.7/\textcolor{cvprblue}{54.1} & 97.1/\textcolor{cvprblue}{96.1} & 43.6/38.3 & \textbf{50.61} & \textcolor{cvprblue}{98.9} & \textcolor{cvprblue}{99.5} & \textcolor{cvprblue}{96.0} & \textcolor{cvprblue}{100} & \textcolor{cvprblue}{89.3} & 93.41\\
\rowcolor{TableOurs}
\quad + \textbf{DriveTTO-S} & \textcolor{cvprblue}{98.7}/\textcolor{cvprblue}{90.8} & \textcolor{cvprblue}{98.0}/\textcolor{cvprblue}{89.2} & \textcolor{cvprblue}{99.8}/\textcolor{cvprblue}{92.3} & \textcolor{cvprblue}{99.8}/\textcolor{cvprblue}{98.7} & 75.8/76.0 & \textcolor{cvprblue}{98.2}/\textcolor{cvprblue}{88.0} & \textcolor{cvprblue}{95.3}/\textcolor{cvprblue}{52.6} & 95.8/\textcolor{cvprblue}{94.1} & \textcolor{cvprblue}{47.1}/\textcolor{cvprblue}{46.7} & 47.82 & \textcolor{cvprblue}{99.2} & \textcolor{cvprblue}{99.4} & \textcolor{cvprblue}{97.2} & \textcolor{cvprblue}{99.9} & \textcolor{cvprblue}{90.1} & \textbf{94.23}\\
\addlinespace[2.5pt]
RAP-DINO~\cite{rap2025} & 97.8/80.9 & 96.7/81.4 & 97.8/86.7 & 99.8/98.1 & 85.9/88.5 & 97.1/76.5 & 96.4/52.6 & 96.7/95.4 & 65.3/54.4 & 38.62 & 99.1 & 98.9 & 96.7 & 100 & 90.4 & 93.82\\
\quad + ToAD & \textcolor{cvprblue}{98.3}/\textcolor{cvprblue}{87.0} & 96.0/\textcolor{cvprblue}{84.5} & \textcolor{cvprblue}{99.2}/\textcolor{cvprblue}{91.3} & 99.8/\textcolor{cvprblue}{98.2} & 83.2/86.6 & \textcolor{cvprblue}{98.2}/\textcolor{cvprblue}{84.2} & 92.7/49.7 & \textcolor{cvprblue}{97.6}/\textcolor{cvprblue}{95.6} & \textcolor{cvprblue}{71.1}/\textcolor{cvprblue}{57.0} & 45.72 & \textcolor{cvprblue}{99.2} & \textcolor{cvprblue}{99.4} & \textcolor{cvprblue}{97.0} & 100 & \textcolor{cvprblue}{90.8} & 94.46\\
\rowcolor{TableOurs}
\quad + \textbf{DriveTTO-S} & \textcolor{cvprblue}{98.4}/\textcolor{cvprblue}{91.2} & \textcolor{cvprblue}{98.0}/\textcolor{cvprblue}{91.7} & \textcolor{cvprblue}{99.3}/\textcolor{cvprblue}{93.6} & 99.8/\textcolor{cvprblue}{98.9} & 76.3/77.5 & \textcolor{cvprblue}{98.9}/\textcolor{cvprblue}{88.9} & 93.8/\textcolor{cvprblue}{55.1} & \textcolor{cvprblue}{97.6}/\textcolor{cvprblue}{98.0} & 58.7/\textcolor{cvprblue}{58.3} & \textbf{52.55} & \textcolor{cvprblue}{99.3} & \textcolor{cvprblue}{99.6} & \textcolor{cvprblue}{97.4} & 100 & 90.3 & \textbf{94.57}\\
\addlinespace[2.5pt]
DrivoR~\cite{kirby2026registers} & 99.1/92.3 & 98.2/91.6 & 99.3/97.3 & 99.8/99.1 & 75.4/75.6 & 98.7/90.6 & 94.9/56.1 & 97.6/98.4 & 70.2/44.8 & 54.56 & 98.9 & 99.2 & 96.9 & 100 & 91.6 & 94.54\\
\quad + ToAD & 98.9/\textcolor{cvprblue}{93.4} & 98.2/\textcolor{cvprblue}{93.7} & 99.3/\textcolor{cvprblue}{97.5} & 99.8/98.8 & 74.7/74.2 & \textcolor{cvprblue}{98.9}/\textcolor{cvprblue}{91.6} & 92.9/\textcolor{cvprblue}{56.7} & 97.6/97.7 & 65.8/\textcolor{cvprblue}{46.7} & 55.64 & 98.9 & \textcolor{cvprblue}{99.4} & 96.8 & 100 & \textcolor{cvprblue}{91.8} & 94.72\\
\rowcolor{TableOurs}
\quad + \textbf{DriveTTO-S} & 99.1/\textcolor{cvprblue}{93.2} & \textcolor{cvprblue}{98.7}/\textcolor{cvprblue}{93.9} & \textcolor{cvprblue}{99.4}/96.7 & \textcolor{cvprblue}{100}/98.8 & 74.8/75.3 & \textcolor{cvprblue}{98.9}/\textcolor{cvprblue}{91.1} & 94.0/\textcolor{cvprblue}{57.4} & 97.6/98.2 & 64.9/\textcolor{cvprblue}{45.9} & \textbf{56.23} & \textcolor{cvprblue}{99.0} & \textcolor{cvprblue}{99.4} & 96.9 & 100 & \textcolor{cvprblue}{92.0} & \textbf{94.87}\\
\bottomrule
\end{tabularx}
\end{table*}

\subsection{Temporal fusion and relative selection}
\paragraph{Temporal fusion.}
\looseness=-1
A proposal may supply a useful turn while a control correction improves the approach. Matched-time fusion therefore allows the contribution of each branch to vary along the horizon (Figure~\ref{fig:framework}(b)). We form complete paths using either constant or time-varying weights.

We first choose the branch output that a blend must improve on. This fusion reference $e$ is $c$ only if $g^c_1>g^a_1$ and $g^c_2>g^a_2$. Otherwise $e=a$. The control weight $w_t$ combines the two poses at time $t$ to form $\tau^w$,
\begin{equation}
\begin{aligned}
p_t^w&=(1-w_t)p_t^a+w_t p_t^c,\\
\psi_t^w&=\arg\!\left((1-w_t)e^{\mathrm i\psi_t^a}+w_t e^{\mathrm i\psi_t^c}\right).
\end{aligned}
\label{eq:fusion}
\end{equation}
\looseness=-1
The profile set $\mathcal W$ contains five constant weights and two time-varying profiles, $w_t=h(s_t)$ and $1-h(s_t)$, where $s_t=(t-1)/(T-1)$ and $h(s)=s^2(3-2s)$. A changing weight introduces the velocity term $\dot w(p^c-p^a)$, so the scorers evaluate each blended trajectory in full.

\paragraph{Relative selection.}
Only blends that both scorers prefer to $e$ in their native contexts can replace it. We return the eligible blend with the largest minimum gain, or $e$ if none qualifies. Rollback entries at strengths $0/.5/.75$ provide scoring context and are ineligible for replacement. Exact-array deduplication retains the first source.

\endgroup
\section{Experiments}
\label{sec:experiments}
We test whether DriveTTO improves benchmark scores, transfers across planners, and recovers native failures. Ablations test which parts of Figure~\ref{fig:framework} contribute to those gains. Fixed-pool controls isolate how the scorers rank candidates and screen updates.

\subsection{Experimental setup}
\label{sec:protocol}
\looseness=-1
We evaluate frozen planners on 12,146 Navtest scenes (PDMS and EPDMS), 450 original Navhard scenes with 5,462 linked continuations (two-stage EPDMS), and 345 closed-loop HUGSIM scenarios~\cite{dauner2024navsim,cao2025pseudo,zhou2025hugsim}. Controlled Navtest PDMS comparisons instantiate Algorithm~\ref{alg:drivetto} with frozen ChainFlow DiffGRPO providing $n,\mathcal P_1,S_1$ and DrivoR Navsim-v1 providing $\mathcal P_2,S_2$. Native returns $n$. Both models use camera images, ego history, and a navigation command to evaluate eight poses at half-second intervals. Table~\ref{tab:configuration_map} lists the checkpoints and search schedules for the other settings. For these controlled Navtest comparisons, we average three search seeds per scene and resample 44 parent logs for paired intervals. These intervals are conditional on configurations chosen using evaluation scenes (Appendix~\ref{app:datasets}).

\subsection{Benchmark performance and planner transfer}
\paragraph{Navtest performance.}
\label{sec:pdms}
\label{sec:navtest_epdms}
\looseness=-1
DriveTTO leads the listed methods on aggregate PDMS and EPDMS, with lower extended comfort than SUV (Tables~\ref{tab:pdms} and~\ref{tab:navtest_epdms}). The benchmark PDMS is 95.54, rounded to 95.5, whereas controlled Full averages 95.51 over three seeds (Table~\ref{tab:failure_recovery}). Published checkpoints and inference budgets differ, so we isolate search effects in Section~\ref{sec:system_comparison}.

\setcounter{dbltopnumber}{2}
\begin{table}[!tb]
\begin{minipage}{\linewidth}
\captionsetup{type=table}
\caption{\textbf{Navtest PDMS.} Selected published methods and five components. C denotes comfort. Higher is better.}
\label{tab:pdms}
\centering
\fontsize{8.5}{9.5}\selectfont
\setlength{\tabcolsep}{1.7pt}
\renewcommand{\arraystretch}{1.05}
\begin{tabularx}{\linewidth}{@{}l*{5}{>{\raggedleft\arraybackslash}X}r@{}}
\toprule
\rowcolor{TableHeader}Method & NC & DAC & TTC & C & EP & PDMS\\
\midrule
TransFuser~\cite{chitta2023transfuser} & 97.7 & 92.8 & 92.8 & \textbf{100.0} & 79.2 & 84.0\\
Hydra-MDP~\cite{li2024hydramdp} & 98.3 & 96.0 & 94.6 & \textbf{100.0} & 78.7 & 86.5\\
DiffusionDrive~\cite{diffusiondrive2024} & 98.2 & 96.2 & 94.7 & \textbf{100.0} & 82.2 & 88.1\\
SUV~\cite{suv2026} & 99.1 & 97.8 & 96.7 & \textbf{100.0} & 84.6 & 90.8\\
Auto-JEPA~\cite{autojepa2026} & 98.4 & 98.3 & 95.0 & \textbf{100.0} & 87.1 & 91.3\\
iPad~\cite{ipad2025} & 98.6 & 98.3 & 94.9 & 99.9 & 88.0 & 91.7\\
SparseDriveV2~\cite{sparsedrivev22026} & 98.5 & 98.4 & 95.0 & 99.9 & 88.6 & 92.0\\
DriveSuprim~\cite{drivesuprim2025} & 98.6 & 98.6 & 95.5 & \textbf{100.0} & 91.3 & 93.5\\
RAP~\cite{rap2025} & 99.1 & 98.9 & 96.7 & \textbf{100.0} & 90.3 & 93.8\\
CLOVER~\cite{clover2026} & 98.6 & 99.1 & 95.8 & \textbf{100.0} & 92.8 & 94.5\\
DrivoR~\cite{kirby2026registers} & 99.1 & 99.2 & 96.9 & \textbf{100.0} & 91.6 & 94.6\\
ChainFlow-VLA~\cite{wang2026chainflow} & 99.2 & 99.0 & 97.2 & 99.9 & 91.9 & 94.8\\
\midrule
\rowcolor{TableOurs}
\textbf{DriveTTO} & \textbf{99.3} & \textbf{99.6} & \textbf{97.3} & \textbf{100.0} & \textbf{92.9} & \textbf{95.5}\\
\bottomrule
\end{tabularx}
\end{minipage}\par
\vspace{\floatsep}
\begin{minipage}{\linewidth}
\captionsetup{type=table}
\caption{\textbf{Navtest EPDMS.} Full-split results under the corrected evaluator. DriveVLA-W0, EponaV2, and Metis follow SUV's comparison~\cite{suv2026}. Higher is better.}
\label{tab:navtest_epdms}
\centering
\fontsize{7.5}{9.5}\selectfont
\setlength{\tabcolsep}{.75pt}
\renewcommand{\arraystretch}{1.05}
\begin{tabular*}{\linewidth}{@{\extracolsep{\fill}}l*{10}{r}@{}}
\toprule
\rowcolor{TableHeader}Method & NC & DAC & DDC & TL & EP & TTC & LK & HC & EC & EPDMS\\
\midrule
DriveVLA-W0~\cite{drivevlaw02025} & 98.4 & 95.2 & 99.4 & \textbf{99.9} & 86.6 & 97.9 & 97.8 & 98.3 & 82.7 & 86.9\\
EponaV2~\cite{eponav22026} & 98.5 & 97.4 & 99.5 & \textbf{99.9} & 87.9 & 98.1 & 97.7 & 98.2 & 77.4 & 88.9\\
Auto-JEPA~\cite{autojepa2026} & 98.5 & 98.7 & 98.3 & 99.7 & 90.5 & 97.9 & 94.7 & 97.8 & 75.2 & 89.1\\
Metis~\cite{metis2026} & 98.4 & 97.2 & 99.6 & 99.8 & 87.8 & 97.7 & 97.8 & \textbf{98.4} & 88.0 & 89.5\\
DriveFuture~\cite{drivefuture2026} & 98.8 & 99.1 & 99.6 & \textbf{99.9} & 86.6 & 98.4 & 96.4 & 98.3 & 74.8 & 89.9\\
SparseDriveV2~\cite{sparsedrivev22026} & 98.1 & 98.1 & 99.6 & 99.8 & \textbf{91.1} & 97.3 & 96.9 & 98.2 & 78.4 & 90.1\\
CLOVER~\cite{clover2026} & \textbf{99.4} & 99.3 & 99.5 & 99.8 & 86.9 & \textbf{98.9} & 95.2 & 98.2 & 75.5 & 90.4\\
SUV~\cite{suv2026} & 99.1 & 97.8 & \textbf{99.7} & 99.8 & 87.8 & 98.7 & \textbf{98.1} & \textbf{98.4} & \textbf{88.3} & 91.0\\
\midrule
\rowcolor{TableOurs}
\textbf{DriveTTO} & 99.3 & \textbf{99.6} & \textbf{99.7} & 99.7 & 90.5 & \textbf{98.9} & 95.2 & 98.3 & 80.7 & \textbf{92.3}\\
\bottomrule
\end{tabular*}
\end{minipage}
\end{table}

\paragraph{Transfer across frozen planners.}
\label{sec:transfer}
\looseness=-1
To test transfer, DriveTTO-S and ToAD~\cite{xu2026toad} receive each target planner's native plan and top-64 proposals with the same frozen DrivoR scorer. DriveTTO-S reuses that bank and scorer in both roles, using Navsim-v1 on Navtest and Navsim-v2 on Navhard. It improves all six native planners on both benchmarks (Table~\ref{tab:transfer}) and outscores reproduced ToAD on all six planners on Navtest and four on Navhard. For RAP-DINO on Navhard, second-stage collision avoidance and drivable-area compliance improve while progress falls.

\begin{table*}[t]
\caption{\textbf{Navhard EPDMS.} Published results with standard per-scene inputs. Components report S1/S2. Blue marks component gains over the DrivoR row for ToAD and DriveTTO. Bold marks the best EPDMS. Local reproductions appear in Table~\ref{tab:transfer}.}
\label{tab:navhard_epdms}
\centering
\fontsize{8.5}{9.5}\selectfont
\setlength{\tabcolsep}{1.5pt}
\renewcommand{\arraystretch}{1.04}
\begin{tabularx}{\linewidth}{@{}l*{9}{>{\raggedleft\arraybackslash}X}r@{}}
\toprule
\rowcolor{TableGroup}
& \multicolumn{9}{c}{\textbf{Stage components} (S1/S2)} & \textbf{Overall}\\
\cmidrule(lr){2-10}\cmidrule(l){11-11}
\rowcolor{TableHeader}
Planner / variant & NC & DAC & DDC & TL & EP & TTC & LK & HC & EC & EPDMS $\uparrow$\\
\midrule
CLOVER~\cite{clover2026} & 99.4/89.8 & 98.0/88.2 & 99.9/93.2 & 100/99.0 & 75.7/74.2 & 98.9/87.7 & 94.4/52.6 & 97.1/97.0 & 56.0/57.3 & 48.3\\
GTRS-E~\cite{gtrs2025} & 98.9/92.3 & 99.3/93.3 & 99.8/94.6 & 99.8/99.2 & 75.2/73.1 & 98.4/91.2 & 96.0/53.9 & 97.6/96.7 & 51.6/56.8 & 49.4\\
Gigapixel~\cite{gigapixel2026} & 99.4/93.7 & 95.8/92.9 & 99.4/96.0 & 99.8/98.9 & 68.1/62.6 & 99.6/90.7 & 91.8/60.2 & 97.6/98.2 & 49.8/58.8 & 50.1\\
DriveZero~\cite{drivezero2026} & 98.7/90.5 & 98.9/89.2 & 99.9/94.6 & 99.8/98.6 & 79.5/81.2 & 98.7/88.3 & 95.8/57.2 & 97.6/94.5 & 51.6/41.3 & 51.5\\
DriveFuture~\cite{drivefuture2026} & 99.8/90.6 & 99.8/87.5 & 100/94.1 & 99.6/99.1 & 85.7/84.6 & 99.8/88.8 & 98.7/58.3 & 97.6/93.5 & 66.2/45.6 & 55.5\\
\addlinespace[2.5pt]
DrivoR~\cite{kirby2026registers} & 99.1/92.3 & 98.2/91.6 & 99.3/97.3 & 99.8/99.1 & 75.4/75.7 & 98.7/90.6 & 94.9/56.1 & 97.6/98.4 & 70.2/44.7 & 54.6\\
\quad + ToAD~\cite{xu2026toad} & 98.9/\textcolor{cvprblue}{93.6} & 98.2/\textcolor{cvprblue}{94.0} & \textcolor{cvprblue}{99.4}/\textcolor{cvprblue}{97.5} & \textcolor{cvprblue}{100}/98.8 & 74.6/74.1 & \textcolor{cvprblue}{98.9}/\textcolor{cvprblue}{92.0} & 94.0/\textcolor{cvprblue}{57.1} & 97.6/98.0 & 66.7/\textcolor{cvprblue}{47.3} & 56.3\\
\midrule
\rowcolor{TableOurs}
\textbf{DriveTTO} & 99.1/\textcolor{cvprblue}{93.9} & 98.0/\textcolor{cvprblue}{94.7} & \textcolor{cvprblue}{99.6}/\textcolor{cvprblue}{97.4} & \textcolor{cvprblue}{100}/\textcolor{cvprblue}{99.3} & 74.0/74.3 & \textcolor{cvprblue}{99.1}/\textcolor{cvprblue}{92.1} & 91.6/\textcolor{cvprblue}{58.4} & 97.6/96.3 & 65.3/\textcolor{cvprblue}{47.9} & \textbf{57.1}\\
\bottomrule
\end{tabularx}
\end{table*}
\begin{table*}[!t]
\caption{\textbf{HUGSIM component and overall scores.} Overall scores weight all 345 scenarios equally. DriveTTO uses observed commands. C denotes comfort and HD denotes HD-Score. Baselines follow ToAD~\cite{xu2026toad} and DrivoR~\cite{kirby2026registers}. Appendix~\ref{app:hugsim_results} gives matched-command controls and CL-PDMS.}
\label{tab:hugsim}
\centering
\fontsize{7}{9.5}\selectfont
\setlength{\tabcolsep}{.8pt}
\renewcommand{\arraystretch}{1.05}
\begin{tabularx}{\linewidth}{@{}l*{26}{>{\raggedleft\arraybackslash}X}@{}}
\toprule
\rowcolor{TableGroup}& \multicolumn{6}{c}{KITTI-360} & \multicolumn{6}{c}{nuScenes} & \multicolumn{6}{c}{PandaSet} & \multicolumn{6}{c}{Waymo} & \multicolumn{2}{c}{\textbf{Overall}}\\
\cmidrule(lr){2-7}\cmidrule(lr){8-13}\cmidrule(lr){14-19}\cmidrule(lr){20-25}\cmidrule(l){26-27}
\rowcolor{TableHeader}Method & NC & DAC & TTC & C & RC & HD & NC & DAC & TTC & C & RC & HD & NC & DAC & TTC & C & RC & HD & NC & DAC & TTC & C & RC & HD & RC & HD\\
\midrule
UniAD~\cite{uniad2022} & 48.1 & 85.1 & 19.8 & 37.2 & 23.5 & 5.9 & \textbf{71.8} & 99.2 & 63.3 & 79.3 & 49.1 & 37.5 & \textbf{68.2} & \textbf{100} & \textbf{58.1} & 77.3 & 49.5 & 36.7 & \textbf{76.9} & 94.3 & 66.3 & 71.3 & 56.2 & 44.4 & 45.9 & 32.7\\
VAD~\cite{jiang2023vad} & 24.4 & 83.2 & 15.0 & 94.2 & 17.1 & 3.9 & 57.2 & 97.6 & 42.9 & 97.4 & 43.1 & 24.5 & 42.4 & \textbf{100} & 26.5 & 94.9 & 38.7 & 15.4 & 46.6 & 85.5 & 32.6 & 94.9 & 29.4 & 11.0 & 31.4 & 13.4\\
GTRS-A~\cite{gtrs2025} & 39.8 & 86.1 & 33.8 & 95.5 & 27.5 & 15.3 & 61.0 & 99.9 & 54.7 & 94.6 & 58.4 & 42.9 & 51.6 & \textbf{100} & 43.3 & 95.2 & 53.0 & 33.4 & 58.8 & 96.7 & 52.5 & 95.7 & 47.1 & 33.4 & 46.3 & 31.7\\
GTRS-D~\cite{gtrs2025} & 48.2 & 90.8 & 43.7 & 94.1 & 28.3 & 18.9 & 70.4 & 99.8 & 63.7 & 96.3 & 58.5 & 46.7 & 56.0 & \textbf{100} & 48.8 & 91.9 & 54.2 & 36.6 & 65.3 & 96.8 & 60.9 & 95.0 & 52.3 & 40.9 & 48.6 & 36.7\\
DrivoR~\cite{kirby2026registers} & 44.0 & 89.9 & 40.8 & 93.6 & 28.8 & 17.7 & 57.0 & \textbf{100} & 51.3 & 93.0 & 54.2 & 39.7 & 50.8 & 99.9 & 44.0 & 89.8 & 58.3 & 35.1 & 64.0 & 96.8 & 58.6 & 94.1 & 56.5 & 44.2 & 49.8 & 35.7\\
DrivoR + ToAD~\cite{xu2026toad} & \textbf{49.5} & 90.9 & \textbf{46.1} & \textbf{99.1} & 26.9 & 18.4 & 69.1 & \textbf{100} & \textbf{63.9} & \textbf{98.3} & \textbf{60.9} & \textbf{49.9} & 57.3 & \textbf{100} & 51.6 & \textbf{98.4} & 54.5 & 38.0 & 70.3 & 96.7 & \textbf{66.6} & \textbf{97.9} & 51.1 & 42.8 & 48.5 & 38.3\\
\midrule
\rowcolor{TableOurs}
\textbf{DriveTTO} & 47.0 & \textbf{91.1} & 42.8 & 90.7 & \textbf{33.2} & \textbf{21.6} & 55.5 & 99.8 & 48.6 & 95.2 & 54.1 & 38.5 & 50.0 & \textbf{100} & 42.9 & 93.8 & \textbf{60.6} & \textbf{39.0} & 70.1 & \textbf{99.0} & 66.4 & 93.0 & \textbf{65.4} & \textbf{52.9} & \textbf{54.5} & \textbf{40.1}\\
\bottomrule
\end{tabularx}
\par\vspace{2pt}
{\fontsize{8}{9}\selectfont\textbf{DriveTTO overall components}\quad \textit{NC} 58.2\quad \textit{DAC} 97.5\quad \textit{TTC} 53.0\quad \textit{Comfort} 93.2\quad \textit{CL-PDMS} 51.3.}
\end{table*}

\paragraph{Two-stage continuation.}
\label{sec:navhard_epdms}
\label{sec:warmup}
\looseness=-1
Navhard evaluates subsequent driving through continuations weighted by proximity to the selected endpoint. DriveTTO exceeds published DrivoR + ToAD in aggregate EPDMS and second-stage drivable-area compliance (Table~\ref{tab:navhard_epdms}). On the eight equally weighted pairs in ToAD's warmup subset, the same frozen predictions score 57.89 EPDMS compared with 61.61 for native DrivoR (Table~\ref{tab:warmup}). The aggregate gain therefore does not hold on every subset. Component means alone do not determine the weighted EPDMS products (Appendix~\ref{app:warmup}).

\paragraph{Closed-loop driving.}
\label{sec:hugsim}
\looseness=-1
In closed-loop HUGSIM, each plan affects subsequent observations. Under the same fixed-straight command, route completion rises from 48.2 to 52.5 and HD-Score from 34.0 to 38.1 for base DriveTTO versus reproduced DrivoR (Appendix~\ref{app:hugsim_results}). With observed commands, DriveTTO reaches 54.5 route completion and 40.1 HD-Score overall (Table~\ref{tab:hugsim}). DrivoR was not evaluated under this command setting.

\subsection{Recovery of native planning failures}
\label{sec:failure_recovery}
\looseness=-1
To quantify improvements in difficult scenes, we count recovered native zero-score plans alongside newly introduced failures. Of 198 native zeros, Full recovers 101.67 per run on average (51.35\%) and introduces 41.00 new zeros (Table~\ref{tab:failure_recovery}). Final zeros decrease to 137.33, reducing the overall rate from 1.63\% to 1.13\%. Recovery outweighs new failures under this metric, and gains over Native concentrate on low-scoring plans (Figure~\ref{fig:failure_recovery_distribution}).

\subsection{Search and refinement ablations}
\label{sec:system_comparison}
\begin{figure}[!htb]
\centering
\includegraphics[trim=20bp 0bp 0bp 0bp,clip,width=\columnwidth]{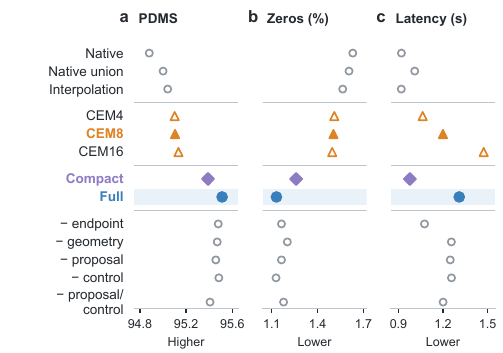}
\caption{\looseness=-1 \textbf{Quality, failures, and computation.} Thirteen Navtest configurations. Panels show (a) mean PDMS, (b) mean per-seed zero-score rate, and (c) median encoding/search latency. Minus denotes stage or branch removal with downstream reruns. Section~\ref{sec:protocol} specifies averaging. Paired intervals appear in Figure~\ref{fig:paired_quality_differences} and Appendix~\ref{app:branch_intervals}.}
\label{fig:quality_latency}
\end{figure}

\paragraph{Structured search at comparable cost.}
\looseness=-1
To compare search strategies at similar cost, we fix both models and the input observations. We select CEM8 by latency before official scoring. Full exceeds CEM8 by 0.41 PDMS (95\% CI [0.11,0.73]) with fewer zero-score plans at similar median latency (Figure~\ref{fig:quality_latency}). Doubling the CEM budget increases latency and leaves mean PDMS below Full.

\paragraph{Search beyond reranking.}
\label{sec:cost}
\looseness=-1
To isolate gains beyond native proposals, we compare DriveTTO-S with top-64 reranking. It improves all five Navtest planners tested (Appendix~\ref{app:transfer_effects}), with a median search-time ratio of 3.8 to ToAD using cached features (Table~\ref{tab:transfer_phases}). On Navhard, smoothing yields no gain for DrivoR or Hydra-MDP. DriveTTO-S improves on reranking for Hydra-MDP (95\% CI [0.21,4.54] EPDMS), while the DrivoR interval spans zero. A separate five-round ToAD run with a different sampling stream shows no clear advantage for either search method (Appendix~\ref{app:refinement_controls}).

\begin{figure}[!htb]
\centering
\includegraphics[width=\columnwidth]{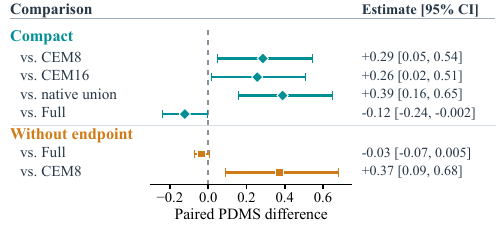}
\caption{\looseness=-1 \textbf{Simplified search configurations.} Paired PDMS differences with individual exploratory 95\% intervals from 2,000 bootstrap draws over 44 logs, with seeds averaged per scene. Endpoint removal bypasses native-path initialization, and Compact retains proposal refinement with constant fusion.}
\label{fig:paired_quality_differences}
\end{figure}
\paragraph{Endpoint and branch construction.}
\looseness=-1
To assess what each stage adds to the complete system, we remove it and rerun downstream search. Removing both refinement branches lowers mean PDMS (Figure~\ref{fig:quality_latency}(a)). Without endpoint search, the system still outperforms CEM8, while its paired difference from Full has an interval that includes zero (Figure~\ref{fig:paired_quality_differences}). Compact retains initial mixing, proposal refinement, and constant fusion. It preserves about four-fifths of Full's gain over Native. The noninferiority test against Full is inconclusive (Appendix~\ref{app:branch_intervals}).

\paragraph{Feedback coordinates and matched-time fusion.}
\looseness=-1
The lower panels of Figure~\ref{fig:framework} motivate two controlled comparisons. In (a), recording utility at the requested endpoint changes the result in only one scene relative to using the reached endpoint. In (b), allowing time-varying weights on identical branch outputs changes mean PDMS by $-.0015$ (95\% CI [$-.02,.01$]). Intervals for the residual rollout control also include zero (Appendix~\ref{app:implementation_controls}).

\begin{table}[!htb]
\caption{\looseness=-1 \textbf{Candidate opportunity and output.} $O$ is the best available PDMS and $J$ the selected PDMS. Single ($s$) and joint ($d$) selection match update counts.}
\label{tab:opportunity}
\centering\small\setlength{\tabcolsep}{3pt}
\begin{tabularx}{\columnwidth}{@{}>{\raggedright\arraybackslash}Xrr>{\columncolor{TableOurs}}r@{}}
\toprule
\rowcolor{TableHeader}
& \multicolumn{2}{c}{Expansion vs. top-64} & \cellcolor{TableOursHead}Same pool\\
\cmidrule(lr){2-3}\cmidrule(l){4-4}
Planner & $\Delta O$ & $\Delta J_s$ & \cellcolor{TableOursHead}$J_d-J_s$\\
\midrule
ZTRS~\cite{ztrs2025} & +1.47 & -1.49 & +1.77\\
GTRS~\cite{gtrs2025} & +0.50 & -1.50 & +1.45\\
Hydra-MDP~\cite{li2024hydramdp} & +0.49 & -1.75 & +1.85\\
iPad~\cite{ipad2025} & +0.36 & -1.88 & +2.34\\
RAP-DINO~\cite{rap2025} & +0.45 & -1.49 & +2.25\\
DrivoR~\cite{kirby2026registers} & +0.22 & -1.89 & +2.18\\
\bottomrule
\end{tabularx}

\end{table}

\subsection{Candidate access and scorer reliability}
\label{sec:mechanism_ablation}
\label{sec:selector_ablation}
\label{sec:opportunity_analysis}
\paragraph{Availability versus selection.}
\looseness=-1
We evaluate saved pools offline to separate limited candidate access from scoring errors. Restricting the full banks of ZTRS, GTRS, and Hydra-MDP to their top 64 removes 6.02, 3.46, and 3.21 points of oracle PDMS, respectively (Table~\ref{tab:native_access}). Across all six planners, expansion raises the best available PDMS and lowers the PDMS selected by one scorer (Table~\ref{tab:opportunity}). Table~\ref{tab:opportunity_full} gives the intervals.

\begin{figure*}[!t]
\centering
\includegraphics[width=\textwidth]{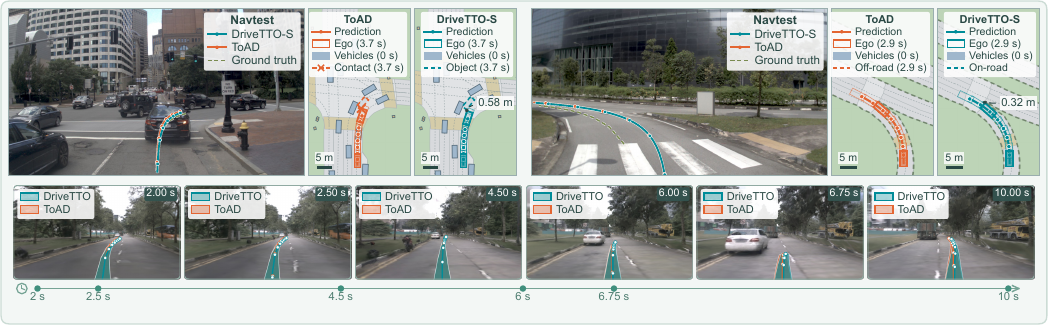}
\caption{\textbf{Selected trajectory comparisons.} Navtest shows full front views and separate ToAD/DriveTTO-S BEVs with matched bounds. Orange/teal denote ToAD/DriveTTO and dashed green denotes human motion. Footprints mark the first ToAD failure, with simultaneous DriveTTO-S obstacle or boundary clearance. Camera paths use calibrated ground-plane projection. HUGSIM reruns ToAD on DriveTTO observations and ego histories, projecting both plans from the same ego pose. Appendix~\ref{app:qualitative} includes regressions.}
\label{fig:qualitative_main}
\end{figure*}

\paragraph{Dual scoring for ranking and screening.}
\label{sec:maximin_ablation}
\looseness=-1
We add 960 diagnostic candidates to each top-64 bank. Joint scoring improves output at matched update counts (Table~\ref{tab:opportunity}), although different scenes can receive updates. Table~\ref{tab:ranking_screening} separates candidate ranking from update screening on these fixed pools.

\looseness=-1
After removing copies of the reference, we require both policies to select a replacement from the same pool. Joint ranking gains 1.45--2.35 PDMS across six planners. Gains remain positive on the same 2,429 scenes selected by identifier. With the DrivoR proposal fixed, joint screening alone gains 1.21--2.03 points.

\looseness=-1
The two controls test separate interventions, so their gains cannot be added. In Full, changing the ranking rule at four stages while retaining both scorers for screening has little effect on mean PDMS (Appendix~\ref{app:terminal_ranking}).

\begin{table}[!t]
\caption{\textbf{Separating ranking from screening.} Joint-minus-single Navtest PDMS with 95\% intervals below. Ranking forces an update, screening fixes the proposal, and the last column fixes the same 20\% of scenes. Appendix~\ref{app:selection_interventions} details inference.}
\label{tab:ranking_screening}
\centering
\fontsize{8.5}{10}\selectfont
\setlength{\tabcolsep}{2pt}
\renewcommand{\arraystretch}{1.08}
\begin{tabularx}{\columnwidth}{@{}>{\raggedright\arraybackslash}Xrrr@{}}
\toprule
\rowcolor{TableHeader}
Planner & Ranking & Screening & \shortstack[r]{Ranking\\same 20\%}\\
\midrule
ZTRS~\cite{ztrs2025} & \shortstack[r]{+1.77\\{\fontsize{7.5}{8}\selectfont[1.42,2.08]}} & \shortstack[r]{+1.56\\{\fontsize{7.5}{8}\selectfont[1.24,1.87]}} & \shortstack[r]{+0.269\\{\fontsize{7.5}{8}\selectfont[0.162,0.396]}}\\
\rowcolor{TableAlt}
GTRS~\cite{gtrs2025} & \shortstack[r]{+1.45\\{\fontsize{7.5}{8}\selectfont[1.14,1.78]}} & \shortstack[r]{+1.21\\{\fontsize{7.5}{8}\selectfont[0.93,1.52]}} & \shortstack[r]{+0.279\\{\fontsize{7.5}{8}\selectfont[0.153,0.421]}}\\
Hydra-MDP~\cite{li2024hydramdp} & \shortstack[r]{+1.85\\{\fontsize{7.5}{8}\selectfont[1.52,2.23]}} & \shortstack[r]{+1.41\\{\fontsize{7.5}{8}\selectfont[1.10,1.76]}} & \shortstack[r]{+0.354\\{\fontsize{7.5}{8}\selectfont[0.220,0.512]}}\\
\rowcolor{TableAlt}
iPad~\cite{ipad2025} & \shortstack[r]{+2.35\\{\fontsize{7.5}{8}\selectfont[1.88,2.86]}} & \shortstack[r]{+1.92\\{\fontsize{7.5}{8}\selectfont[1.49,2.39]}} & \shortstack[r]{+0.402\\{\fontsize{7.5}{8}\selectfont[0.253,0.563]}}\\
RAP-DINO~\cite{rap2025} & \shortstack[r]{+2.25\\{\fontsize{7.5}{8}\selectfont[1.83,2.70]}} & \shortstack[r]{+2.03\\{\fontsize{7.5}{8}\selectfont[1.68,2.44]}} & \shortstack[r]{+0.388\\{\fontsize{7.5}{8}\selectfont[0.237,0.580]}}\\
\rowcolor{TableAlt}
DrivoR~\cite{kirby2026registers} & \shortstack[r]{+2.22\\{\fontsize{7.5}{8}\selectfont[1.84,2.65]}} & \shortstack[r]{+1.99\\{\fontsize{7.5}{8}\selectfont[1.60,2.42]}} & \shortstack[r]{+0.458\\{\fontsize{7.5}{8}\selectfont[0.288,0.624]}}\\
\bottomrule
\end{tabularx}
\end{table}
\begin{table}[!htb]
\vspace*{5pt}
\caption{\looseness=-1 \textbf{Frozen warmup predictions.} Eight equally weighted pairs of original Navhard scenes and their linked continuations. DR/CF denote DrivoR Navsim-v2/ChainFlow.}
\label{tab:warmup}
\centering\fontsize{8}{9.5}\selectfont
\setlength{\tabcolsep}{2pt}\renewcommand{\arraystretch}{1.08}
\begin{tabularx}{\columnwidth}{@{}l*{4}{>{\centering\arraybackslash}X}@{}}
\toprule
\rowcolor{TableHeader}
& Native & ToAD~\cite{xu2026toad} & DriveTTO-S & DriveTTO\\
\midrule
Models & DR & DR & DR & CF+DR\\
\rowcolor{TableAlt}
EPDMS $\uparrow$ & \textbf{61.61} & 56.53 & 60.33 & 57.89\\
$\Delta$ Native & 0.00 & $-5.08$ & $-1.27$ & $-3.72$\\
\bottomrule
\end{tabularx}
\end{table}
\begin{table}[!t]
\vspace*{5pt}
\caption{\textbf{Navtest recovery and regressions.} Counts average three seeds. $R/A/Z$ denote recovered/new/final zeros, with $Z=198-R+A$. Low averages PDMS on 530 Native-$\leq80$ scenes. Drops count Native-relative, seed-mean PDMS losses $>20$.}
\label{tab:failure_recovery}
\centering\fontsize{8.5}{10}\selectfont
\setlength{\tabcolsep}{2.5pt}
\renewcommand{\arraystretch}{1.08}
\begin{tabularx}{\columnwidth}{@{}>{\raggedright\arraybackslash}Xrrrrrr@{}}
\toprule
\rowcolor{TableHeader}
Method & PDMS $\uparrow$ & $R\uparrow$ & $A\downarrow$ & $Z\downarrow$ & Low $\uparrow$ & Drops $\downarrow$\\
\midrule
Native & 94.88 & 0.00 & 0.00 & 198.00 & 37.59 & 0\\
CEM8 & 95.10 & 16.67 & 1.33 & 182.67 & 41.75 & 11\\
\rowcolor{TableOurs}
\textbf{Full} & \textbf{95.51} & \textbf{101.67} & 41.00 & \textbf{137.33} & \textbf{61.56} & 118\\
Early & 95.39 & 51.67 & 16.33 & 162.67 & 50.97 & 52\\
Terminal & 95.06 & 10.67 & 3.33 & 190.67 & 40.99 & 21\\
\bottomrule
\end{tabularx}
\end{table}

\subsection{Acceptance and remaining regressions}
\label{sec:reference_history}
\paragraph{Reference retention.}
\looseness=-1
Retaining the current plan can preserve an earlier error. Full loses over 20 PDMS points relative to Native in 118 scenes after seed averaging (Appendix~\ref{app:reference_chain}). Against ToAD, DriveTTO-S avoids a collision and a road departure in two Navtest examples (Figure~\ref{fig:qualitative_main}). Figure~\ref{fig:navtest_toad_tto_cases} includes losses against ToAD.

\par

\paragraph{Verification timing.}
\label{sec:early_verification}
\looseness=-1
Checking the initial blend (Early) reduces large losses from 118 to 52 scenes, while recovery of native zeros falls from 51\% to 26\%. Checking only the final output (Terminal) leaves 21 large losses and recovers still fewer native zeros (Table~\ref{tab:failure_recovery}, Appendix~\ref{app:early_verification}).
\par

\setcounter{dbltopnumber}{2}
\raggedbottom
\section{Discussion}
\looseness=-1
Across six planners, expanding the candidate pool raises the best available Navtest score, yet a single scorer selects worse plans on average. A second scorer improves ranking and screening on fixed pools. In the full pipeline, changing the ranking rule at four stages has little effect on mean PDMS.

\newpage
\paragraph{Limitations.}
\looseness=-1
Both scorers can favor a harmful update. Configuration selection used evaluation scenes, so gains need confirmation on independent recordings. Full takes a median of 1.31\,s for encoding and search on Navtest. HUGSIM waits for each plan, leaving performance under real-time deadlines untested.

\section{Conclusion}
\looseness=-1
DriveTTO extends a frozen planner's proposal pool at inference time. Scores from completed paths guide local endpoint search, and two frozen scorers evaluate proposal, control, and fused trajectories against a retained plan. On Navtest, DriveTTO recovers 51.35\% of native zero-score cases on average and lowers the overall zero-score rate from 1.63\% to 1.13\%. Its single-scorer variant improves all six native planners tested on both Navtest and Navhard. In closed-loop HUGSIM, the observed-command configuration achieves an HD-Score of 40.1. These results support pairing local search with an explicit decision about when to replace the current plan.
\par
\nocite{drivinggpt2024}
{\linespread{1}\fontsize{9}{9.55}\selectfont
\setlength{\bibsep}{0pt}
\bibliographystyle{ieeenat_fullname}\bibliography{references}}
\flushcolsend
\clearpage
\flushbottom
\appendix
\setcounter{figure}{0}
\renewcommand{\thefigure}{S\arabic{figure}}
\renewcommand{\theHfigure}{supp.\arabic{figure}}
\setcounter{table}{0}
\setcounter{equation}{0}
\renewcommand{\thetable}{S\arabic{table}}
\renewcommand{\theequation}{S\arabic{equation}}
\renewcommand{\theHtable}{supp.\arabic{table}}
\renewcommand{\theHequation}{supp.\arabic{equation}}
\maketitlesupplementary
\newcommand{\suppTableStyle}{\fontsize{8.5}{10}\selectfont\setlength{\tabcolsep}{4pt}\renewcommand{\arraystretch}{1.12}}

\noindent\textbf{Reading guide.}
Appendices~\ref{app:datasets}--\ref{app:derivation} specify evaluation, controls, and operators. Appendix~\ref{app:qualitative} collects qualitative results. Blue highlights studied configurations and gray separates groups.
\begingroup
\linespread{.975}\selectfont
\setlength{\parskip}{1pt plus .6pt minus .5pt}
\setlength{\abovedisplayskip}{5pt plus 2pt minus 1pt}
\setlength{\belowdisplayskip}{5pt plus 2pt minus 1pt}
\setlength{\textfloatsep}{10pt plus 2pt minus 2pt}
\setlength{\floatsep}{8pt plus 2pt minus 1pt}
\setlength{\intextsep}{10pt plus 2pt minus 2pt}
\section{Evaluation protocols}
\label{app:datasets}
\subsection{Observations and benchmark metrics}
Navtest has 12,146 scenes~\cite{dauner2024navsim}. Navhard has 450 original and 5,462 synthetic observations~\cite{cao2025pseudo}. Inputs comprise current images, three preceding frames, observed ego motion, navigation commands, and identifiers for sampling/features. Agents output eight ego-frame poses at 0.5\,s intervals. Future states and evaluator caches are used only offline. 

\looseness=-1
PDMS multiplies no-at-fault collision (NC) and drivable-area compliance (DAC) by a weighted mean of progress (EP), time-to-collision (TTC), and comfort, with weights 5/5/2. EPDMS additionally penalizes driving-direction and traffic-light violations, filters violations shared by the human driver, and extends the averaged terms to lane keeping, history comfort, and extended comfort. Extended comfort differs from the ego-only tracking proxy used to construct candidates.

\looseness=-1
Navhard multiplies first-stage EPDMS by a weighted mean of continuation scores. Weights decay with squared distance from the reached endpoint to each linked start, using Gaussian variance $.1\,\mathrm m^2$. Products are computed before averaging. Pairing is unavailable to standard agents. Warmup contains 16 original and 204 synthetic observations, averaged within eight equally weighted paired groups from seven logs.

\subsection{Closed-loop protocol}
\label{app:hugsim_protocol}
HUGSIM~\cite{zhou2025hugsim,gigapixel2026} has 345 scenarios, with 79/87/51/128 from KITTI-360/nuScenes/PandaSet/Waymo. HD-Score multiplies maximum route completion (capped at one) by time-averaged CL-PDMS. CL-PDMS multiplies NC and DAC by the TTC/comfort weighted mean with weights 5/2. Reported RC, HD-Score, and CL-PDMS receive equal episode weights. Runs use seed 2, identical calibration, a 400-step cap, 0.25\,s simulation steps, and a 50\,ms iLQR limit. Simulation waits for planning. Inputs use rendered images and history at $-1.5,-1,-.5,0$\,s, initially left-padded. First-frame predictions agree within $10^{-6}$ across native/DriveTTO PyTorch 2.8.0/2.5.1. Time-limited iLQR remains variable. End-to-end latency and missed deadlines are unmeasured.

\subsection{Selection exposure and statistical inference}
\label{app:reproducibility}\label{app:bootstrap}
All scenes, including zeros, enter aggregates. Configuration selection used evaluation data, so reported intervals exclude selection uncertainty and do not provide independent-data confirmation. Unless stated otherwise, Navtest uses 2,000 scene-weighted bootstrap draws over 44 parent logs after averaging seeds 0/1/2. Earlier pool diagnostics use 136 recording segments, transfer uses 20,000 draws, and HUGSIM resamples 49 physical scene groups from single-seed rollouts. Sign tests assume exchangeable group-sum signs. Holm families comprise three system and two terminal-ranking contrasts. Compact noninferiority uses margin $-.20$. Navhard controls follow Section~\ref{app:refinement_controls}. Scores are reported on a 0--100 scale and differences are points, calculated before rounding.

\paragraph{Published comparisons.}\label{app:comparison_sources}
Published rows retain their original assets and budgets. Navtest EPDMS uses corrected-evaluator records, ChainFlow uses trainval PDMS, and Auto-JEPA follows its Navsim-v1 table. Published Navhard DrivoR/ToAD use 134k SimScale examples, DriveFuture uses V2-99, and DriveZero's 51.5 row is unscaled~\cite{kirby2026registers,xu2026toad,drivefuture2026,drivezero2026}. Common inference inputs do not imply matched training resources.
\section{Implementation and controlled experiments}
\label{app:parameters}\label{app:implementation}\label{app:ablations}
\subsection{Frozen models and benchmark schedules}
\label{app:schedules}\label{app:hugsim_adapter}
PDMS uses ChainFlow DiffGRPO (103k trainval, epoch 6, step 11,291) and DrivoR Navsim-v1. EPDMS uses 30-epoch DrivoR Navsim-v2, trained on 85k scenes plus 134k SimScale examples. Table~\ref{tab:configuration_map} specifies setting-dependent operators.
\begin{table}[!tbp]
\caption{\textbf{Benchmark configurations.} All weights are frozen. Shared operators use setting-specific contexts, budgets, and margins.}
\label{tab:configuration_map}
\centering\suppTableStyle\setlength{\tabcolsep}{2pt}
\begin{tabularx}{\linewidth}{@{}>{\raggedright\arraybackslash}p{.20\linewidth}>{\raggedright\arraybackslash}p{.25\linewidth}>{\raggedright\arraybackslash}X@{}}
\toprule
\rowcolor{TableHeader}
\textbf{Setting} & \textbf{Frozen scoring models} & \textbf{Candidate construction and available inputs}\\
\midrule
Navtest PDMS & ChainFlow~\cite{wang2026chainflow} + DrivoR~\cite{kirby2026registers} Navsim-v1 & Endpoint completion, shared geometric center, parallel proposal/control refinement, and temporal fusion.\\
\rowcolor{TableAlt}
Navtest EPDMS & ChainFlow~\cite{wang2026chainflow} + DrivoR~\cite{kirby2026registers} Navsim-v2 & Endpoint and geometric center followed by terminal-preserving tracking corrections.\\
Navhard standard & ChainFlow~\cite{wang2026chainflow} + DrivoR~\cite{kirby2026registers} Navsim-v2 & Navsim-v2 shooting anchor and sequential refinement with terminal-XY-constrained residuals.\\
\mbox{DriveTTO-S} & DrivoR~\cite{kirby2026registers} Navsim-v1 or Navsim-v2 & Target native plan and top-64 bank; portable endpoint, geometric, proposal/control and fusion operators.\\
\rowcolor{TableAlt}
HUGSIM & ChainFlow~\cite{wang2026chainflow} + DrivoR~\cite{kirby2026registers} Navsim-v1 & Online reference refinement with fixed-straight or observed commands; the latter supplies the main result.\\
\bottomrule
\end{tabularx}

\end{table}
\looseness=-1
Endpoint search uses $2\times128$ requests. Geometry uses two rounds of 24 samples/four elites, control one round of 32/four. Navsim-v2 shooting uses five rounds of 64/eight. Navhard adds 16 fixed-XY residuals, with seed 2 and token-ordered batches of 32. EPDMS tests eight tracking corrections. HUGSIM uses three 32-sample control rounds, .75/.25 DrivoR/ChainFlow gain weights, and dual-positive acceptance. Observed commands change navigation inputs only.
\subsection{System, ranking, and failure controls}
\label{app:confirmation}\label{app:confirmation_inference}
Native returns ChainFlow. Deterministic union and interpolation retain Native and compare both banks or five native-output blends using dual-positive admission and minimum-gain ranking. CEM4/8/16 repeat the control sampling and objective in Section~\ref{sec:control_refinement}, with equal old/elite updates and a .20 scale floor. Branch ablations bypass one generator and rerun all downstream stages. Endpoint removal also bypasses native-path initialization. Compact retains initial mixing, proposals, and constant fusion, replacing the control branch with the shared center.

\paragraph{Inference and cost.}\label{app:branch_intervals}
Full seed means are 95.54/95.49/95.51. The three planned tests use Section~\ref{app:bootstrap}'s protocol. Full exceeds union and CEM8 (adjusted $p=.0105$ and $.0360$, respectively). Compact minus Full is $-.12$ [$-.24,-.0022$], with one-sided lower bound $-.25$ and adjusted noninferiority $p=.0955$. The endpoint-removal intervals in main Figure~\ref{fig:paired_quality_differences} are descriptive, outside this family. Timing uses 64 predefined scenes, two isolated repeats, one process per GPU, and includes both encoders through final trajectory, excluding loading, input construction, and simulation. CEM8 selection used timing only.

\paragraph{Terminal ranking.}\label{app:terminal_ranking}
Mean/DrivoR ranking replaces minimum ranking at proposal, control-proposal, control-arm, and fusion selection, preserving dual-positive admission and minimum-based elite updates. Minimum-minus-mean/DrivoR is $.004$ [$-.03,.04$]/$.0055$ [$-.02,.03$], both adjusted $p=1$. Under minimum ranking with zero-rank fallback, extra gating changes only 22 control-arm choices because admission and ranking use different contexts.

\paragraph{Individual operator controls.}
\label{app:implementation_controls}\label{app:same_branch_controls}\label{app:feedback_coordinates}\label{app:residual_placement}
To distinguish operator properties from measured improvements, we also hold the surrounding computation fixed. Temporal rather than constant fusion changes PDMS by $-.0015$ [$-.02,.01$] on identical branch outputs. Recording utility at requested endpoints changes the result in one scene, with a mean PDMS difference of $.0000248$ [0,.0000854]. Transport--reconstruction intervals at scales .25/.50/1.00 include zero. These exploratory tests use 2,000 scene-weighted bootstrap draws over 44 parent logs on development-exposed scenes. They do not establish a resolved mean gain from each operator separately.

\paragraph{Failure accounting.}\label{app:failure_recovery}
Zeros use unrounded $J_{i,s}=0$ and coincide at tolerance $10^{-12}$. Count each seed before averaging. With 198 Native zeros, $Z_s=198-R_s+A_s$. All zeros have NC or DAC zero, whereas TTC zero can retain positive PDMS. Full has 134/141/137 zeros versus CEM8's 182/183/183. Full--Native/Full--CEM8 zero-rate differences are $-.50$ [$-.78,-.25$]/$-.37$ [$-.64,-.13$] percentage points, with 2,000 parent-log draws and no multiplicity adjustment. Table~\ref{tab:failure_recovery_additional} gives additional controls.
\begin{table}[t]
\caption{\textbf{Additional recovery comparisons.} Procedures complement main Table~\ref{tab:failure_recovery}, with the same per-run count definitions and $Z=198-R+A$.}
\label{tab:failure_recovery_additional}
\centering\suppTableStyle
\setlength{\tabcolsep}{3.5pt}
\begin{tabularx}{\columnwidth}{@{}>{\hsize=1.6\hsize\linewidth=\hsize\raggedright\arraybackslash}X>{\hsize=0.85\hsize\linewidth=\hsize\raggedleft\arraybackslash}X>{\hsize=0.85\hsize\linewidth=\hsize\raggedleft\arraybackslash}X>{\hsize=0.85\hsize\linewidth=\hsize\raggedleft\arraybackslash}X>{\hsize=0.85\hsize\linewidth=\hsize\raggedleft\arraybackslash}X@{}}
\toprule
\rowcolor{TableHeader}
\textbf{Method} & \textbf{PDMS} & $Z\downarrow$ & $R\uparrow$ & $A\downarrow$\\
\midrule
Native union & 95.00 & 195.00 & 14.00 & 11.00\\
\rowcolor{TableAlt}
Fixed interpolation & 95.04 & 190.00 & 13.00 & 5.00\\
CEM4 & 95.10 & 183.33 & 16.33 & 1.67\\
\rowcolor{TableAlt}
CEM16 & 95.13 & 181.67 & 17.67 & 1.33\\
\rowcolor{TableOurs}
\textbf{Joint compact} & 95.39 & 153.00 & 77.00 & 32.00\\
\bottomrule
\end{tabularx}
\end{table}
Figure~\ref{fig:failure_recovery_distribution} uses common Native groups $\{0\}$, $(0,50]$, $(50,80]$, $(80,95]$, $(95,100]$, containing 198/15/317/2,343/9,273 scenes. The combined 530-scene low-score gain is 23.98 [19.27,29.51]. Among Native zeros, 67/66/70 Full outputs reach 80 and 48/48/51 reach 95. Main Table~\ref{tab:failure_recovery} includes newly introduced failures.
\begin{figure}[t]
\centering
\includegraphics[width=\columnwidth]{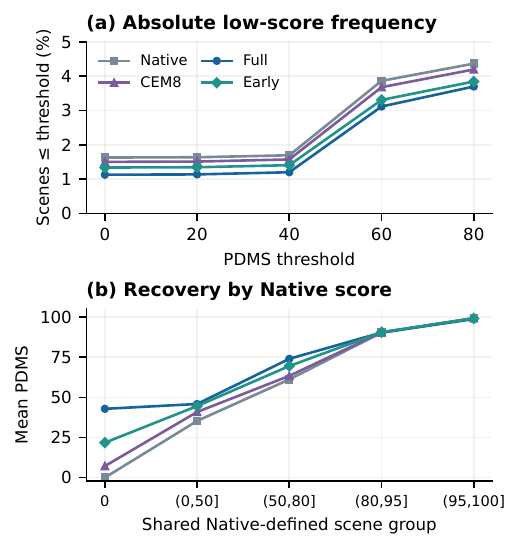}
\caption{\textbf{Recovery beyond exact zeros.} The top panel gives all-scene frequencies at five absolute PDMS thresholds. The bottom panel reports mean scores within common Native-defined groups, with sizes given in the text. Events are counted per seed before averaging counts. Scores also average seeds.}
\label{fig:failure_recovery_distribution}
\end{figure}

\paragraph{Large losses and reference history.}
\label{app:confirmation_tails}\label{app:reference_chain}
Table~\ref{tab:confirmation_tail} reports Native-relative drops after seed averaging. Full--CEM8 $>20$-point frequency differs by .88 percentage points [.64,1.16]. On the 118-scene Full loss set, Native/Full/CEM8/Compact score 92.47/40.52/90.56/57.35. These final-output comparisons quantify regressions without attributing each loss to a particular stage.
\begin{table}[t]
\caption{\textbf{Native-relative PDMS drops.} Counts threshold three-seed scene means. The last column is the percentage of 12,146 scenes.}
\label{tab:confirmation_tail}
\centering\suppTableStyle
\setlength{\tabcolsep}{3pt}
\begin{tabularx}{\columnwidth}{@{}>{\hsize=1.7\hsize\linewidth=\hsize\raggedright\arraybackslash}X>{\hsize=0.825\hsize\linewidth=\hsize\raggedleft\arraybackslash}X>{\hsize=0.825\hsize\linewidth=\hsize\raggedleft\arraybackslash}X>{\hsize=0.825\hsize\linewidth=\hsize\raggedleft\arraybackslash}X>{\hsize=0.825\hsize\linewidth=\hsize\raggedleft\arraybackslash}X@{}}
\toprule
\rowcolor{TableHeader}
\textbf{Method} & $>1$ pt & $>5$ pt & $>20$ pt & $>20$ (\%)\\
\midrule
Native union & 174 & 41 & 25 & .2058\\
\rowcolor{TableAlt}
CEM8 & 37 & 15 & 11 & .0906\\
Joint compact & 1,267 & 129 & 82 & .6751\\
\rowcolor{TableOurs}
\textbf{Full} & 1,555 & 224 & 118 & .9715\\
\bottomrule
\end{tabularx}

\end{table}
Local proposal/control/fusion decisions lower reference quality in 23 of the 118 scenes. Geometry removal returns 40 Full-loss scenes within the threshold and introduces 11, reducing frequency by .24 percentage points [.11,.41].

\paragraph{Early versus terminal verification.}\label{app:early_verification}
Early compares $n$ and $b_0$ in slots 0/1 of each native 64-proposal bank, preserving the other 62 entries. If either gain is nonpositive, Early retains $n$. Downstream budgets and random streams are unchanged, although resetting the reference changes generated candidates. Terminal applies the same gate to saved Full versus Native without rerunning search. Rejecting a recovered output can therefore restore a native failure. The check and analysis were specified before scoring. Early meets only the loss-frequency target. It brings 70 of Full's 118 large-loss scenes back within the threshold and adds four new large losses, leaving 52. Mean quality changes by $-.12$. Its isolated latency is unmeasured.
\subsection{Candidate access, ranking, and screening}
\label{app:opportunity_results}\label{app:opportunity_protocol}
These controls ask whether candidate access or selection limits output quality. The oracle $O$ is the highest official PDMS in a saved pool, and regret is the gap between this score and the selected output. Official scores are used only for offline analysis. Six planners share frozen ChainFlow/DrivoR scorers. Pools add 960 candidates and an exact incumbent to top-64 proposals. Four directions and three amplitudes contribute 32 random and 48 adaptive candidates each. Adaptation uses three 16-sample rounds fitted to DrivoR gains. Endpoint, interior-geometry, timing, and segmented-control perturbations are diagnostic families, distinct from deployed operators.

\begin{table}[!tbp]
\captionsetup{type=table}
\caption{\textbf{Expanded-pool uncertainty.} Regret is oracle minus selected PDMS. Recording-cluster 95\% intervals accompany main Table~\ref{tab:opportunity}. Both policies accept the displayed count.}
\label{tab:opportunity_full}
\centering\suppTableStyle\setlength{\tabcolsep}{1.5pt}
\begin{tabularx}{\linewidth}{@{}>{\hsize=1\hsize\linewidth=\hsize\raggedright\arraybackslash}X>{\hsize=0.95\hsize\linewidth=\hsize\raggedleft\arraybackslash}X>{\hsize=1.15\hsize\linewidth=\hsize\raggedleft\arraybackslash}X>{\hsize=0.8\hsize\linewidth=\hsize\raggedleft\arraybackslash}X>{\hsize=1.1\hsize\linewidth=\hsize\raggedleft\arraybackslash}X@{}}
\toprule
\rowcolor{TableHeader}
 & \multicolumn{2}{c}{\textbf{Pool expansion}} & \multicolumn{2}{c}{\textbf{Joint selection}}\\
\cmidrule(lr){2-3}\cmidrule(l){4-5}
\textbf{Planner} & \shortstack{Regret\\increase} & \shortstack{Output change\\95\% CI} & \shortstack{Accepted\\scenes} & \cellcolor{TableOursHead}\shortstack{Joint gain\\95\% CI}\\
\midrule
ZTRS & +2.96 & [-1.87,-1.13] & 12,143 & [+1.46,+2.10]\\
GTRS & +2.00 & [-1.84,-1.20] & 12,132 & [+1.14,+1.79]\\
Hydra-MDP & +2.24 & [-2.12,-1.42] & 12,127 & [+1.53,+2.23]\\
iPad & +2.24 & [-2.39,-1.45] & 12,124 & [+1.85,+2.83]\\
RAP-DINO & +1.93 & [-1.91,-1.10] & 12,118 & [+1.80,+2.72]\\
DrivoR & +2.11 & [-2.33,-1.46] & 12,038 & [+1.79,+2.57]\\
\bottomrule\end{tabularx}

\end{table}
\begin{table}[!tbp]
\captionsetup{type=table}
\caption{\textbf{Full-bank versus accessible oracle.} Access loss is the full-bank oracle minus the top-64 oracle. $\Delta O$ is the oracle gain from adding generated candidates to the full bank.}
\label{tab:native_access}
\centering\suppTableStyle
\begin{tabularx}{\columnwidth}{@{}>{\hsize=1.45\hsize\linewidth=\hsize\raggedright\arraybackslash}X>{\hsize=0.8\hsize\linewidth=\hsize\raggedleft\arraybackslash}X>{\hsize=0.9\hsize\linewidth=\hsize\raggedleft\arraybackslash}X>{\hsize=1\hsize\linewidth=\hsize\raggedleft\arraybackslash}X>{\hsize=0.85\hsize\linewidth=\hsize\raggedleft\arraybackslash}X@{}}
\toprule
\rowcolor{TableHeader}
\textbf{Planner} & Full $O$ & Top-64 $O$ & \textbf{Access loss} & $\Delta O$\\
\midrule

ZTRS~\cite{ztrs2025} & 99.88 & 93.87 & 6.02 & $<0.01$\\

\rowcolor{TableAlt}
GTRS~\cite{gtrs2025} & 99.88 & 96.42 & 3.46 & $<0.01$\\

Hydra-MDP~\cite{li2024hydramdp} & 99.88 & 96.67 & 3.21 & $<0.01$\\

\rowcolor{TableAlt}
iPad~\cite{ipad2025} & 98.57 & 98.57 & 0.00 & +0.36\\

RAP-DINO~\cite{rap2025} & 98.31 & 98.31 & 0.00 & +0.45\\

\rowcolor{TableAlt}
DrivoR~\cite{kirby2026registers} & 99.45 & 99.45 & 0.00 & +0.22\\

\bottomrule
\end{tabularx}

\end{table}

Shared contexts alternate DrivoR/ChainFlow proposals. Single selection ranks DrivoR gain. Joint selection requires positive DrivoR and nonnegative ChainFlow gains, ranking their minimum. Zero-ranked incumbents win ties. Matched coverage fixes update counts, not scene identities. Recording-cluster intervals and six-planner Holm tests use 2,000 draws over 136 segments, with $p\approx.003$ per contrast (Table~\ref{tab:opportunity_full}).

ZTRS/GTRS/Hydra share a 16,384-entry vocabulary, whereas other full banks have 64 entries. Table~\ref{tab:native_access} separates access loss from additional opportunity. Expanded joint selection versus joint top-64 changes PDMS by +.36/$-.17$/+.04/+.08/+.33/+.31 in row order, so expansion does not improve every planner.

\paragraph{Fixed-pool interventions.}
\label{app:selection_interventions}\label{app:common_mask}
The ranking and screening interventions follow Table~\ref{tab:ranking_screening}. Primary tests use 136 segments, with Holm-adjusted $p=.006$ throughout. Common-mask and sensitivity tests use 44 parent logs, retaining positive primary intervals. After minimum ranking, extra screening changes means by less than $1.2\times10^{-5}$ point. Fixed-pair native/shared contexts change gain signs in 4.65--5.30\% of DrivoR and 7.17--9.38\% of ChainFlow comparisons.
\subsection{Transfer and Navhard controls}
\label{app:endpoint_transfer}\label{app:transfer_protocol}\label{app:refinement_controls}
ToAD and DriveTTO-S share native anchors, top-64 banks, and DrivoR v1/v2 scorers on Navtest/Navhard. ToAD uses ten/five CEM rounds. Navtest uses 64 samples, eight elites, .70 smoothing, .50 position/control-anchor weights, and .05 comfort weight. DriveTTO-S uses endpoint $2\times128$, geometry $2\times24$, control $1\times32$, four elites, and interpolation caps .50/.75. Hydra uses GTRS-distributed weights and ZTRS its VoV release.
\begin{table}[!tbp]
\caption{\textbf{Single-scorer phase medians on H20.} Batch size one with reused features. Head calls are complete scorer invocations with varying candidate-batch sizes.}
\label{tab:transfer_phases}
\centering\suppTableStyle
\setlength{\tabcolsep}{1.5pt}
\begin{tabularx}{\linewidth}{@{}>{\raggedright\arraybackslash}p{.31\linewidth}*{5}{>{\centering\arraybackslash}X}@{}}
\toprule
\rowcolor{TableHeader}
\textbf{Phase} & \textbf{ZTRS} & \shortstack{\textbf{Hydra-}\\\textbf{MDP}} & \textbf{iPad} & \textbf{GTRS} & \shortstack{\textbf{RAP-}\\\textbf{DINO}}\\
 & \cite{ztrs2025} & \cite{li2024hydramdp} & \cite{ipad2025} & \cite{gtrs2025} & \cite{rap2025}\\
\midrule
\rowcolor{TableGroup}\multicolumn{6}{l}{\textbf{One-time preparation (ms)}}\\
Native + forward & 446.8 & 242.7 & 208.2 & 241.7 & 652.0\\
Scorer + encode & 281.3 & 285.6 & 300.3 & 285.0 & 275.9\\
\midrule
\rowcolor{TableGroup}\multicolumn{6}{l}{\textbf{Scoring and search (ms)}}\\
Rerank & 3.3 & 3.5 & 3.8 & 3.6 & 3.1\\
ToAD~\cite{xu2026toad} & 42.9 & 45.5 & 48.8 & 43.4 & 40.3\\
\rowcolor{TableOurs}DriveTTO-S & 161.9 & 170.7 & 185.2 & 165.2 & 153.9\\
\midrule
\shortstack[l]{Head calls\\(DriveTTO-S)} & 42 & 42 & 41 & 42 & 41\\
\bottomrule
\end{tabularx}

\label{tab:transfer_cost}
\end{table}
\paragraph{Smoothing and reranking controls.}
Smoothing reconstructs and rolls out bounded bicycle-model controls without search. Reranking selects unchanged top-64 proposals. New Navhard runs use seed 2, token-derived sampling, and batch size one, reproducing all 5,912 native trajectories. DriveTTO-S--rerank is +2.41 [.21,4.54] for Hydra and +1.66 [$-.10,3.45$] for DrivoR. Against the new five-round ToAD stream, differences are $-1.82$ [$-4.62,1.01$]/$-.03$ [$-1.73,1.69$], from 10,000 draws over 225 paired groups. iPad controls remain unavailable.

\paragraph{Navtest transfer.}\label{app:transfer_effects}\label{app:transfer_timing}
For ZTRS/GTRS/Hydra/iPad/RAP-DINO, top-64 reranking scores 90.56/93.24/93.27/93.68/94.08. DriveTTO-S adds .46/.28/.19/.55/.49. Transfer intervals use 20,000 recording-cluster draws on one inference seed. Table~\ref{tab:transfer_phases} reports CUDA-synchronized H20 phase medians with cached features, excluding loading and pool preparation. Separate phase medians are not end-to-end latency.

\paragraph{Warmup scope.}
\label{app:warmup}
Warmup reuses eight development-exposed groups and earlier ToAD predictions (55.64 full-Navhard EPDMS). Component means differ from means of products with method-dependent continuation weights. This subset is not independent confirmation.
\subsection{HUGSIM adapter and input ablations}
\label{app:hugsim_results}
Table~\ref{tab:hugsim_variants} compares reproduced DrivoR and base DriveTTO under fixed-straight commands. Base DriveTTO adds ChainFlow and refinement. The last three rows independently modify the base.
\begin{table}[!tbp]
\caption{\textbf{HUGSIM adapter controls.} Equal-episode means on 345 scenarios. Final three rows independently change base DriveTTO. Metrics follow Appendix~\ref{app:hugsim_protocol}.}
\label{tab:hugsim_variants}
\centering\suppTableStyle
\setlength{\tabcolsep}{.9pt}
\begin{tabularx}{\linewidth}{@{}>{\raggedright\arraybackslash}p{.24\linewidth}*{7}{>{\centering\arraybackslash}X}@{}}
\toprule
\rowcolor{TableGroup}
 & \multicolumn{2}{c}{\shortstack{\textbf{Overall}\\\textbf{performance}}} & \multicolumn{5}{c}{\shortstack{\textbf{Safety and}\\\textbf{motion quality}}}\\
\cmidrule(lr){2-3}\cmidrule(l){4-8}
\rowcolor{TableHeader}
\textbf{Configuration} & \textbf{RC} & \textbf{HD} & \textbf{NC} & \textbf{DAC} & \textbf{TTC} & \shortstack{\textbf{Com-}\\\textbf{fort}} & \shortstack{\textbf{CL-}\\\textbf{PDMS}}\\
\midrule
DrivoR~\cite{kirby2026registers} (reproduced) & 48.2 & 34.0 & 55.5 & 96.4 & 50.2 & 93.9 & 48.7\\
\rowcolor{TableOurs}
\textbf{DriveTTO (base)} & 52.5 & 38.1 & 57.9 & 96.5 & 52.4 & 93.1 & 50.1\\
+ execution-heading consensus & 51.6 & 37.3 & 57.5 & 96.6 & 52.2 & 93.3 & 49.8\\
\rowcolor{TableAlt}
+ progress-product ranking & 51.2 & 36.5 & 56.0 & 96.5 & 50.0 & 93.8 & 48.0\\
\rowcolor{TableOurs}
\textbf{+ observed navigation command} & \textbf{54.5} & \textbf{40.1} & \textbf{58.2} & \textbf{97.5} & \textbf{53.0} & 93.2 & \textbf{51.3}\\
\bottomrule
\end{tabularx}

\end{table}
\looseness=-1
Observed commands change inputs only. Their RC/HD gain intervals are [$-1.00,6.40$]/[$-.80,5.90$]. Of 22,880 decisions, 3,075 use non-straight commands. HD changes by +4.00 in 176 turn-metadata scenarios and $-.20$ in 169 straight-only scenarios. Observed-command DrivoR, equal-model union/CEM, and repeated suites remain unevaluated.
\section{Operator implementation}
\label{app:derivation}\label{app:executable_schedule}
The main algorithm defines the reference sequence. This section supplies parameters and properties omitted there. All banks preserve insertion order on ties. Interpolation uses $A=\{0,.25,.5,.75,1\}$ and circular headings. Native comparison slots 0/1 hold reference/candidate. Zero residuals and rejected updates copy the stored array.

\subsection{Endpoint completion and feedback}
\label{app:endpoint}
Endpoint completion preserves the intermediate geometry of a native path while changing its destination. Native-path initialization first groups paths by endpoint and motion compatibility. In each of two rounds, up to four groups propose 64 requests, retain two by packed ChainFlow utility, and attempt two parent completions each. Each of eight shortlisted paths is compared separately with Native. Acceptance requires gain $>.002$, bank-argmax status, and NC/DAC/TTC decreases at most $10^{-4}$.

Particle search stores completed paths, achieved terminal states, utilities, and parent weights. Packed scoring inserts up to eight completions into non-anchor slots, retaining native proposals elsewhere. Utility is $s_{\rm nat}+S_1(c;\mathcal B)-S_1(b_{\rm nat};\mathcal B)$, expressing the paired gain on Native's original score scale. Eight-neighbor distances sum normalized squared XY, wrapped-heading, and speed differences. XY bandwidth is median nearest-neighbor separation floored at .1\,m, with heading/speed scales .35\,rad and 3\,m/s. Sixteen completion slots allow three contraction checks each. Achieved-state utility subtracts .002 reach cost, .002 support cost, and .001 dynamics cost. Reach measures normalized terminal error, support measures parent deviations, and dynamics uses kinematic checks. Valid-parent fractions are recorded separately.

\looseness=-1
For supported seed $\bar p$, parent residual $r=p-\bar p$, $s=(t-1)/(T-1)$, and $h(s)=10s^3-15s^4+6s^5$, completion uses
\begin{equation}
p_t^{\rm new}=\bar p_t+r_t-(1-h(s))r_1-h(s)r_T.
\label{eq:app_completion}
\end{equation}
Both boundary residuals vanish. Heading and speed are recomputed before filtering and full-path scoring. The requested position sets the completion target. Requested heading and speed enter the distance and realization costs.

Three direct paths join optional raw and corrected decoder paths. Re-decoding mixes native autoregressive trajectories and recreated noise by parent weights, with variance/RMS normalization. The corrected path is $d\oplus(r\ominus\bar\tau)$ for best direct path $d$, decoded path $r$, and weighted native decoded mixture $\bar\tau$. Both decoder paths undergo finite-pose, endpoint, parent-support, and motion checks. Final verification places the candidate in first/last non-anchor native slots. Both contexts require gain $>.002$, bank-argmax status, and NC/DAC/TTC/DDC decreases at most $10^{-4}$. Minimum gain ranks eligible paths. Only the first two direct sources and the corrected decoder output may replace the incumbent. Other winning sources trigger exact fallback. Navhard instead fits its shooting Gaussian to elite achieved states and returns the final-mean rollout.

\subsection{Geometry and control}
\label{app:geometry}\label{app:proposal_pool}
Geometric search explores small changes in endpoint and timing around the retained trajectory. It extends the eight-pose path with ego origin and a final secant, then interpolates at index $(1+.01z_s)t$. Add $(t/8)^2\eta_b(.3z_\parallel e_\parallel+.1z_\perp e_\perp)$ metres, where terminal-heading axes define $e_\parallel,e_\perp$ and $\eta_b=\min(1,L_b/(2\,\mathrm m))$. Tangent heading changes clip to $\pm.05$\,rad only for segments longer than .1\,m. Two antithetic CEM rounds clip coefficients to $[-2,2]^3$ and update four-elite mean/population-standard-deviation with equal old/new weight and .2 floor. DrivoR ranks in $[b_0,c,f,\mathcal I_{.5}(f,d_2),d_2]$. The archive includes the final mean and deduplicates at $10^{-6}$. The winner needs positive native-context gain. All five arms are scored jointly, while only $\alpha\leq.75$ can replace $b_0$ after a further native-context check.

\paragraph{Residual rollout.}\label{app:transport}
Observed longitudinal speed initializes $v_0=\max(0,v_{\rm obs})$. At $\Delta t=.5$\,s, nominal segment speed is $\|p_t-p_{t-1}\|/\Delta t$, acceleration its difference from the previous speed, and yaw rate the wrapped heading difference divided by $\Delta t$. Deterministic rollout uses
\begin{equation}
\begin{aligned}
v_t&=\max(0,v_0+\Delta t\sum_{l=1}^t a_l),\quad
\theta_t=\Delta t\sum_{l=1}^t\omega_l,\\
p_t^R-p_{t-1}^R&=\tfrac{\Delta t}2(v_{t-1}+v_t)
\begin{bmatrix}\cos\bar\theta_t\\\sin\bar\theta_t\end{bmatrix},\quad
\bar\theta_t=\tfrac{\theta_{t-1}+\theta_t}2.
\end{aligned}
\label{eq:discrete_rollout}
\end{equation}
To bound the added residual, let $M_p,M_\psi$ be its maximum positional and wrapped-heading magnitudes. We use $\kappa=\min(1,.8/M_p,.06/M_\psi)$, interpreting zero denominators as infinity. Scaling therefore limits changes to .8\,m and .06\,rad up to floating-point rounding. Identical nominal and perturbed rollouts cancel at zero perturbation, and explicit copying preserves rejected references exactly. Neither property guarantees a better driving score.

Four acceleration/yaw-rate coefficient pairs interpolate at 0/1/2/3/4\,s from zero perturbation, scaled by $.1\,\mathrm{m/s^2}$, $.006\,\mathrm{rad/s}$, and $\eta_b$. Sampling, elite selection, and bounded-arm acceptance follow Section~\ref{sec:control_refinement}. Circular interpolation uses the complex-resultant angle, with implementation-defined antipodal cases.

Navhard terminal-XY residual $e(z)$ uses finite-difference Jacobian $J_e$ and damped steps
\begin{equation}
z\leftarrow z-J_e^\top(J_eJ_e^\top+\lambda I)^{-1}e(z).
\label{eq:app_projection}
\end{equation}
This solves a regularized linearized least-squares correction before bounded-step checks. Only converged paths are admitted, sub-tolerance endpoint residual is removed, and bounds apply relative to current and root references. Heading and speed remain free.

\subsection{Terminal-preserving EPDMS correction}
\label{app:epdms_tracking}
The EPDMS configuration corrects tracking mismatch while preserving the selected endpoint. Its ego-only LQR/bicycle tracker $\mathcal E$ uses observed velocity, acceleration, and history. Causal heading differences initialize angular velocity/acceleration clipped to $\pm1$\,rad/s and $\pm2$\,rad/s$^2$. Steering clips to $\pm.6$\,rad and is zero below .5\,m/s. Targets interpolate to .1\,s, outputs return to the eight planning times, and state resets per scene without traffic, maps, or evaluator caches.

For $r_t=\mathcal E(b)_t\ominus b_t$, use $\widetilde r_t=r_t-h(t/T)r_T$ with $h(s)=s^2(3-2s)$. Both residual signs at strengths .25/.50/.75/1 produce eight corrections bounded by .8\,m/.06\,rad. The final stored pose is copied from $b$, with duplicate removal. Proxy error averages XY discrepancy plus twice wrapped-heading discrepancy under re-tracking. Candidate--center pairs use both native and shared contexts, the latter alternating ChainFlow then DrivoR proposals. Replacement maximizes the minimum of four total gains, requiring gain at least .001, NC/DAC/TTC/DDC decreases at most $10^{-4}$, EP decreases at most .01, and proxy reduction at least 5\% with denominator floor $10^{-6}$. Otherwise the center is copied.

\paragraph{Configuration selection.}
\label{app:epdms_evidence}
Two inspected groups of 1,173/1,353 scenes selected the rule. Eligible rules improve archived Navsim-v1 EPDMS, preserve mean NC, and recover at least as many zeros as they introduce in each group, maximizing the smaller gain. Tracking affects 139 full-set scenes. Uncorrected full-set center quality is unavailable, so 92.30 describes the combined checkpoint and shortened schedule. Against archived v1 predictions, the gain interval is [1.10,1.90], with 2,936 improved and 4,754 regressed scenes, 178 recovered zeros, and 68 new zeros. NC/TTC rise while DAC/progress fall.
\flushcolsend
\clearpage
\endgroup
\begingroup
\linespread{1}\selectfont
\setlength{\parskip}{0pt}
\onecolumn
\section{Qualitative results}
\label{app:qualitative}
\label{app:dense_quality_landscapes}
In 16 Navtrain scenes, we complete eight-pose paths around the human endpoint and recompute headings, sampling at 0.50\,m, refining at 0.10\,m, and probing boundaries at 0.05\,m. Panels (a)--(d) show scores, paths, a local enlargement, and the marked slice. Blank cells are unsampled and dots mark off-grid probes. The turn and stop have the fewest grid locations scoring $\geq95$ (197 and 332, Figures~\ref{fig:dense_turn_boundary} and~\ref{fig:dense_stop_boundary}). Figures~\ref{fig:dense_road_boundary} and~\ref{fig:dense_collision_boundary} isolate road and collision boundaries. Figures~\ref{fig:qual_scenes} and~\ref{fig:qual_scenes_more} show 40 selected outputs (PDMS 100), and Figure~\ref{fig:qual_stages} traces five through refinement. Figure~\ref{fig:navtest_toad_tto_cases} compares ToAD and DriveTTO-S improvements and regressions. Appendix~\ref{app:confirmation_tails} reports loss statistics.

\makeatletter
\setlength{\@fptop}{0pt}
\setlength{\@dblfptop}{0pt}
\makeatother

\par\begingroup\centering\captionsetup{skip=3pt}\noindent
\begin{minipage}[t]{.48\textwidth}
\captionsetup{type=figure}\centering\vspace{0pt}
\includegraphics[width=.865\linewidth]{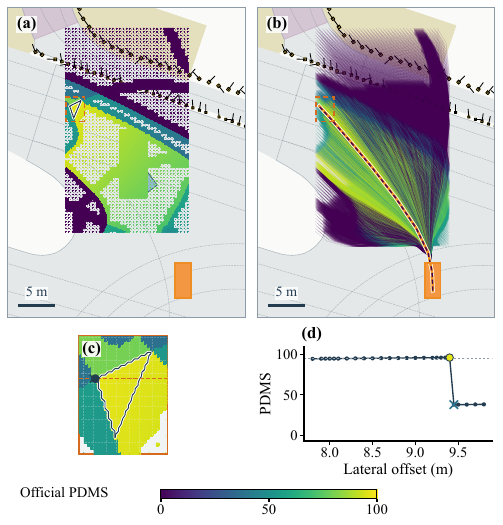}
\caption{\textbf{Turn boundary.} 24,648 paths, scene \texttt{295db25cd2a25fa3}. The marked $0.05$\,m pair scores 96.29/38.02. TTC and comfort change from one to zero, while NC and DAC remain one. The human reference scores 83.15.}
\label{fig:dense_turn_boundary}
\end{minipage}\hfill
\begin{minipage}[t]{.48\textwidth}
\captionsetup{type=figure}\centering\vspace{0pt}
\includegraphics[width=.865\linewidth]{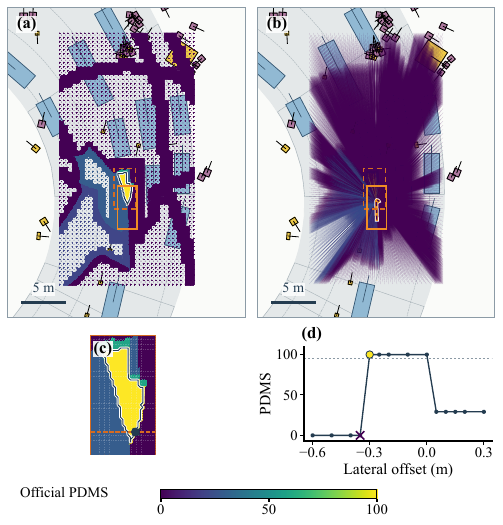}
\caption{\textbf{Narrow region near a stop.} 21,882 paths, scene \texttt{85af90d3198c5fb7}. At forward offset $-0.10$\,m, lateral requests $-0.30$/$-0.35$\,m score 100/0 as NC becomes zero. The high-score slice spans $0.30$\,m. The human reference advances 2.30\,m.}
\label{fig:dense_stop_boundary}
\end{minipage}\par\vspace{2pt}
\begin{minipage}[t]{.48\textwidth}
\captionsetup{type=figure}\centering\vspace{0pt}
\includegraphics[width=.865\linewidth]{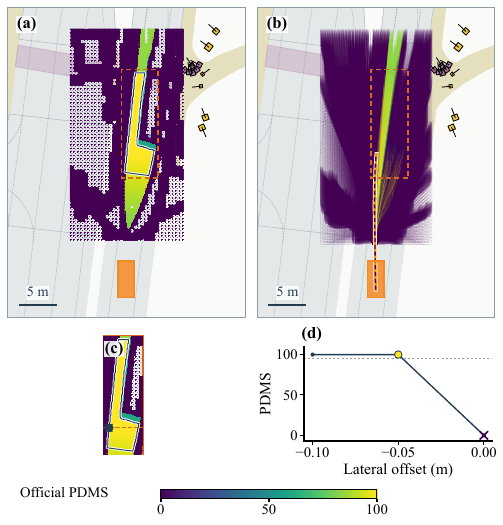}
\caption{\textbf{Drivable-area boundary.} 35,807 paths, scene \texttt{c13dbf32823b5383}. At forward offset $2.15$\,m, lateral requests $-0.05$/$0.00$\,m score 100/0. DAC changes from one to zero, while NC remains one. The human reference scores 95.94.}
\label{fig:dense_road_boundary}
\end{minipage}\hfill
\begin{minipage}[t]{.48\textwidth}
\captionsetup{type=figure}\centering\vspace{0pt}
\includegraphics[width=.865\linewidth]{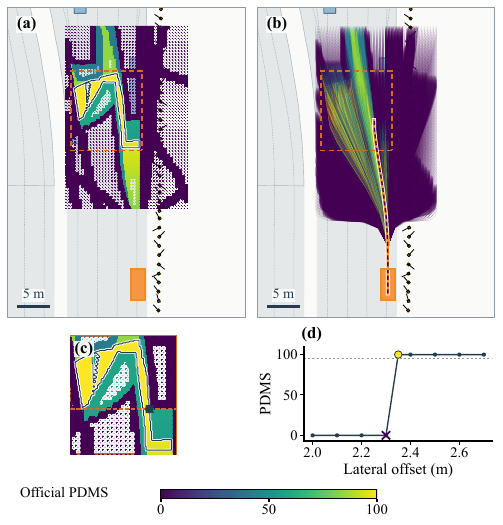}
\caption{\textbf{Collision-related boundary.} 40,857 paths, scene \texttt{ed1254e7bd00593b}. At forward offset $4.50$\,m, lateral requests $2.35$/$2.30$\,m score 100/0. NC becomes zero, while DAC remains one. The high-score band curves around a low-score interior. Human PDMS is 95.06.}
\label{fig:dense_collision_boundary}
\end{minipage}
\par\endgroup

\begin{figure*}[p]
\centering
\begin{minipage}[t]{.179\textwidth}\centering
{\fontsize{8}{9}\selectfont\bfseries Scene 1}\par\vspace{.5pt}
\includegraphics[width=\linewidth]{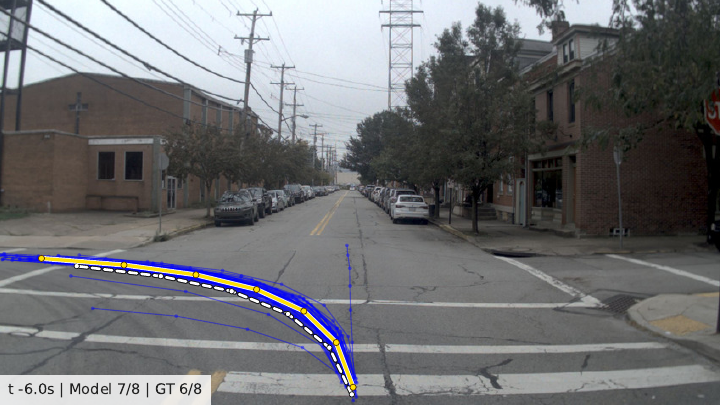}\par\vspace{.5pt}
\includegraphics[width=\linewidth]{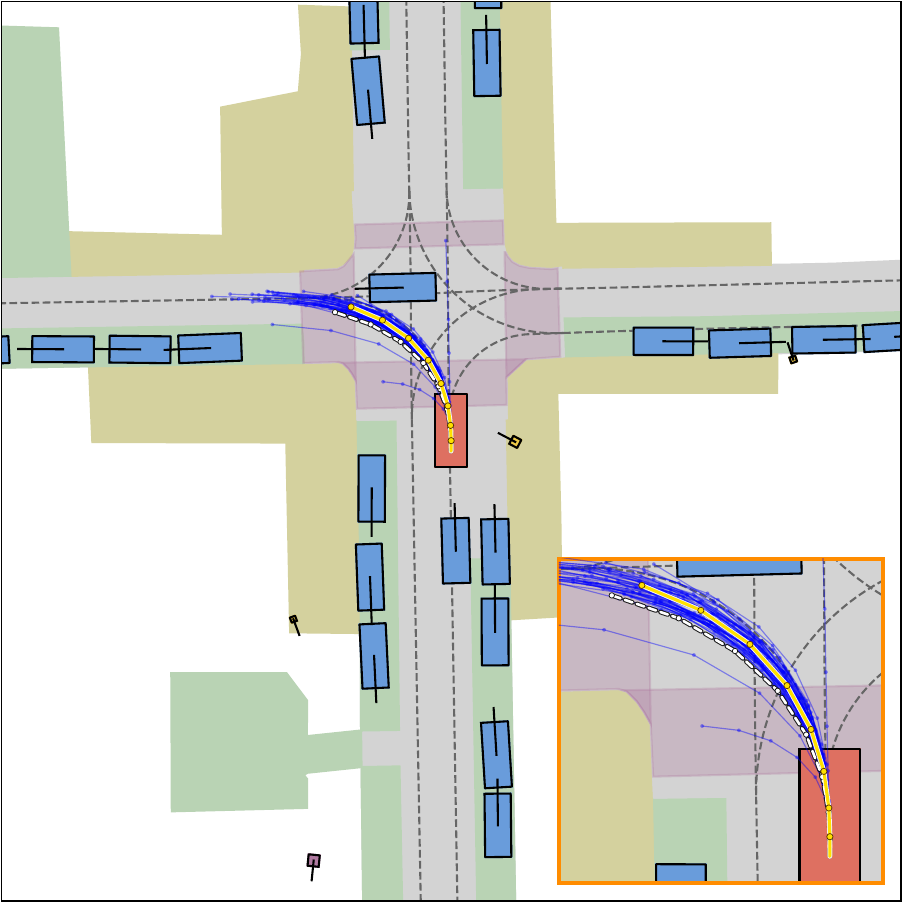}
\end{minipage}\hfill
\begin{minipage}[t]{.179\textwidth}\centering
{\fontsize{8}{9}\selectfont\bfseries Scene 2}\par\vspace{.5pt}
\includegraphics[width=\linewidth]{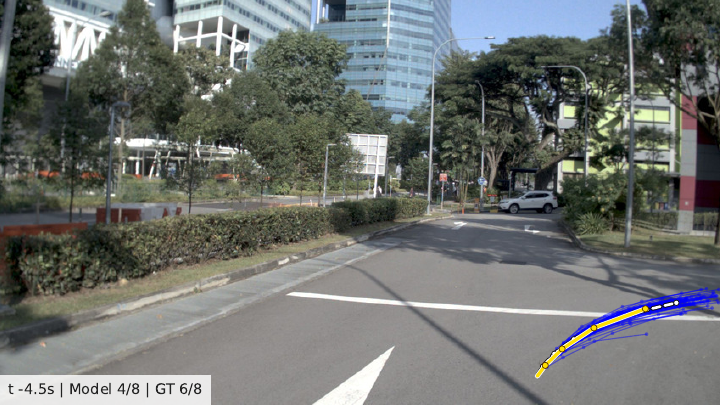}\par\vspace{.5pt}
\includegraphics[width=\linewidth]{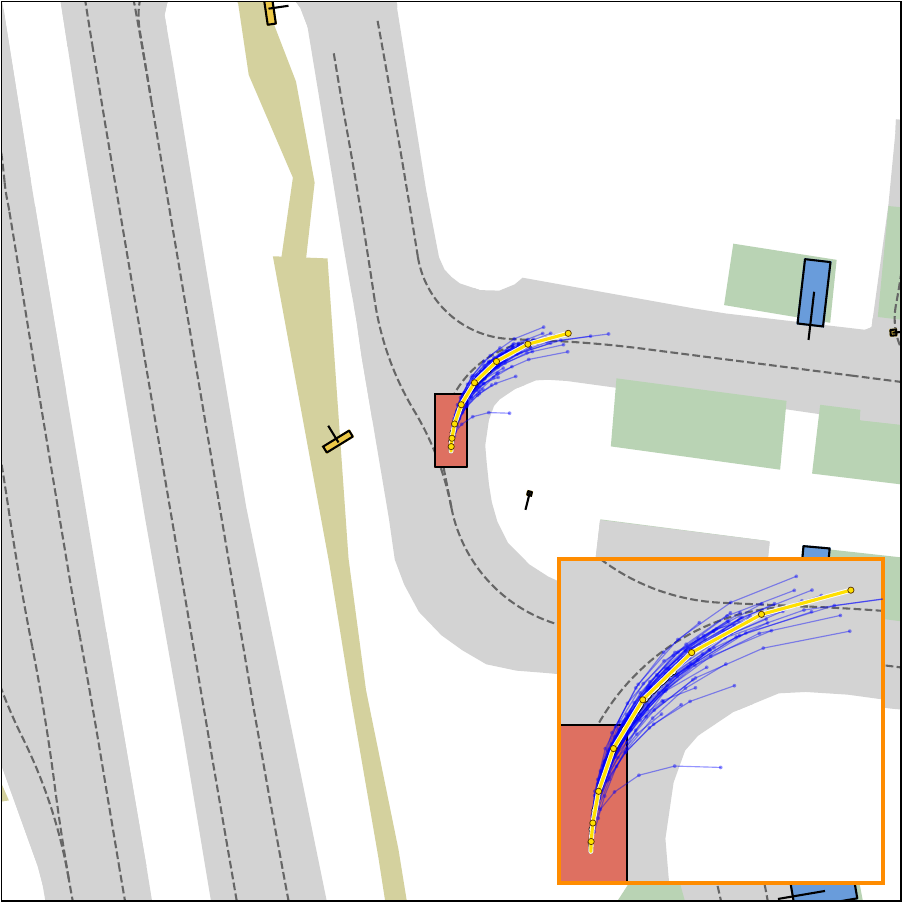}
\end{minipage}\hfill
\begin{minipage}[t]{.179\textwidth}\centering
{\fontsize{8}{9}\selectfont\bfseries Scene 3}\par\vspace{.5pt}
\includegraphics[width=\linewidth]{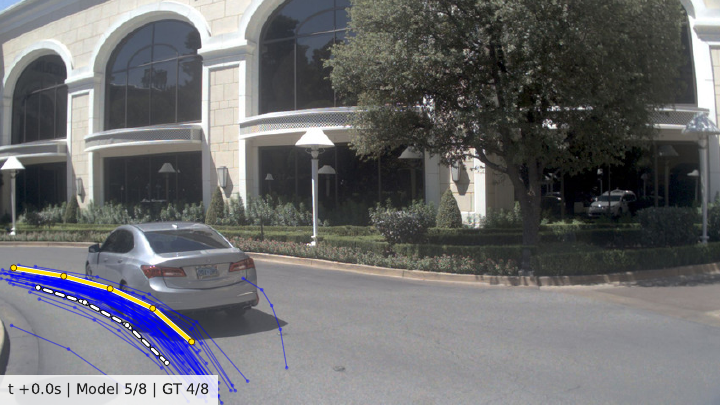}\par\vspace{.5pt}
\includegraphics[width=\linewidth]{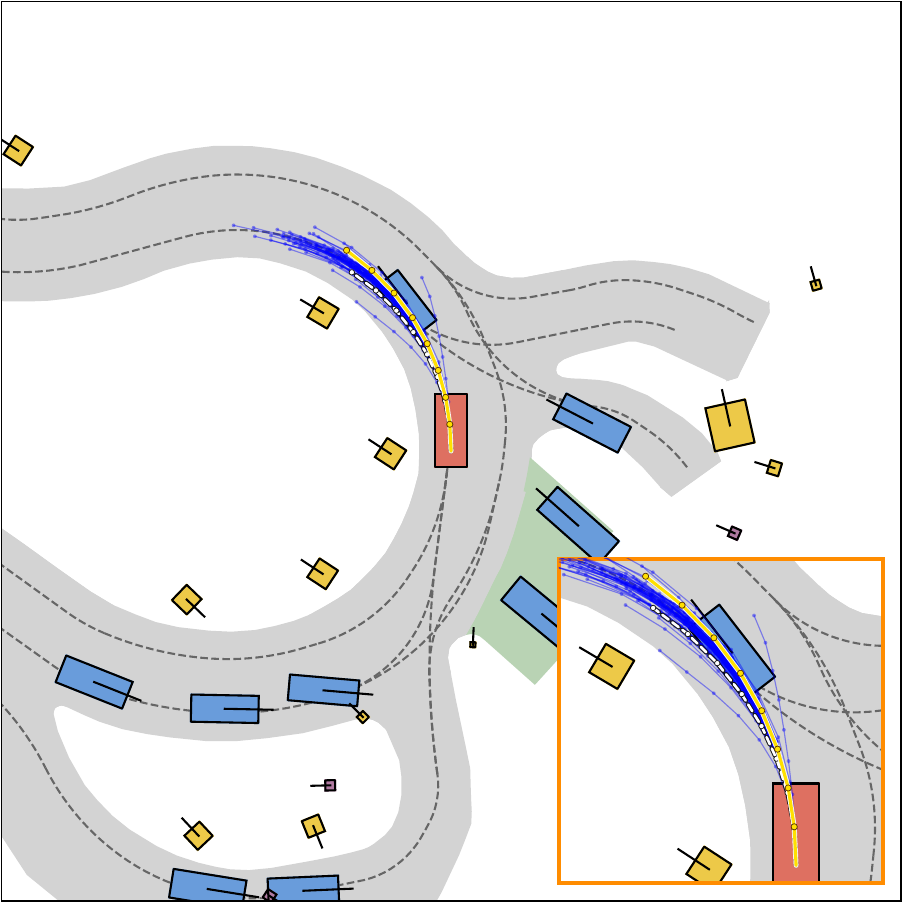}
\end{minipage}\hfill
\begin{minipage}[t]{.179\textwidth}\centering
{\fontsize{8}{9}\selectfont\bfseries Scene 4}\par\vspace{.5pt}
\includegraphics[width=\linewidth]{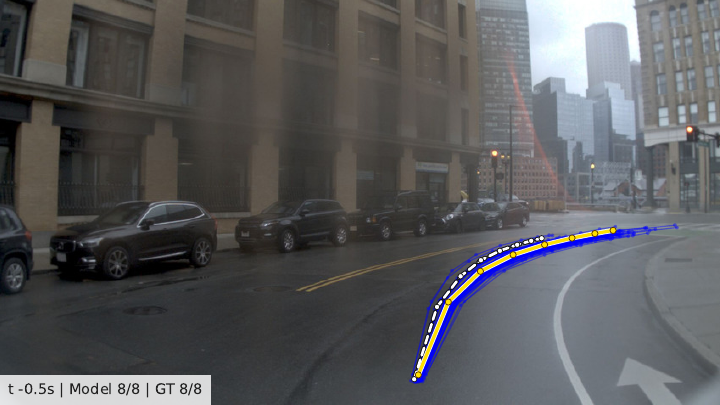}\par\vspace{.5pt}
\includegraphics[width=\linewidth]{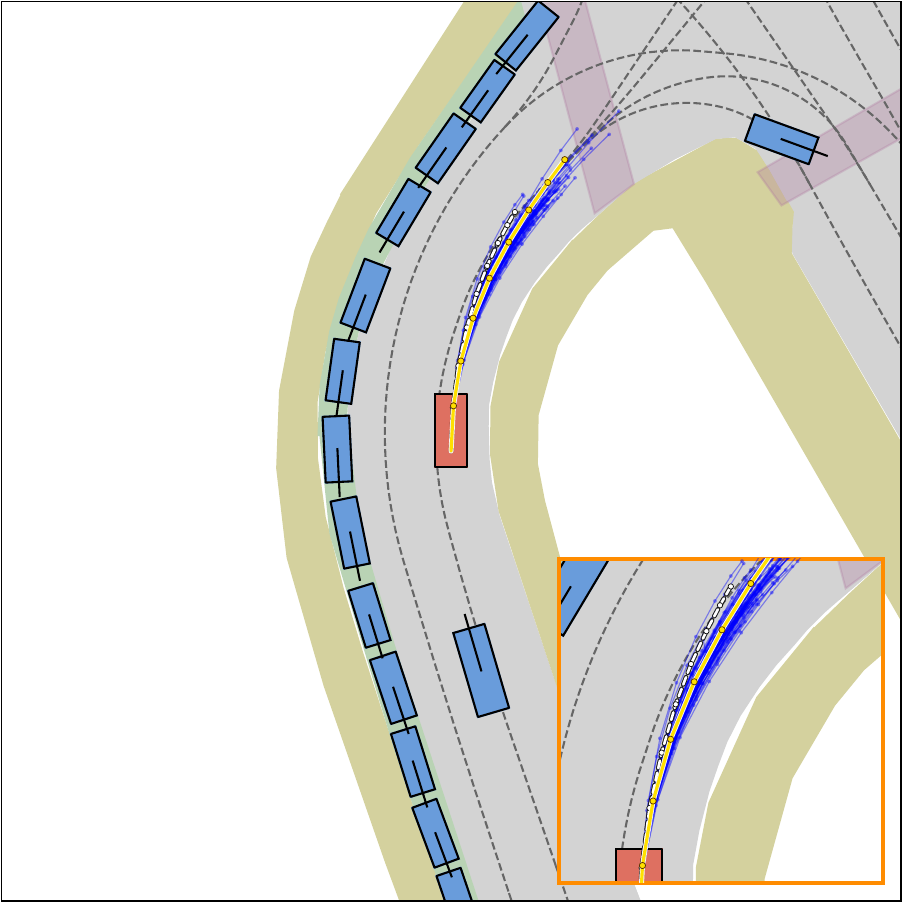}
\end{minipage}\hfill
\begin{minipage}[t]{.179\textwidth}\centering
{\fontsize{8}{9}\selectfont\bfseries Scene 5}\par\vspace{.5pt}
\includegraphics[width=\linewidth]{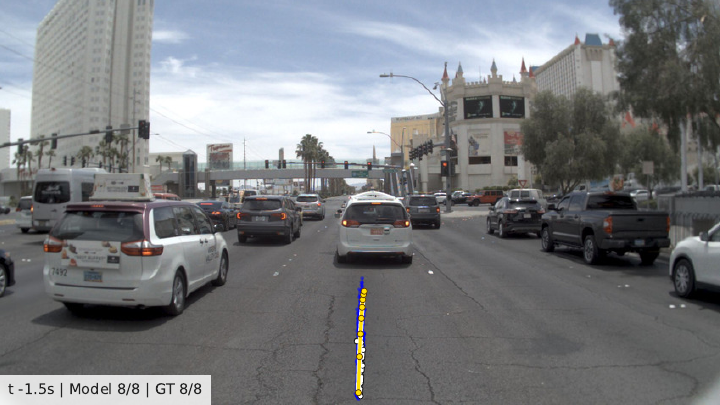}\par\vspace{.5pt}
\includegraphics[width=\linewidth]{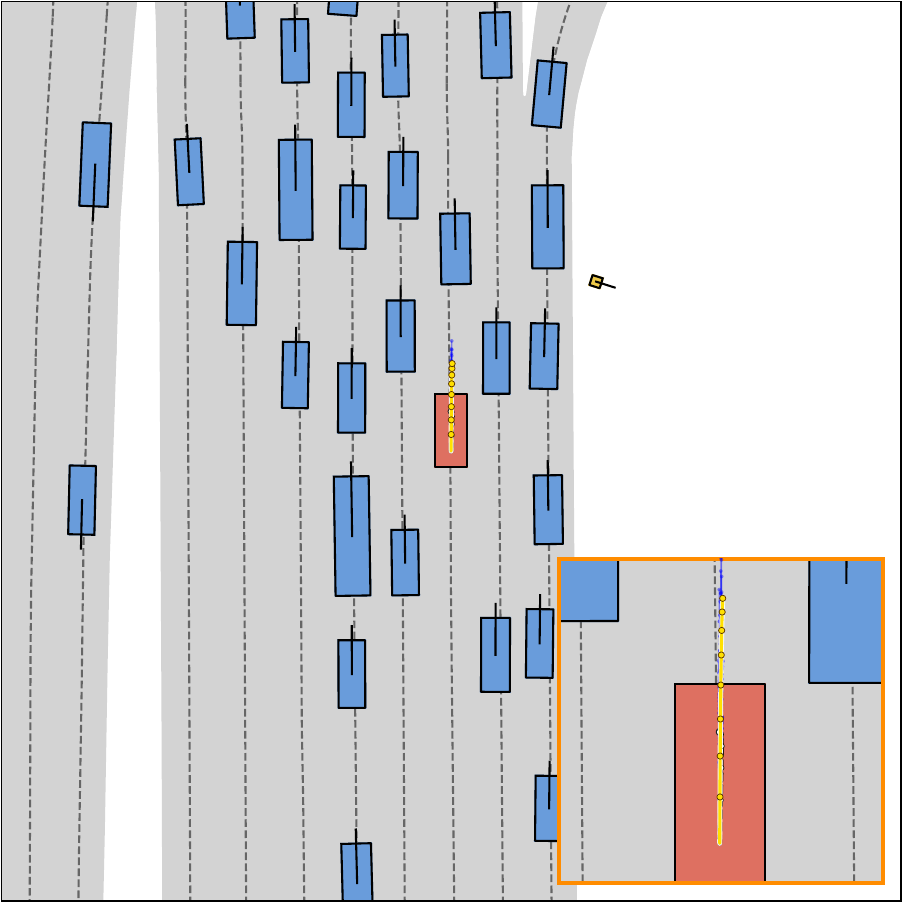}
\end{minipage}\par\vspace{1pt}
\begin{minipage}[t]{.179\textwidth}\centering
{\fontsize{8}{9}\selectfont\bfseries Scene 6}\par\vspace{.5pt}
\includegraphics[width=\linewidth]{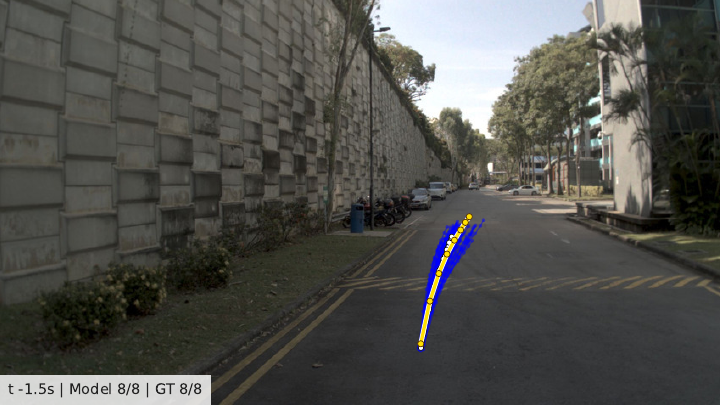}\par\vspace{.5pt}
\includegraphics[width=\linewidth]{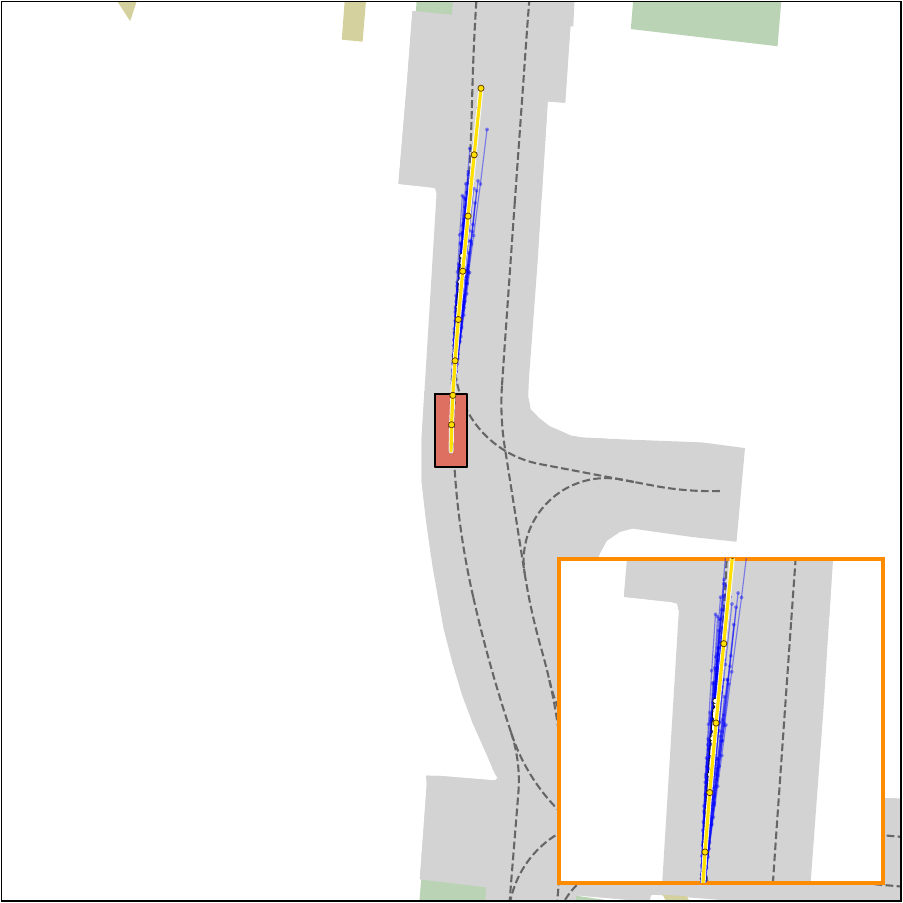}
\end{minipage}\hfill
\begin{minipage}[t]{.179\textwidth}\centering
{\fontsize{8}{9}\selectfont\bfseries Scene 7}\par\vspace{.5pt}
\includegraphics[width=\linewidth]{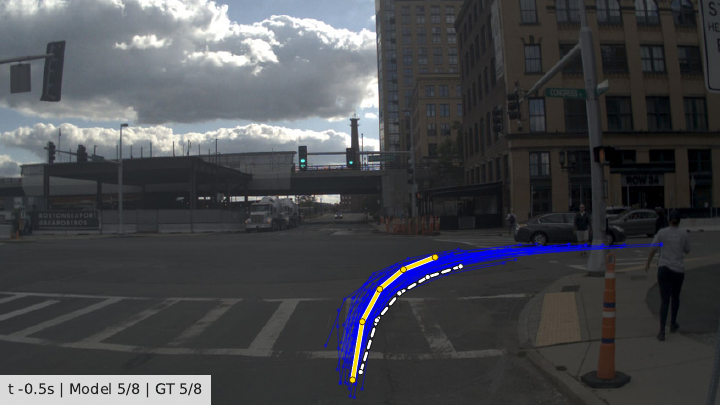}\par\vspace{.5pt}
\includegraphics[width=\linewidth]{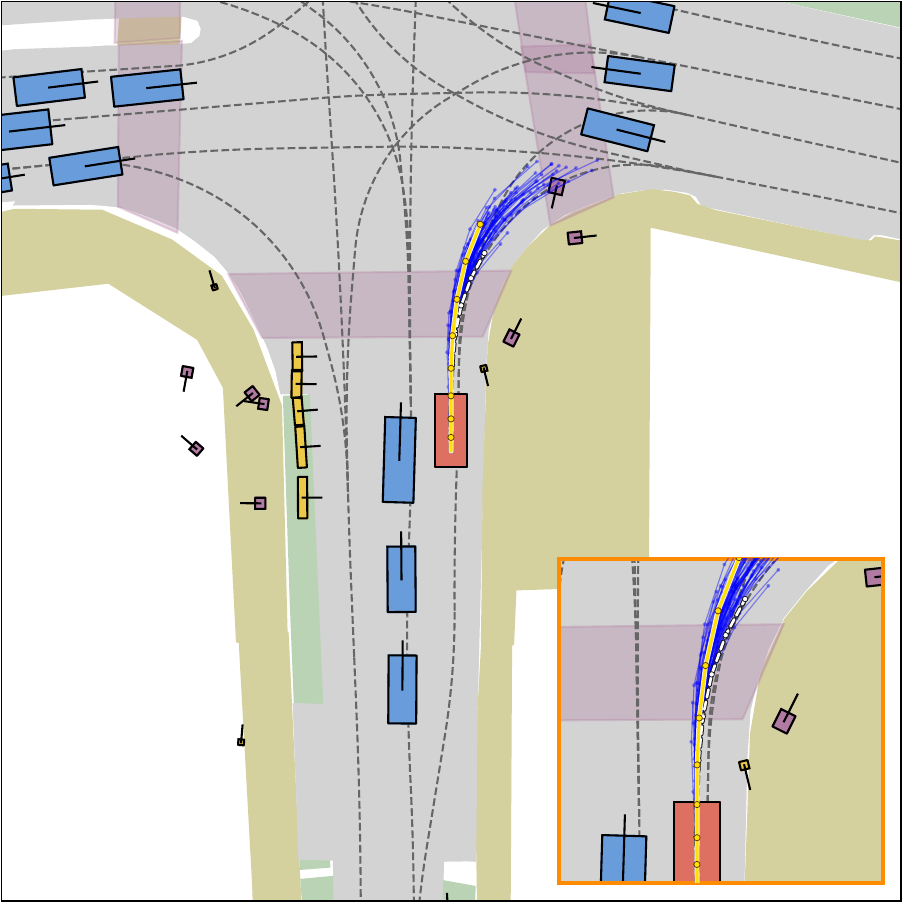}
\end{minipage}\hfill
\begin{minipage}[t]{.179\textwidth}\centering
{\fontsize{8}{9}\selectfont\bfseries Scene 8}\par\vspace{.5pt}
\includegraphics[width=\linewidth]{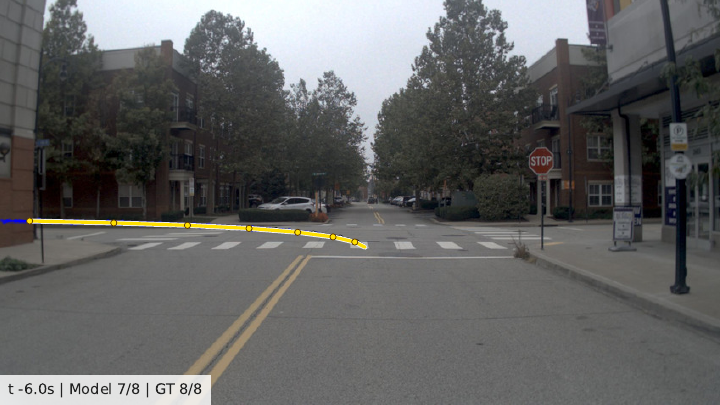}\par\vspace{.5pt}
\includegraphics[width=\linewidth]{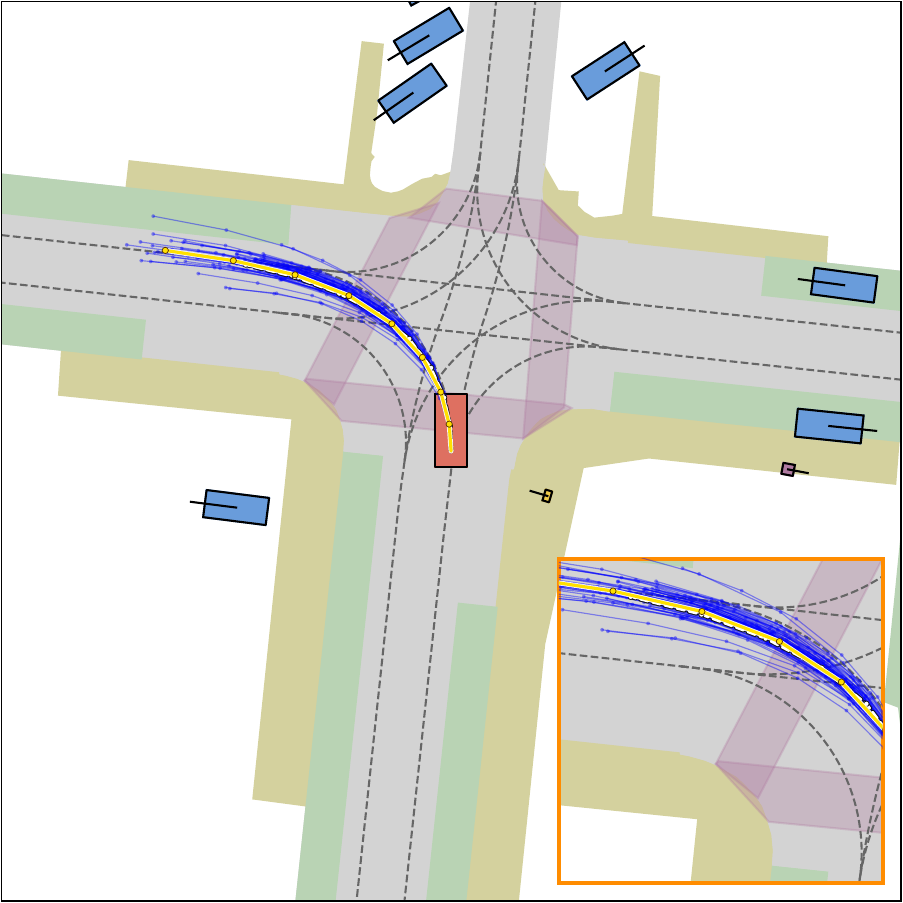}
\end{minipage}\hfill
\begin{minipage}[t]{.179\textwidth}\centering
{\fontsize{8}{9}\selectfont\bfseries Scene 9}\par\vspace{.5pt}
\includegraphics[width=\linewidth]{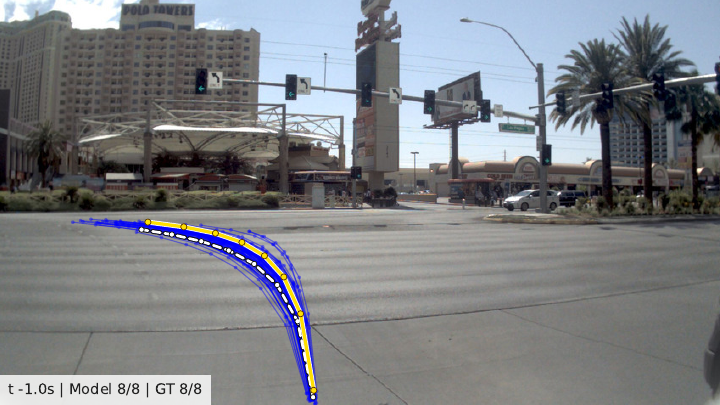}\par\vspace{.5pt}
\includegraphics[width=\linewidth]{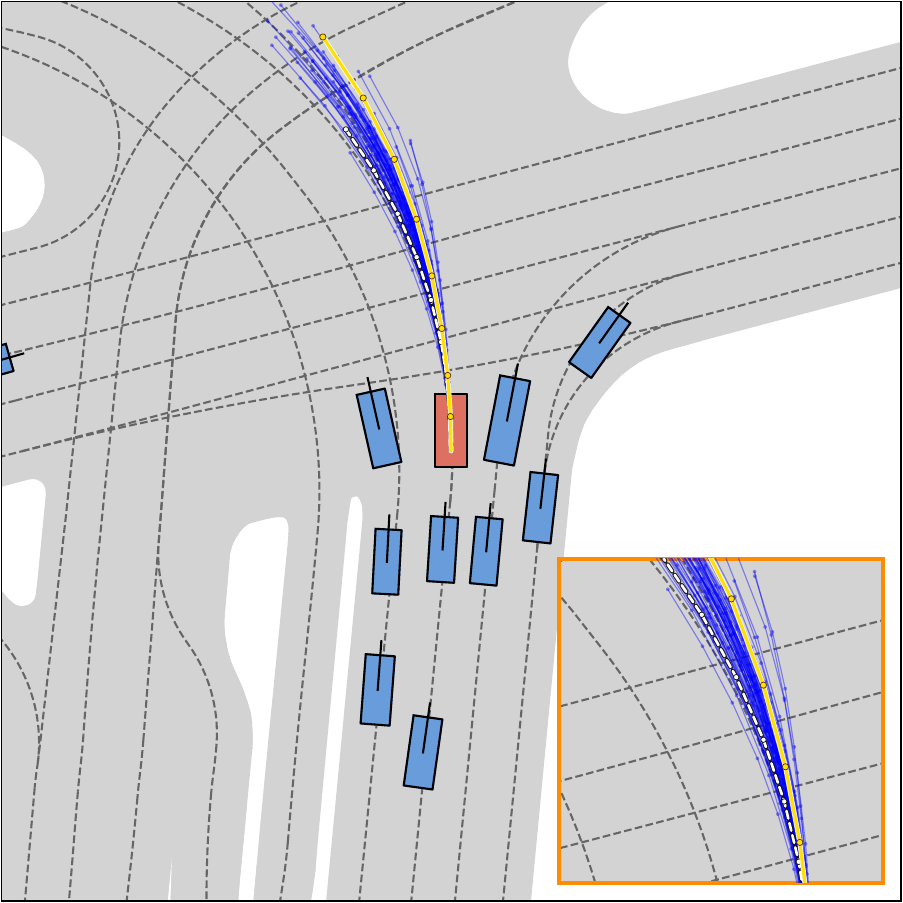}
\end{minipage}\hfill
\begin{minipage}[t]{.179\textwidth}\centering
{\fontsize{8}{9}\selectfont\bfseries Scene 10}\par\vspace{.5pt}
\includegraphics[width=\linewidth]{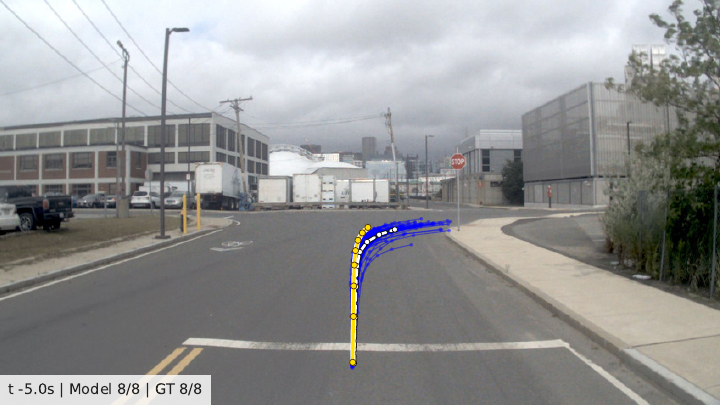}\par\vspace{.5pt}
\includegraphics[width=\linewidth]{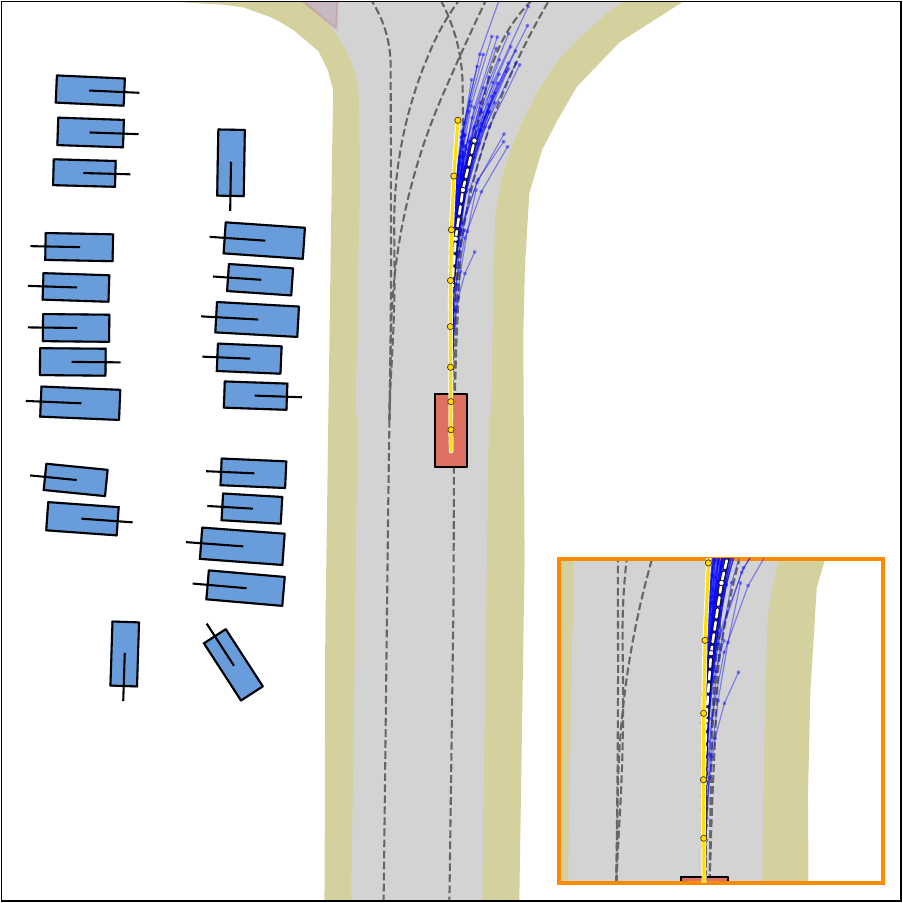}
\end{minipage}\par\vspace{1pt}
\begin{minipage}[t]{.179\textwidth}\centering
{\fontsize{8}{9}\selectfont\bfseries Scene 11}\par\vspace{.5pt}
\includegraphics[width=\linewidth]{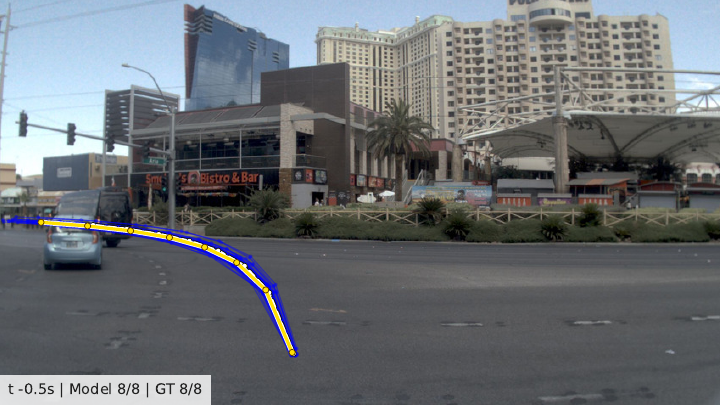}\par\vspace{.5pt}
\includegraphics[width=\linewidth]{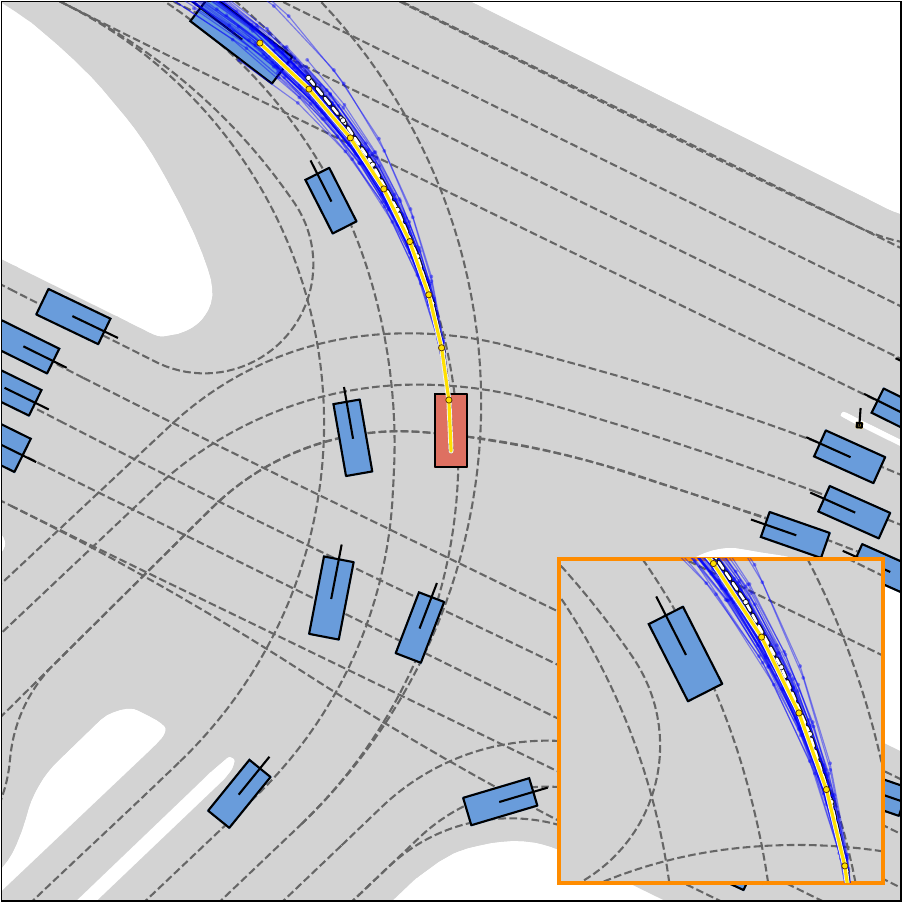}
\end{minipage}\hfill
\begin{minipage}[t]{.179\textwidth}\centering
{\fontsize{8}{9}\selectfont\bfseries Scene 12}\par\vspace{.5pt}
\includegraphics[width=\linewidth]{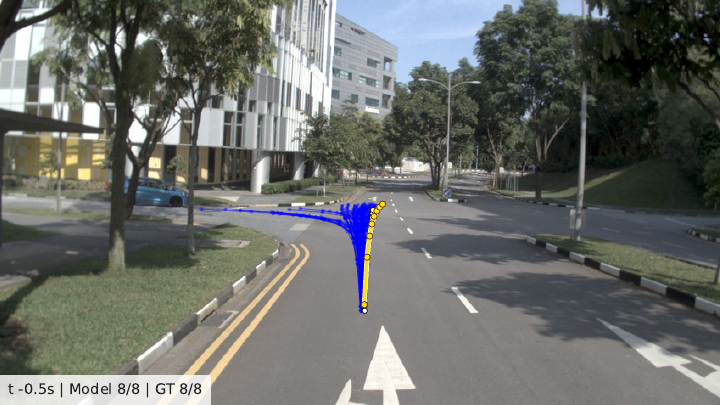}\par\vspace{.5pt}
\includegraphics[width=\linewidth]{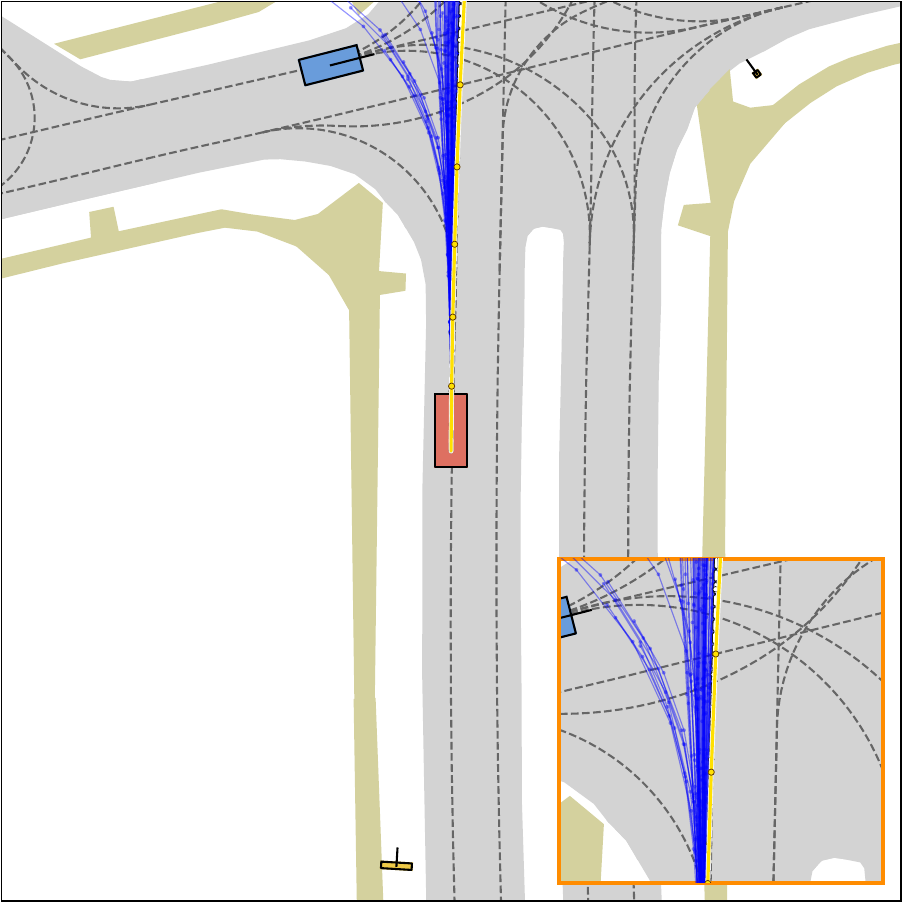}
\end{minipage}\hfill
\begin{minipage}[t]{.179\textwidth}\centering
{\fontsize{8}{9}\selectfont\bfseries Scene 13}\par\vspace{.5pt}
\includegraphics[width=\linewidth]{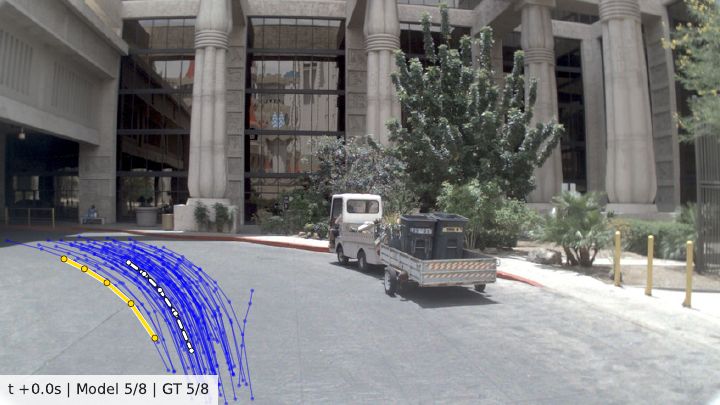}\par\vspace{.5pt}
\includegraphics[width=\linewidth]{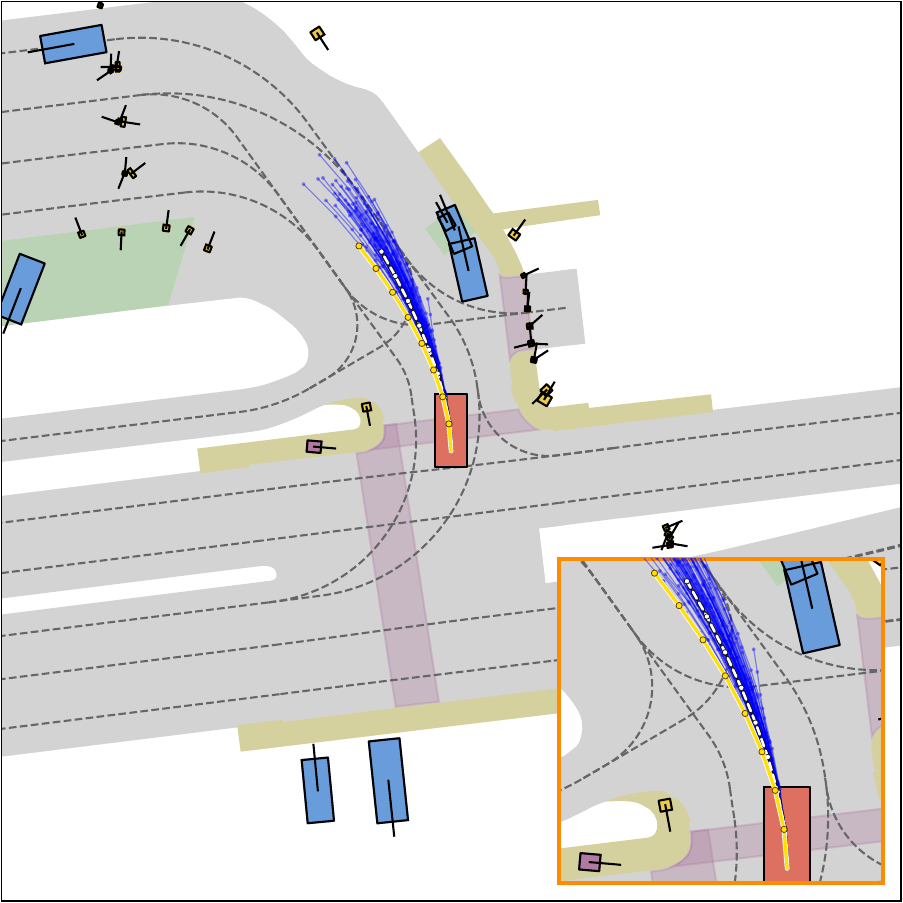}
\end{minipage}\hfill
\begin{minipage}[t]{.179\textwidth}\centering
{\fontsize{8}{9}\selectfont\bfseries Scene 14}\par\vspace{.5pt}
\includegraphics[width=\linewidth]{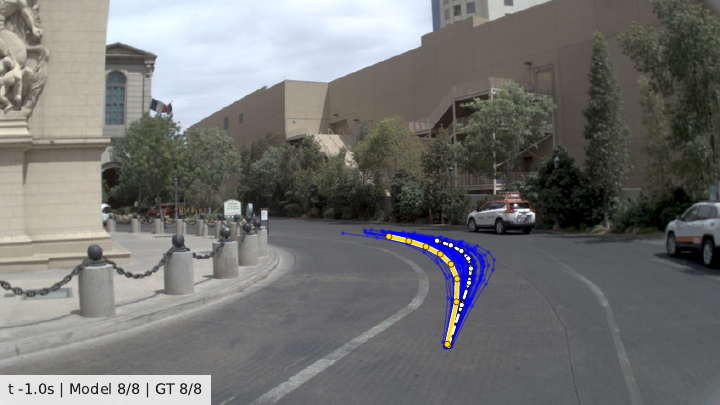}\par\vspace{.5pt}
\includegraphics[width=\linewidth]{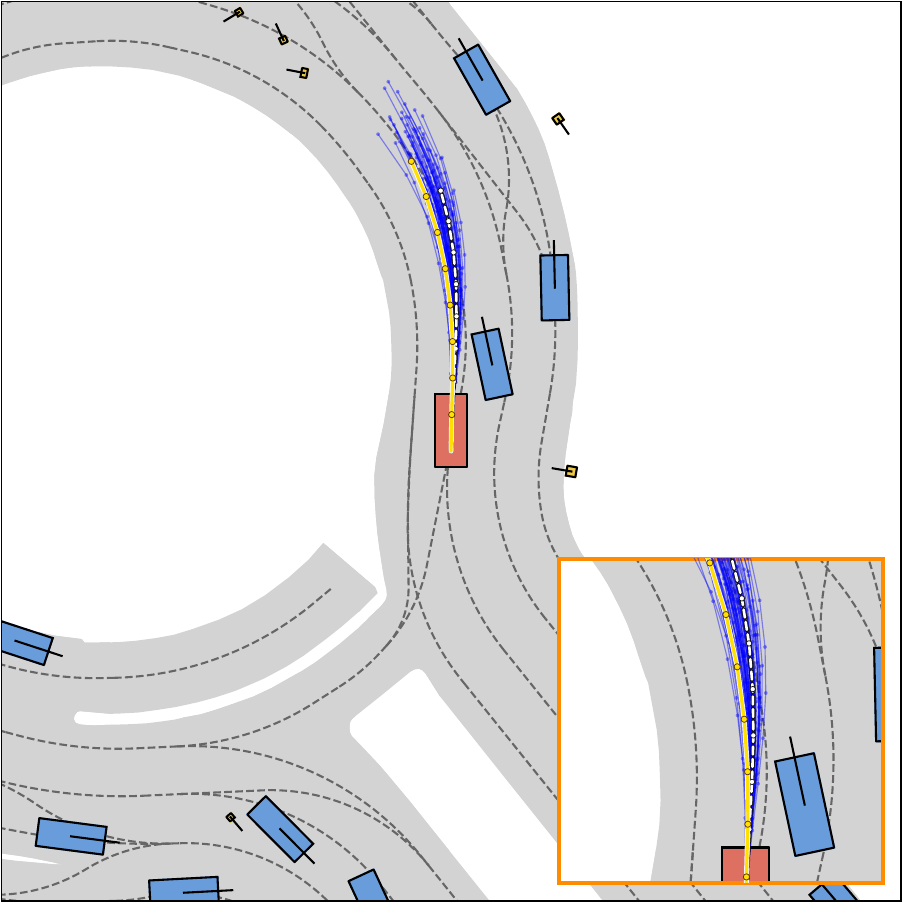}
\end{minipage}\hfill
\begin{minipage}[t]{.179\textwidth}\centering
{\fontsize{8}{9}\selectfont\bfseries Scene 15}\par\vspace{.5pt}
\includegraphics[width=\linewidth]{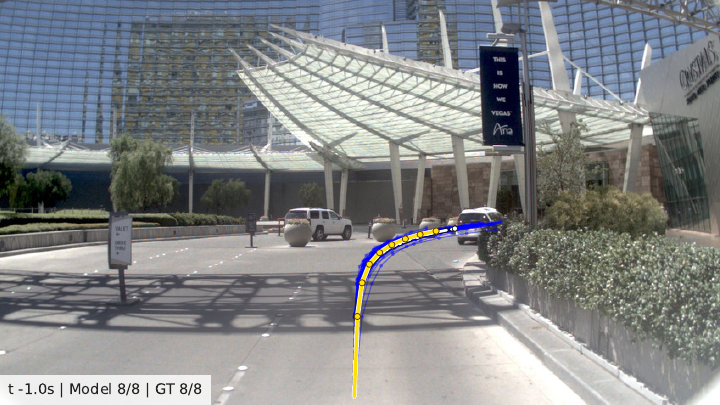}\par\vspace{.5pt}
\includegraphics[width=\linewidth]{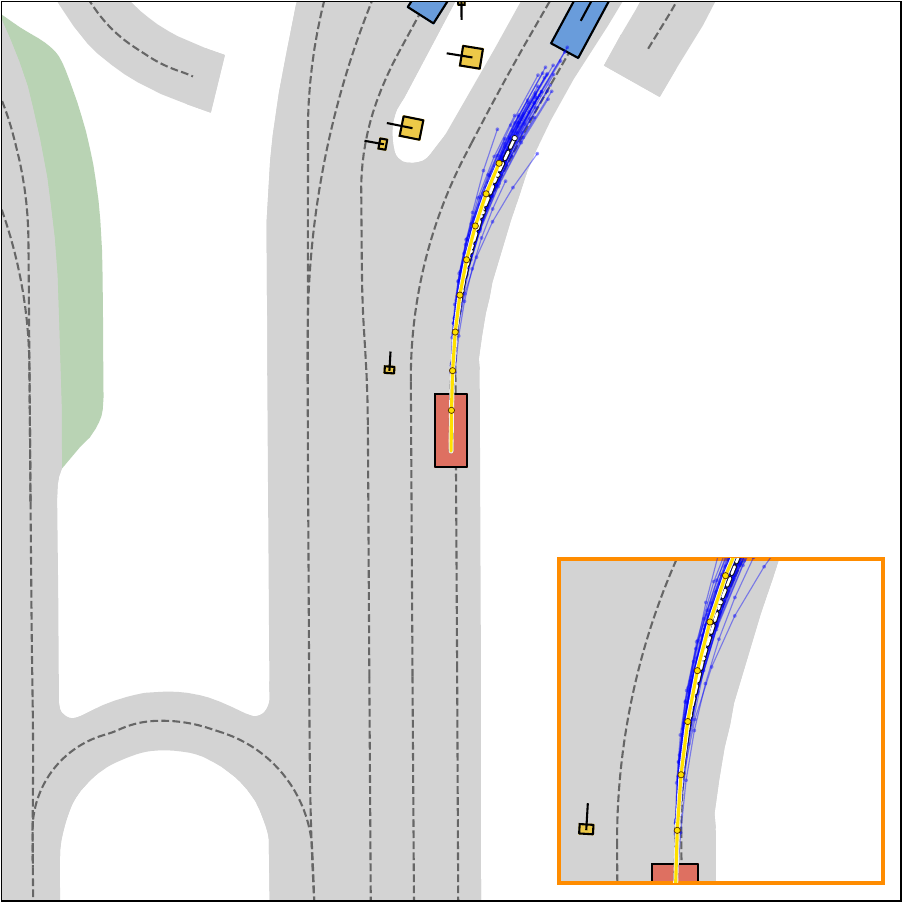}
\end{minipage}\par\vspace{1pt}
\begin{minipage}[t]{.179\textwidth}\centering
{\fontsize{8}{9}\selectfont\bfseries Scene 16}\par\vspace{.5pt}
\includegraphics[width=\linewidth]{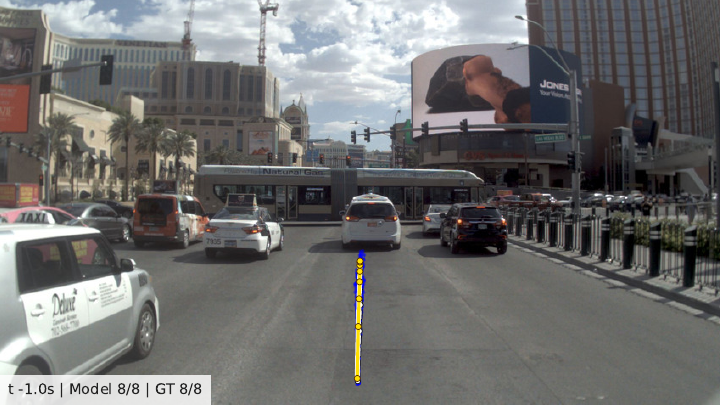}\par\vspace{.5pt}
\includegraphics[width=\linewidth]{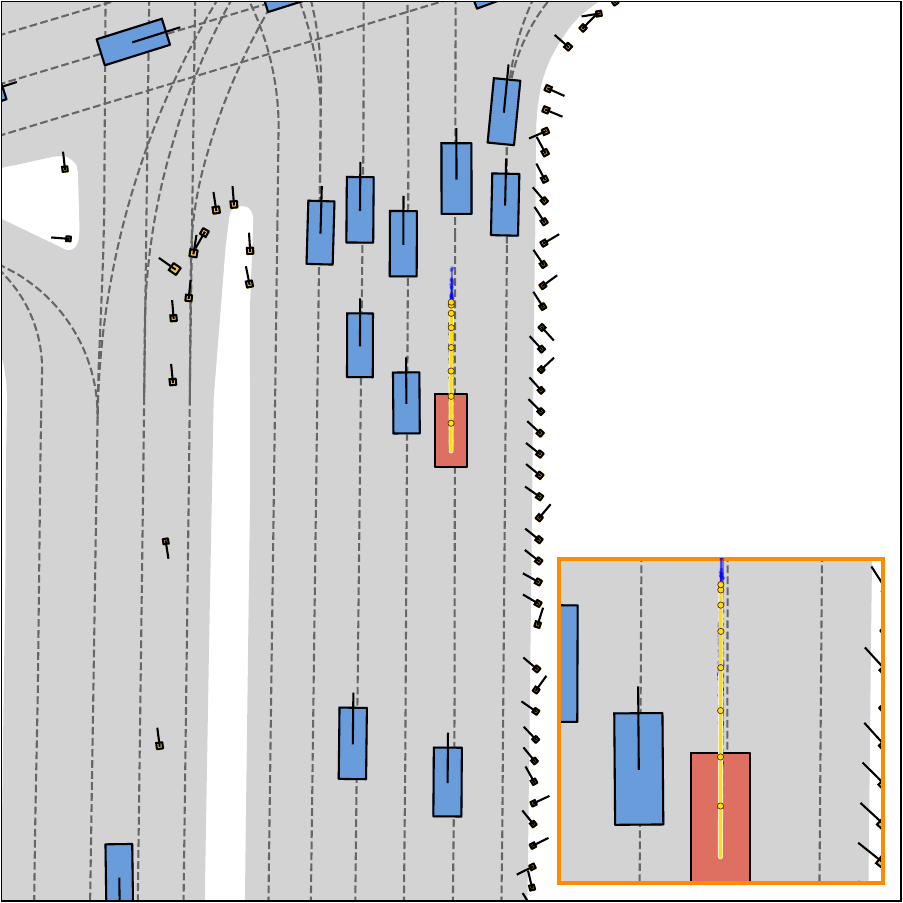}
\end{minipage}\hfill
\begin{minipage}[t]{.179\textwidth}\centering
{\fontsize{8}{9}\selectfont\bfseries Scene 17}\par\vspace{.5pt}
\includegraphics[width=\linewidth]{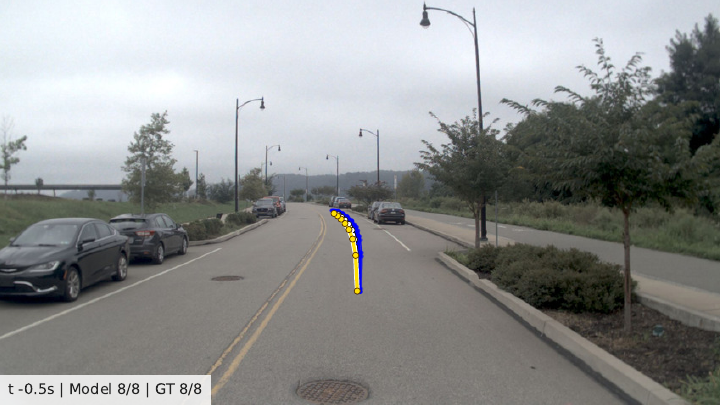}\par\vspace{.5pt}
\includegraphics[width=\linewidth]{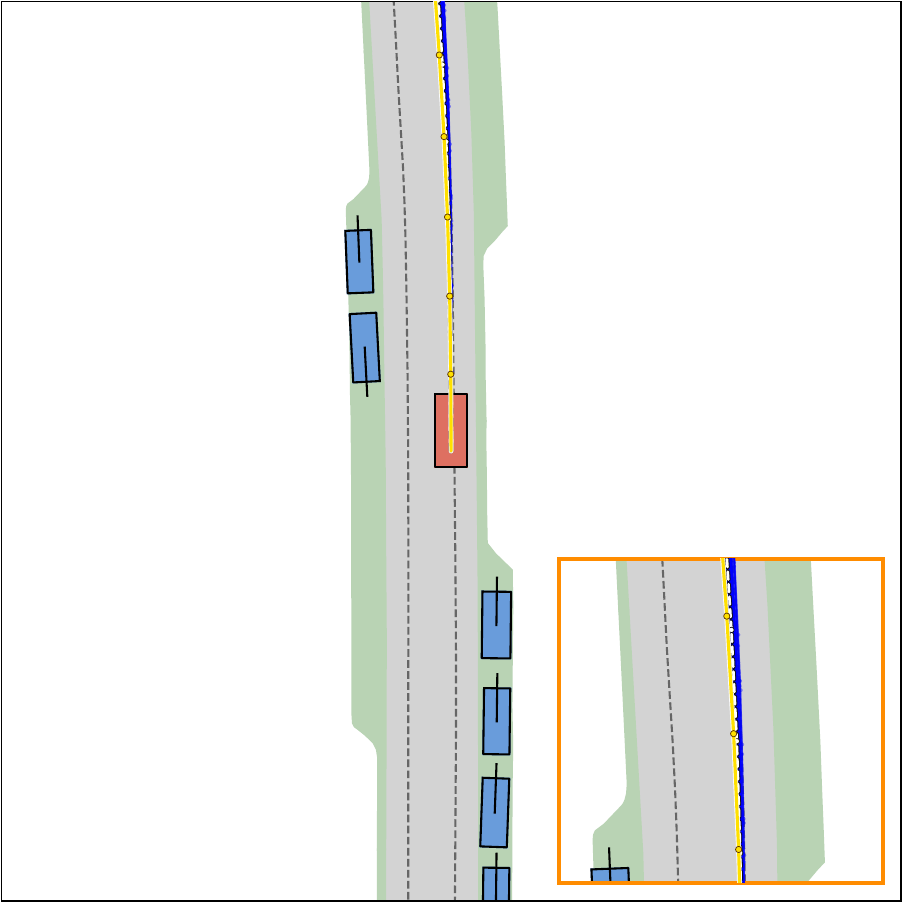}
\end{minipage}\hfill
\begin{minipage}[t]{.179\textwidth}\centering
{\fontsize{8}{9}\selectfont\bfseries Scene 18}\par\vspace{.5pt}
\includegraphics[width=\linewidth]{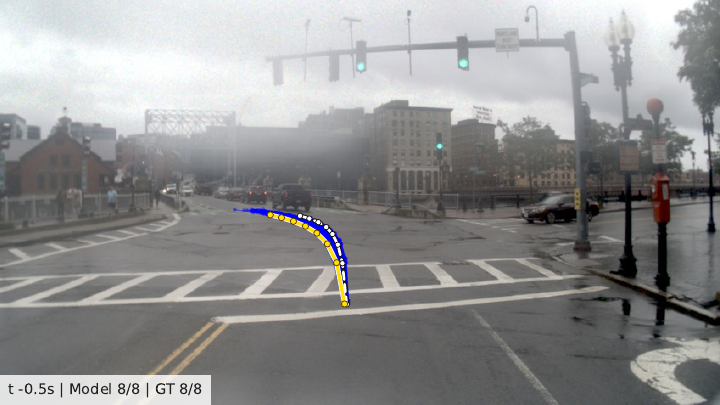}\par\vspace{.5pt}
\includegraphics[width=\linewidth]{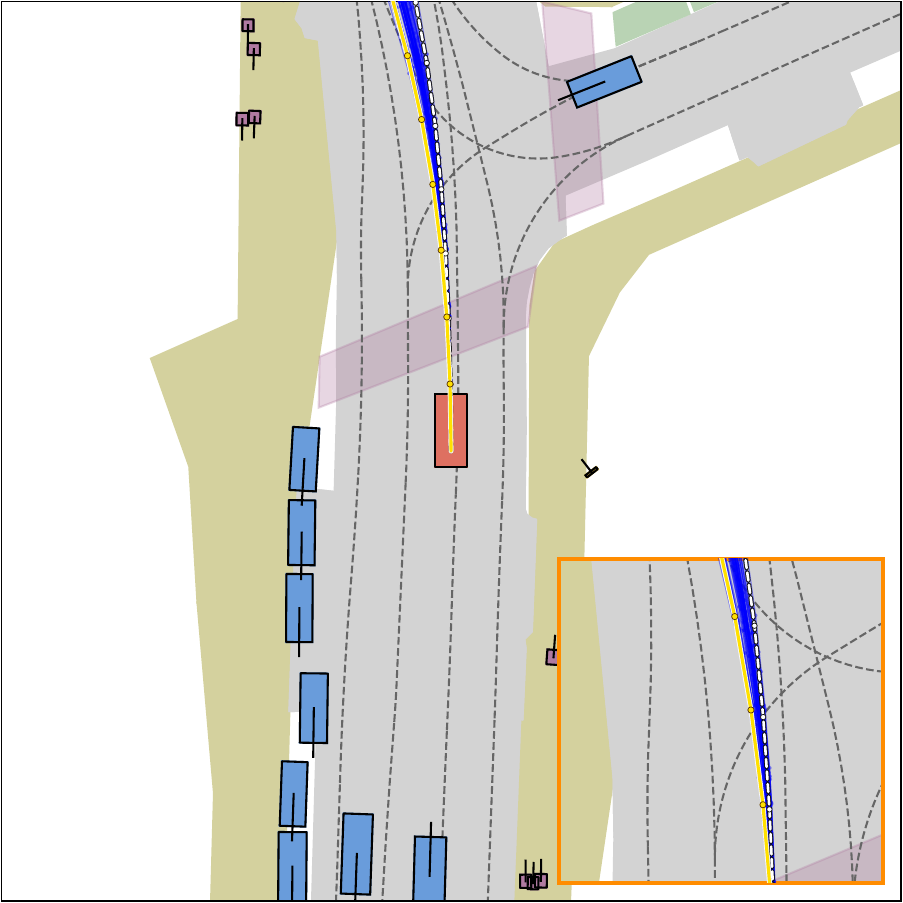}
\end{minipage}\hfill
\begin{minipage}[t]{.179\textwidth}\centering
{\fontsize{8}{9}\selectfont\bfseries Scene 19}\par\vspace{.5pt}
\includegraphics[width=\linewidth]{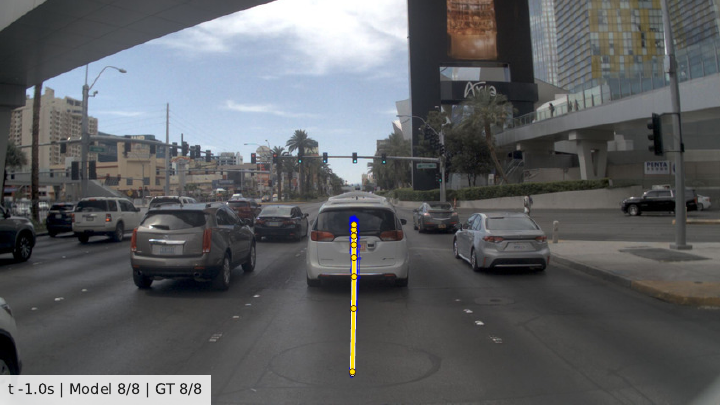}\par\vspace{.5pt}
\includegraphics[width=\linewidth]{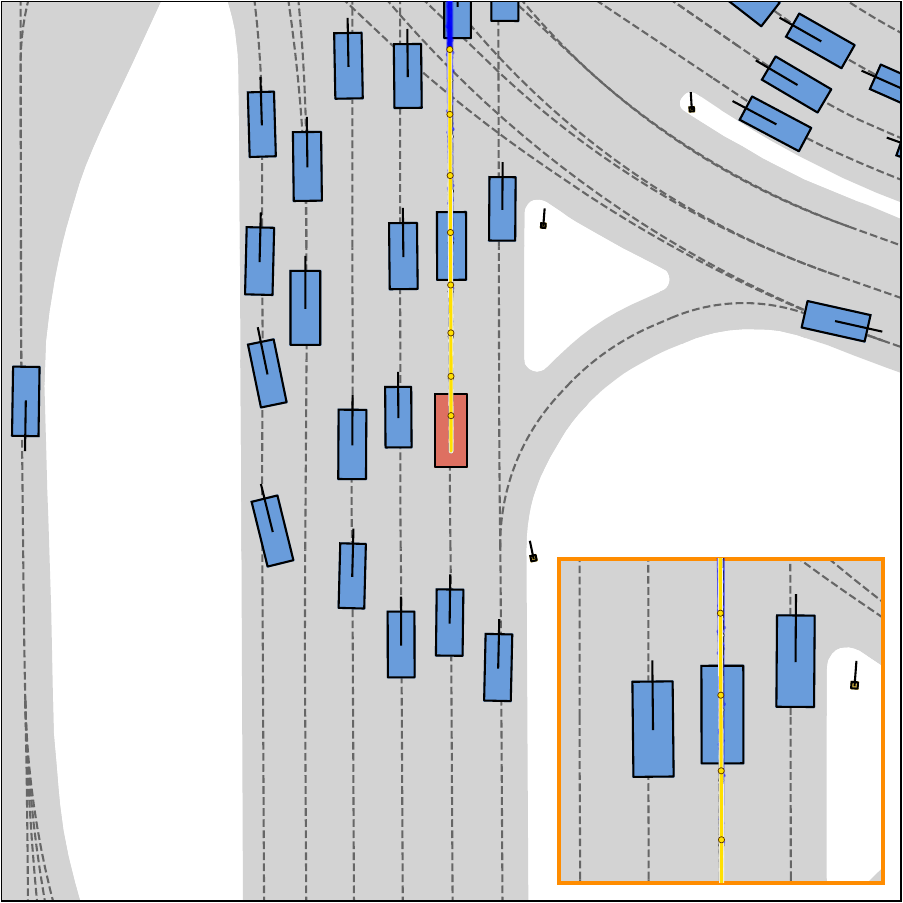}
\end{minipage}\hfill
\begin{minipage}[t]{.179\textwidth}\centering
{\fontsize{8}{9}\selectfont\bfseries Scene 20}\par\vspace{.5pt}
\includegraphics[width=\linewidth]{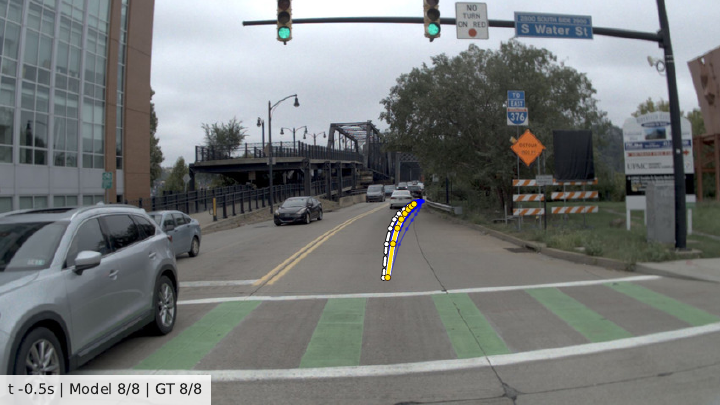}\par\vspace{.5pt}
\includegraphics[width=\linewidth]{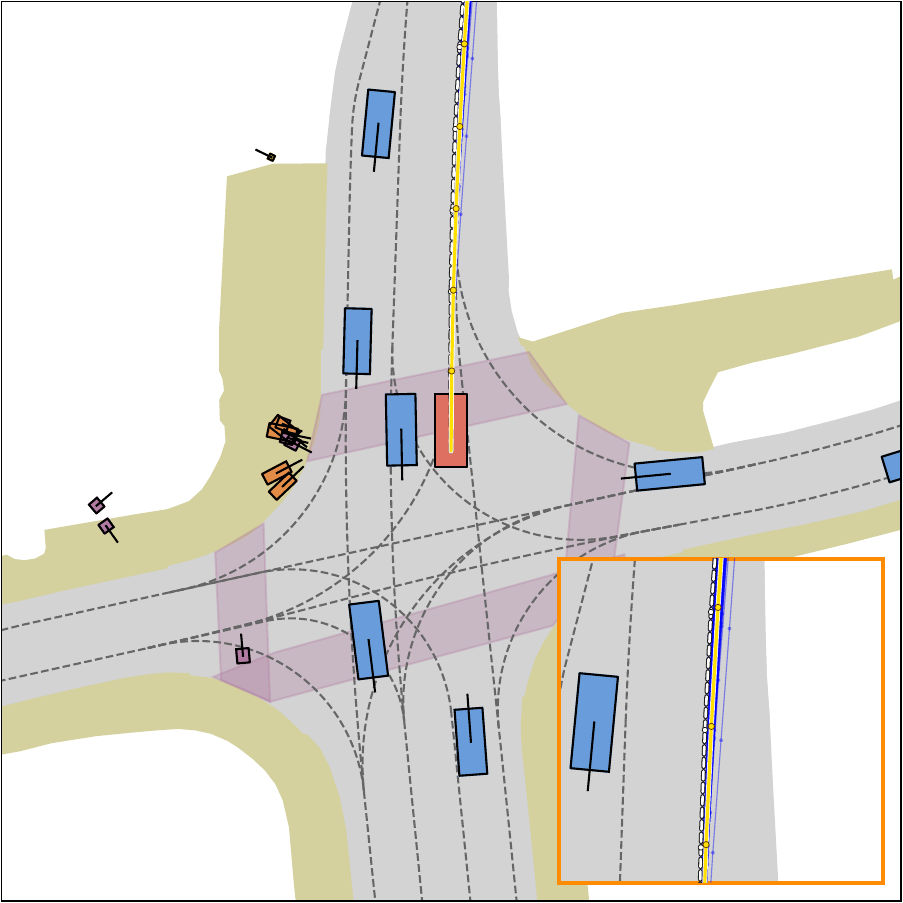}
\end{minipage}\par\vspace{.5pt}
\caption{\textbf{DriveTTO outputs, scenes 1--20.} Camera projections and BEV show blue native proposals, the yellow selected output, and white human motion. The orange insets show how the proposals differ in turning radius at the roundabout (scene 3) and in the right turn (scene 7). All 20 selected outputs score 100 PDMS.}
\label{fig:qual_scenes}
\end{figure*}

\begin{figure*}[p]
\centering
\begin{minipage}[t]{.179\textwidth}\centering
{\fontsize{8}{9}\selectfont\bfseries Scene 21}\par\vspace{.5pt}
\includegraphics[width=\linewidth]{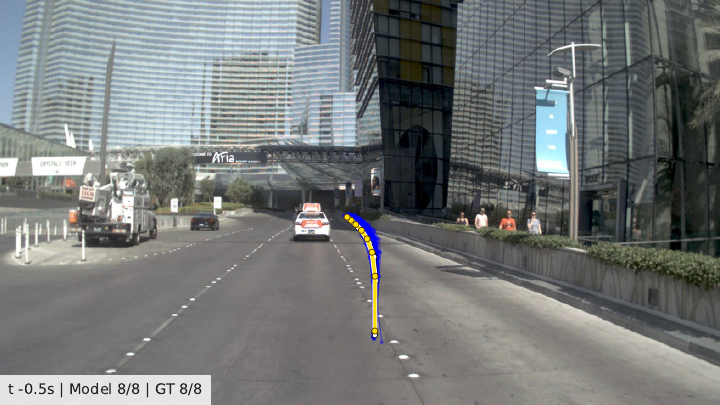}\par\vspace{.5pt}
\includegraphics[width=\linewidth]{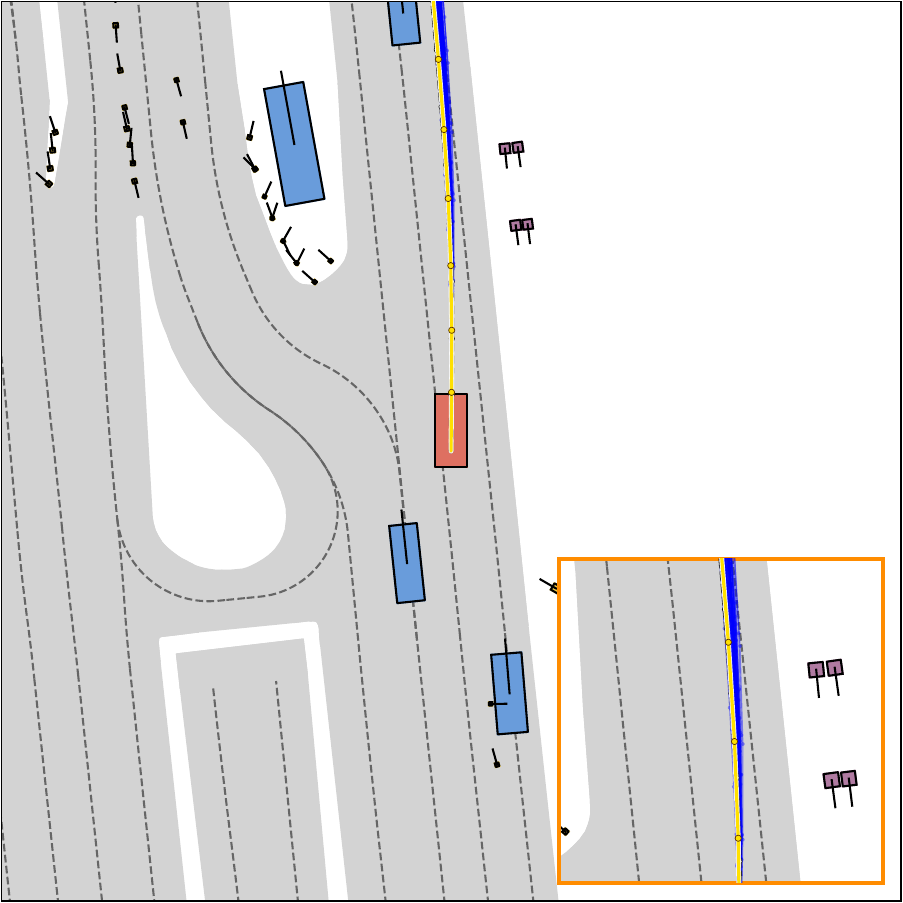}
\end{minipage}\hfill
\begin{minipage}[t]{.179\textwidth}\centering
{\fontsize{8}{9}\selectfont\bfseries Scene 22}\par\vspace{.5pt}
\includegraphics[width=\linewidth]{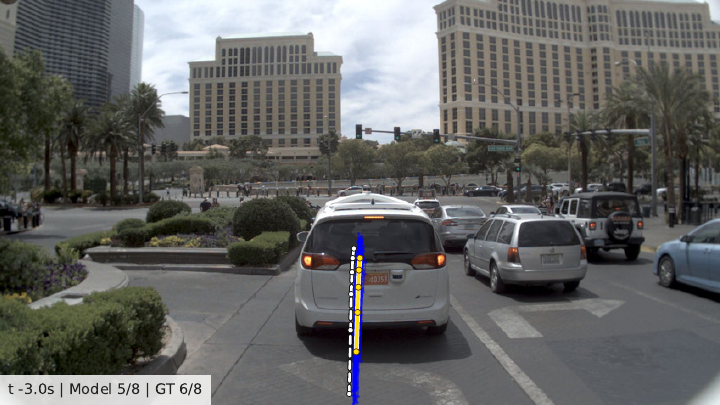}\par\vspace{.5pt}
\includegraphics[width=\linewidth]{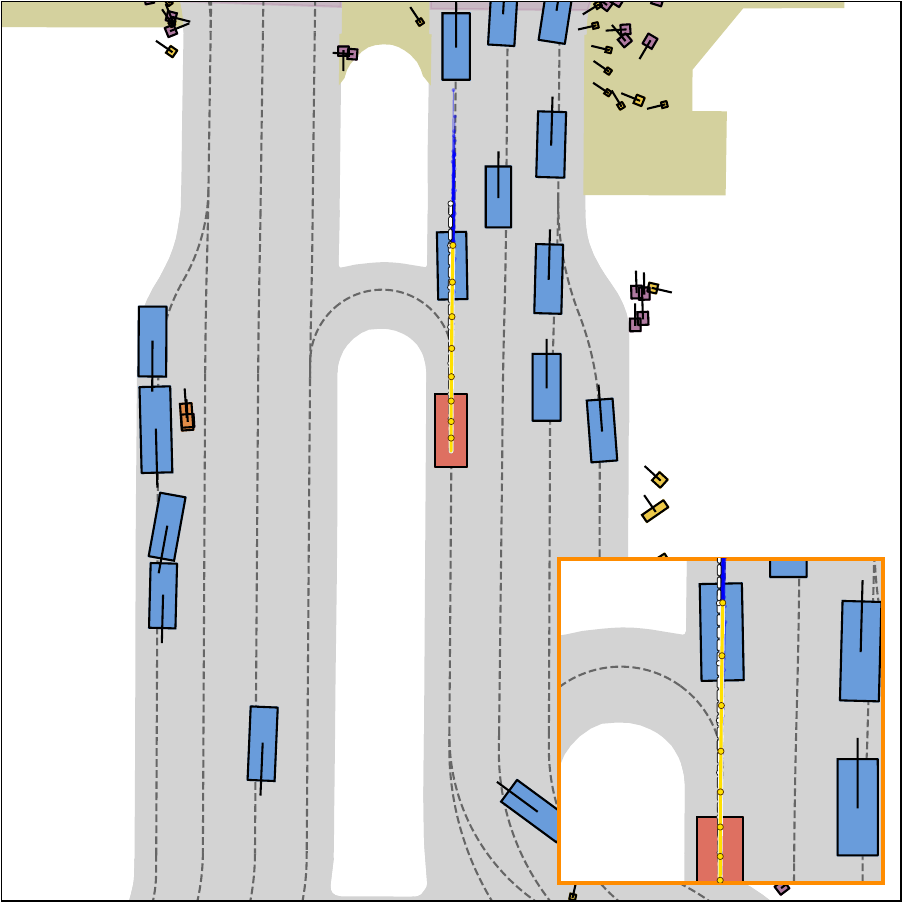}
\end{minipage}\hfill
\begin{minipage}[t]{.179\textwidth}\centering
{\fontsize{8}{9}\selectfont\bfseries Scene 23}\par\vspace{.5pt}
\includegraphics[width=\linewidth]{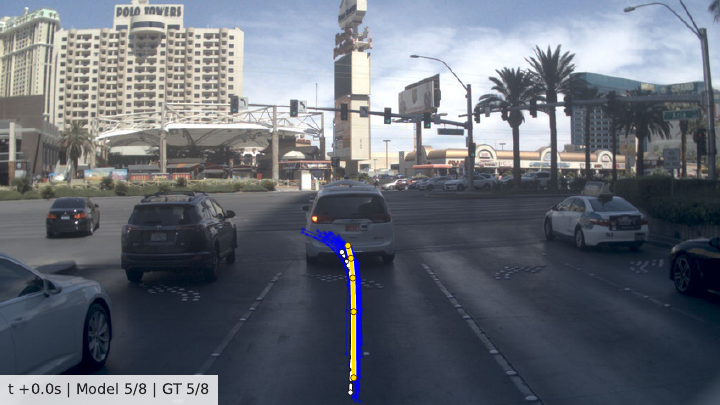}\par\vspace{.5pt}
\includegraphics[width=\linewidth]{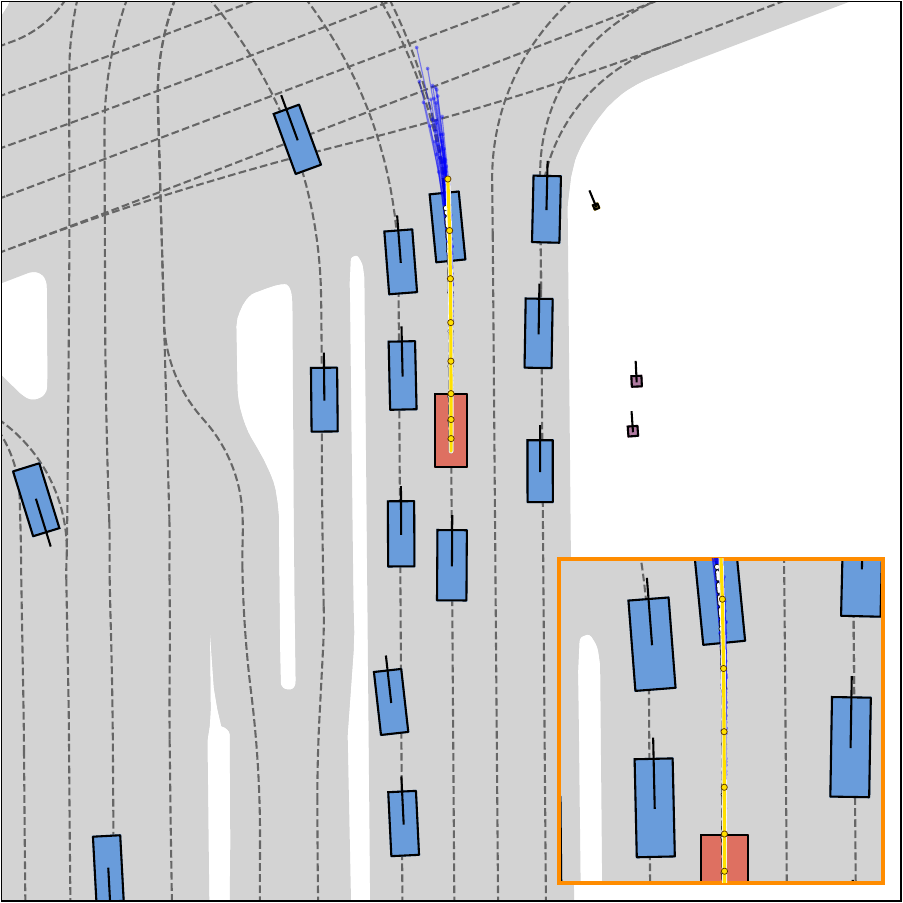}
\end{minipage}\hfill
\begin{minipage}[t]{.179\textwidth}\centering
{\fontsize{8}{9}\selectfont\bfseries Scene 24}\par\vspace{.5pt}
\includegraphics[width=\linewidth]{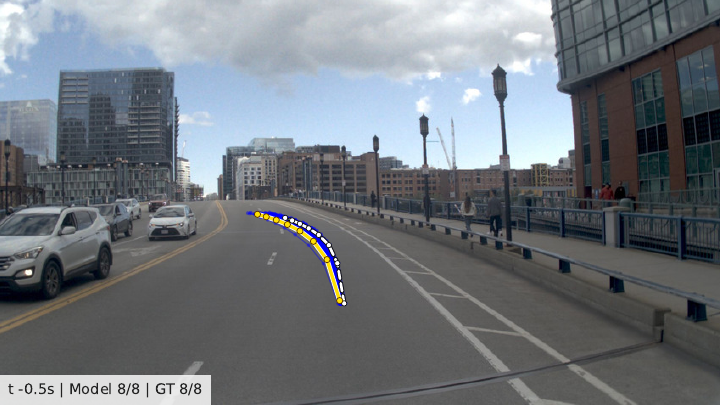}\par\vspace{.5pt}
\includegraphics[width=\linewidth]{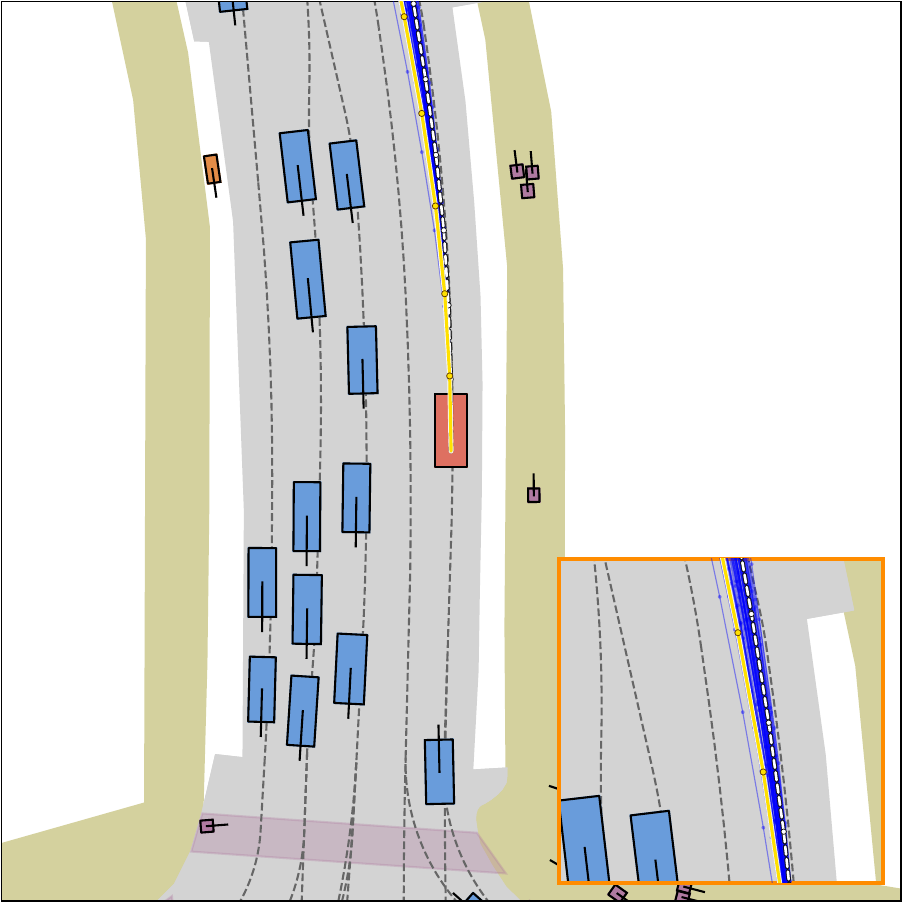}
\end{minipage}\hfill
\begin{minipage}[t]{.179\textwidth}\centering
{\fontsize{8}{9}\selectfont\bfseries Scene 25}\par\vspace{.5pt}
\includegraphics[width=\linewidth]{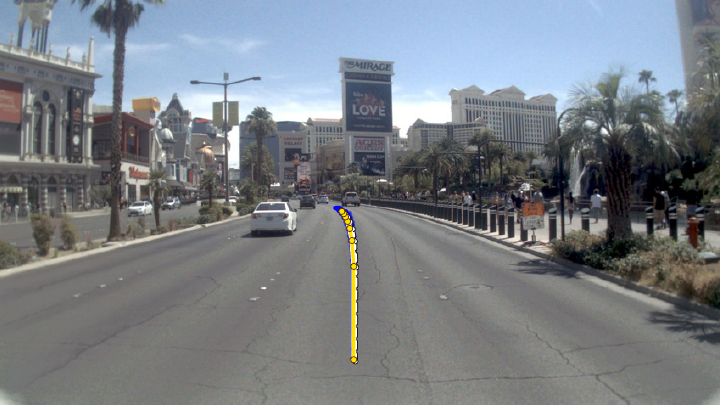}\par\vspace{.5pt}
\includegraphics[width=\linewidth]{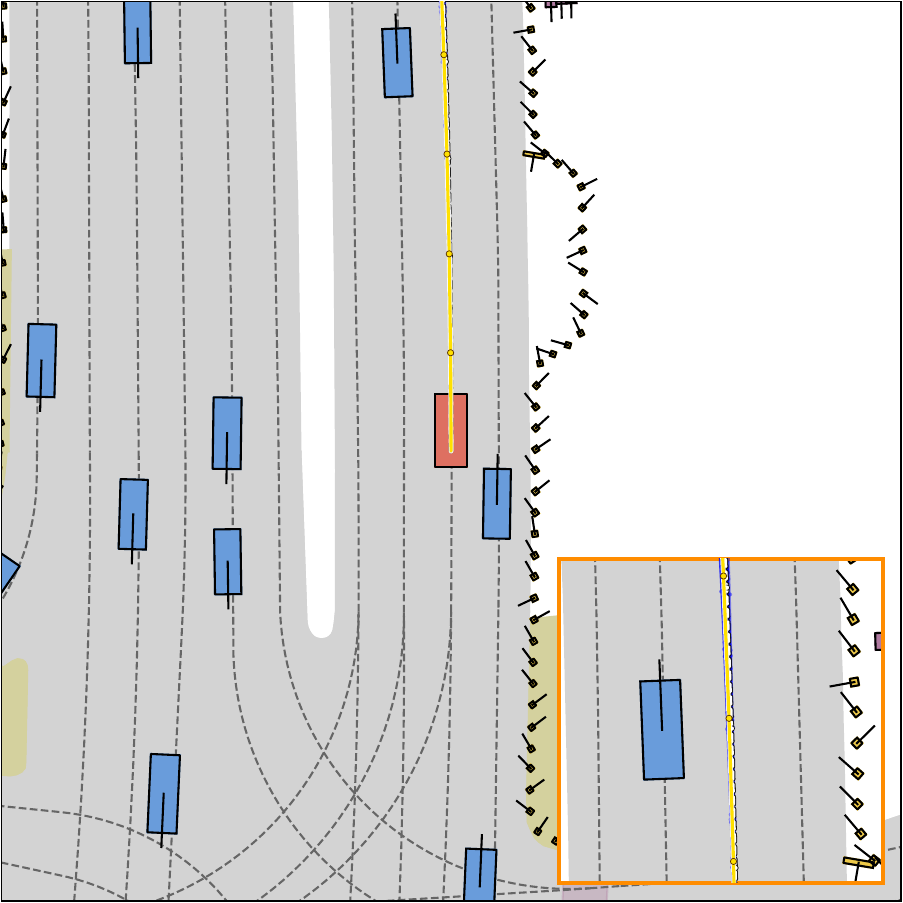}
\end{minipage}\par\vspace{1pt}
\begin{minipage}[t]{.179\textwidth}\centering
{\fontsize{8}{9}\selectfont\bfseries Scene 26}\par\vspace{.5pt}
\includegraphics[width=\linewidth]{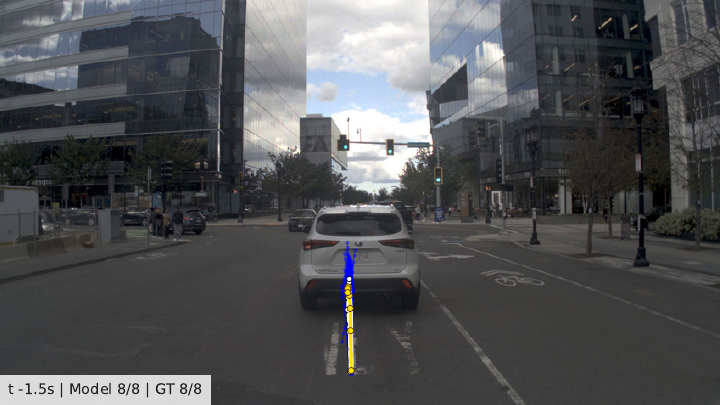}\par\vspace{.5pt}
\includegraphics[width=\linewidth]{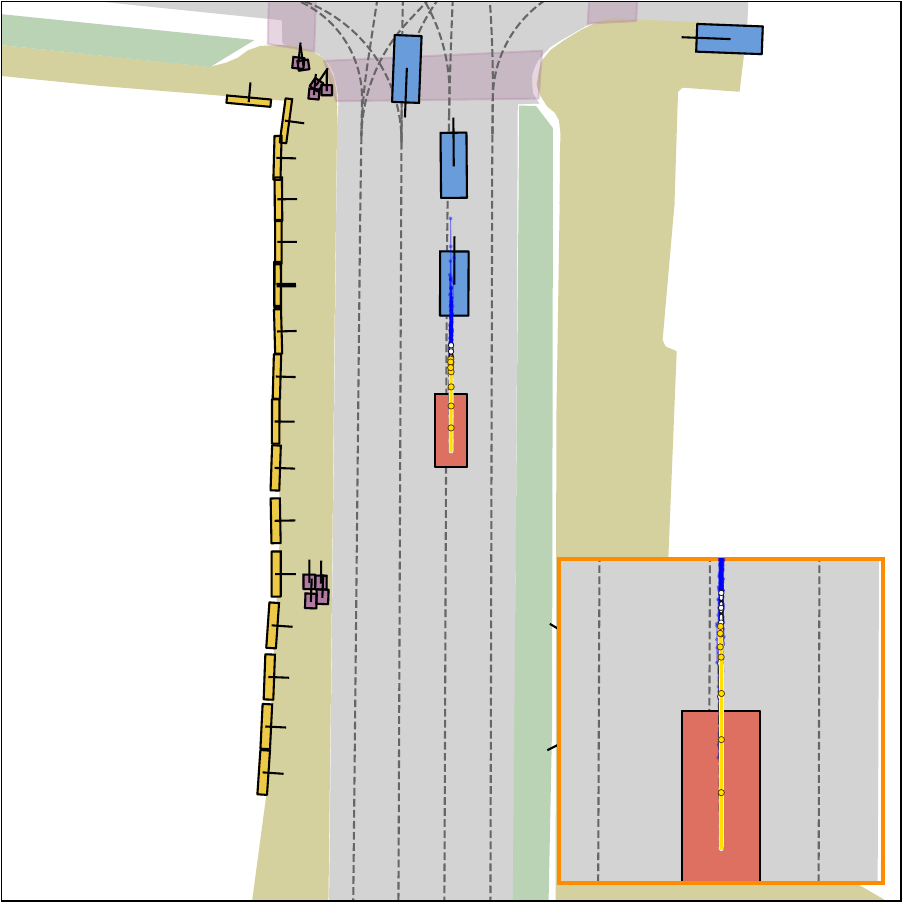}
\end{minipage}\hfill
\begin{minipage}[t]{.179\textwidth}\centering
{\fontsize{8}{9}\selectfont\bfseries Scene 27}\par\vspace{.5pt}
\includegraphics[width=\linewidth]{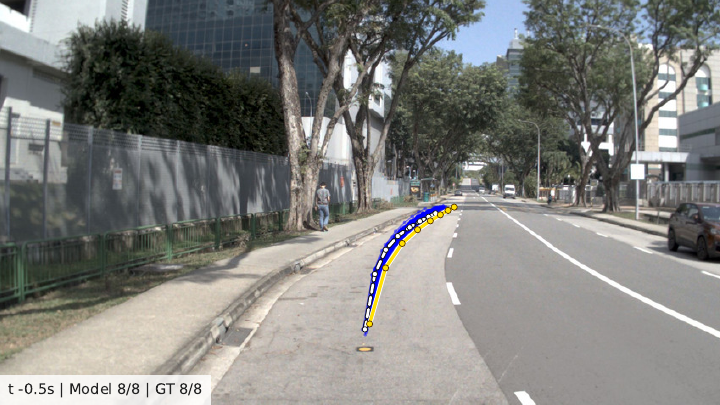}\par\vspace{.5pt}
\includegraphics[width=\linewidth]{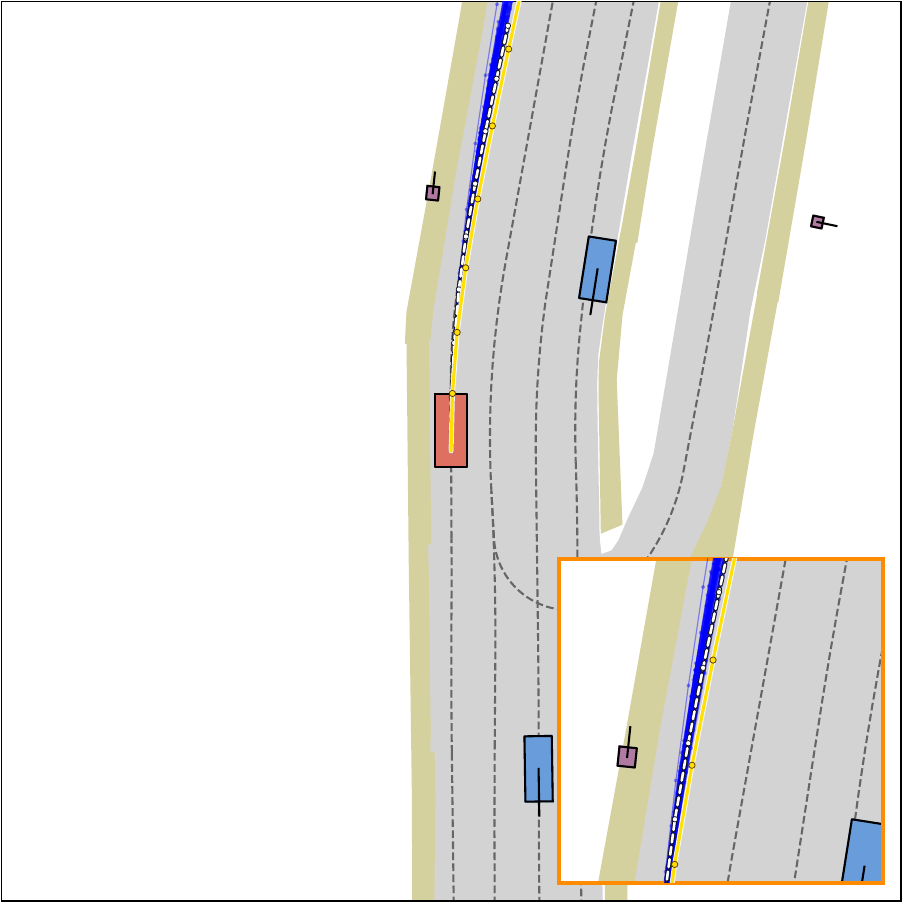}
\end{minipage}\hfill
\begin{minipage}[t]{.179\textwidth}\centering
{\fontsize{8}{9}\selectfont\bfseries Scene 28}\par\vspace{.5pt}
\includegraphics[width=\linewidth]{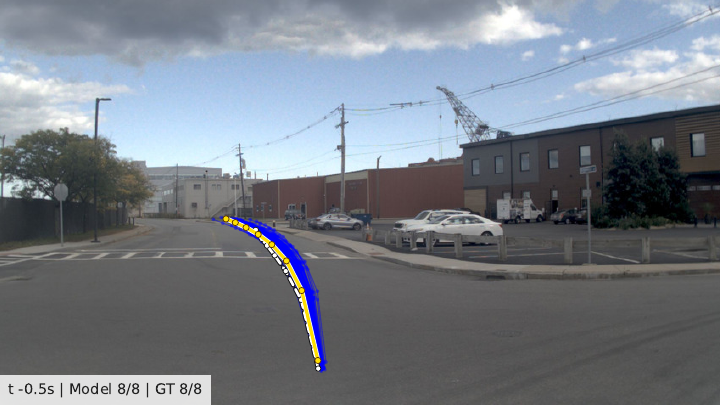}\par\vspace{.5pt}
\includegraphics[width=\linewidth]{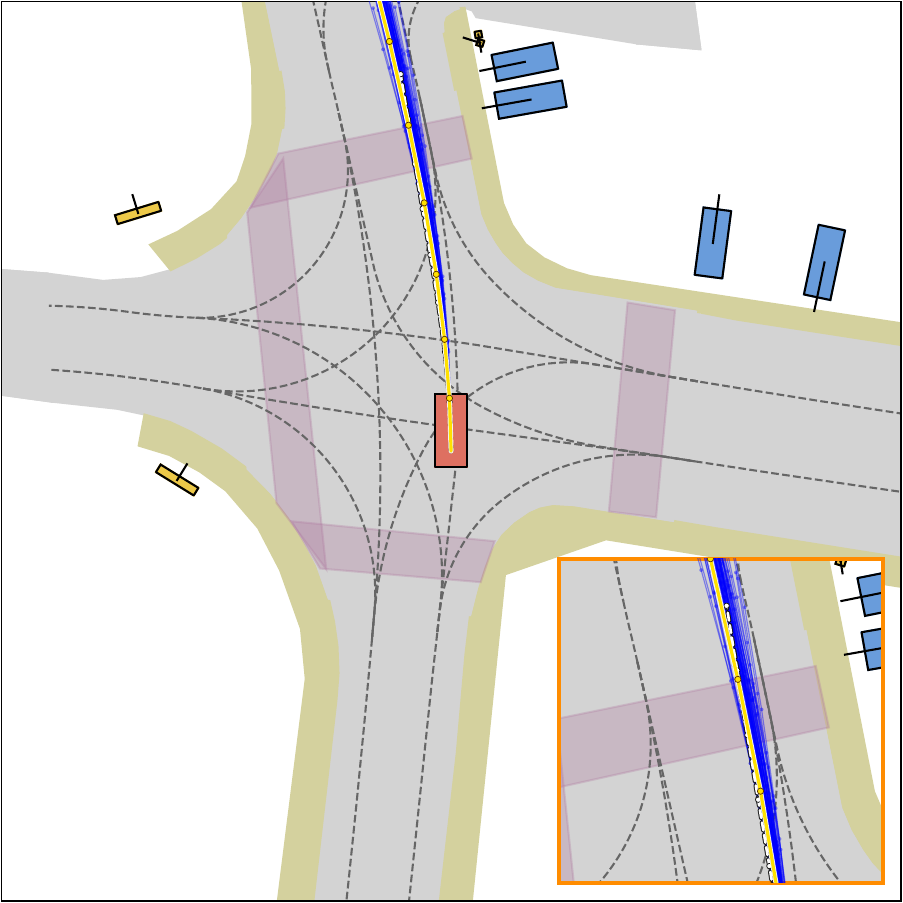}
\end{minipage}\hfill
\begin{minipage}[t]{.179\textwidth}\centering
{\fontsize{8}{9}\selectfont\bfseries Scene 29}\par\vspace{.5pt}
\includegraphics[width=\linewidth]{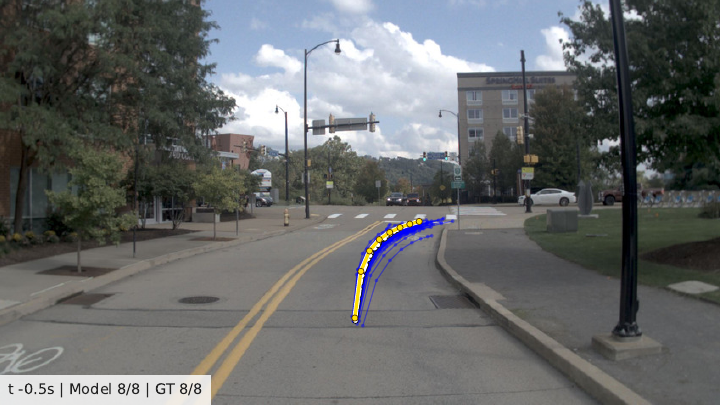}\par\vspace{.5pt}
\includegraphics[width=\linewidth]{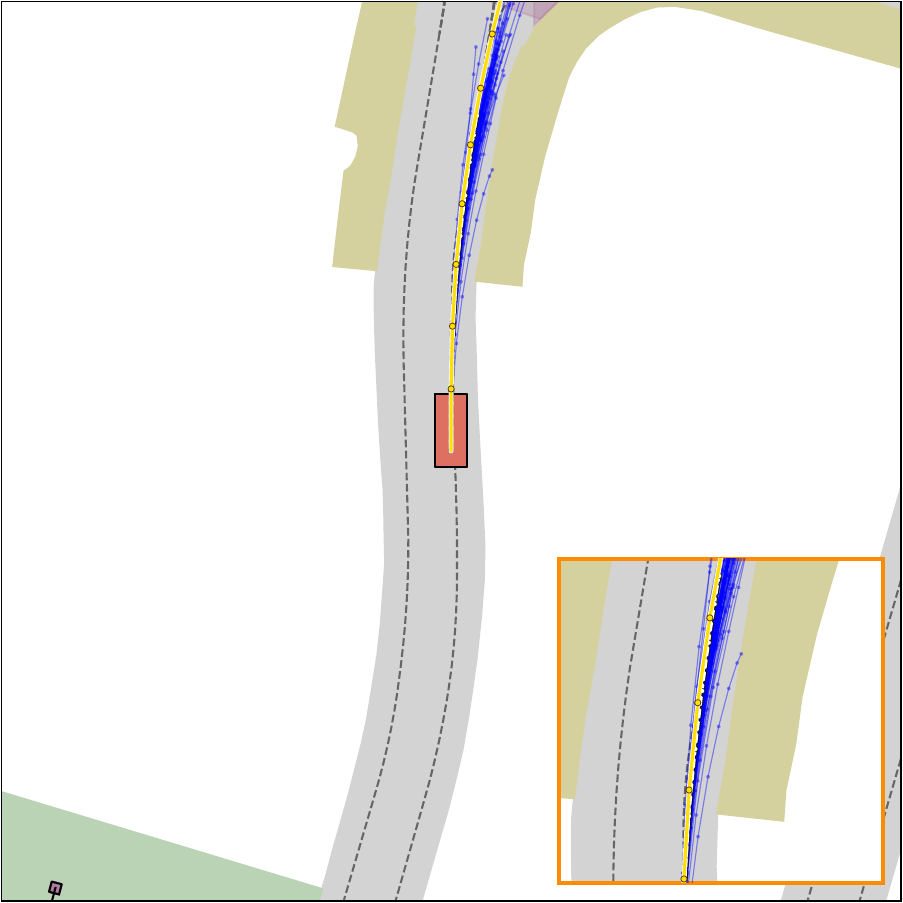}
\end{minipage}\hfill
\begin{minipage}[t]{.179\textwidth}\centering
{\fontsize{8}{9}\selectfont\bfseries Scene 30}\par\vspace{.5pt}
\includegraphics[width=\linewidth]{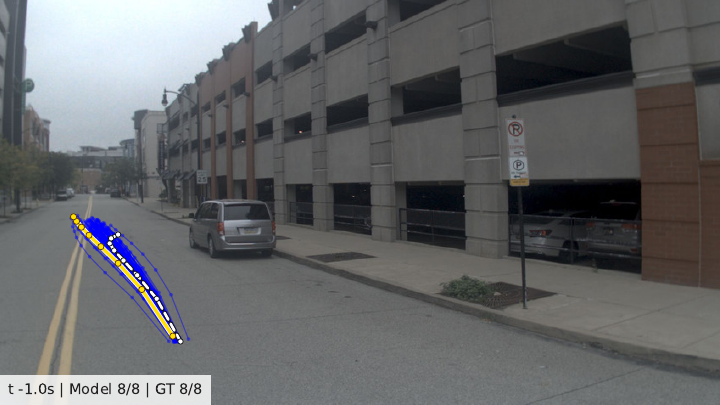}\par\vspace{.5pt}
\includegraphics[width=\linewidth]{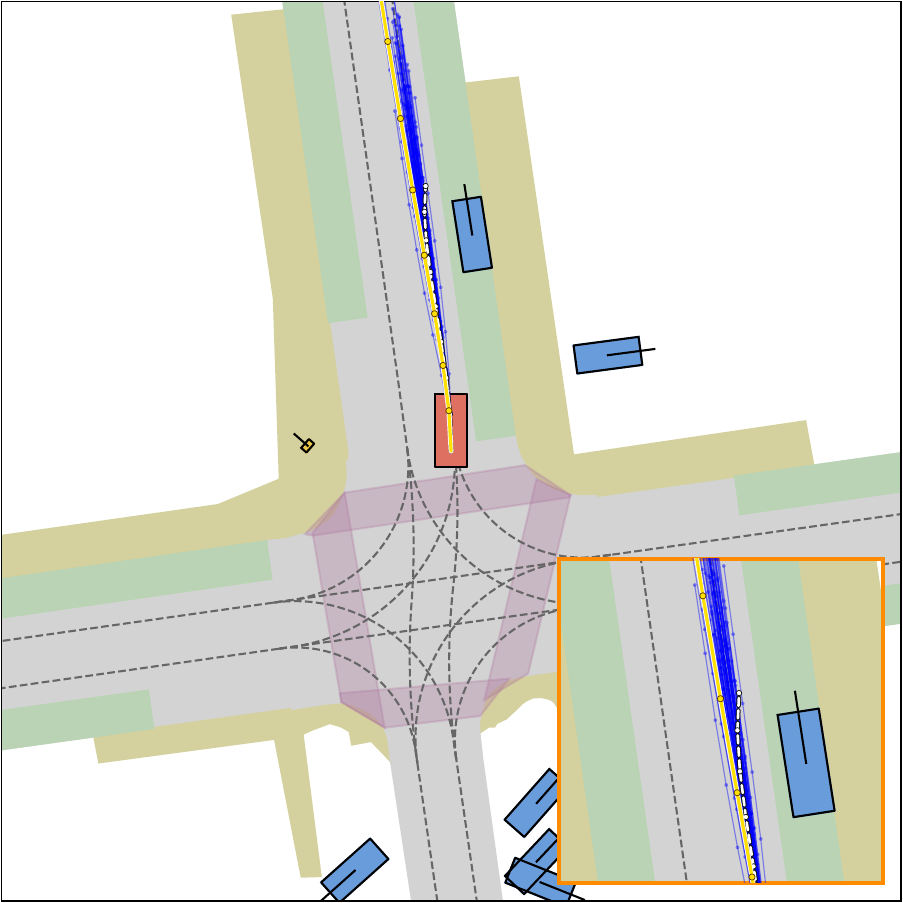}
\end{minipage}\par\vspace{1pt}
\begin{minipage}[t]{.179\textwidth}\centering
{\fontsize{8}{9}\selectfont\bfseries Scene 31}\par\vspace{.5pt}
\includegraphics[width=\linewidth]{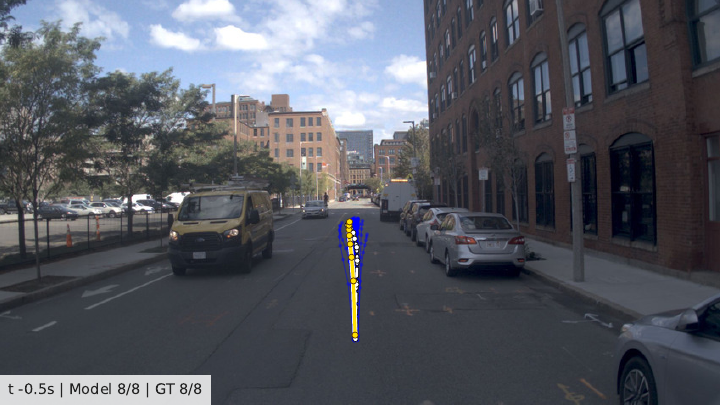}\par\vspace{.5pt}
\includegraphics[width=\linewidth]{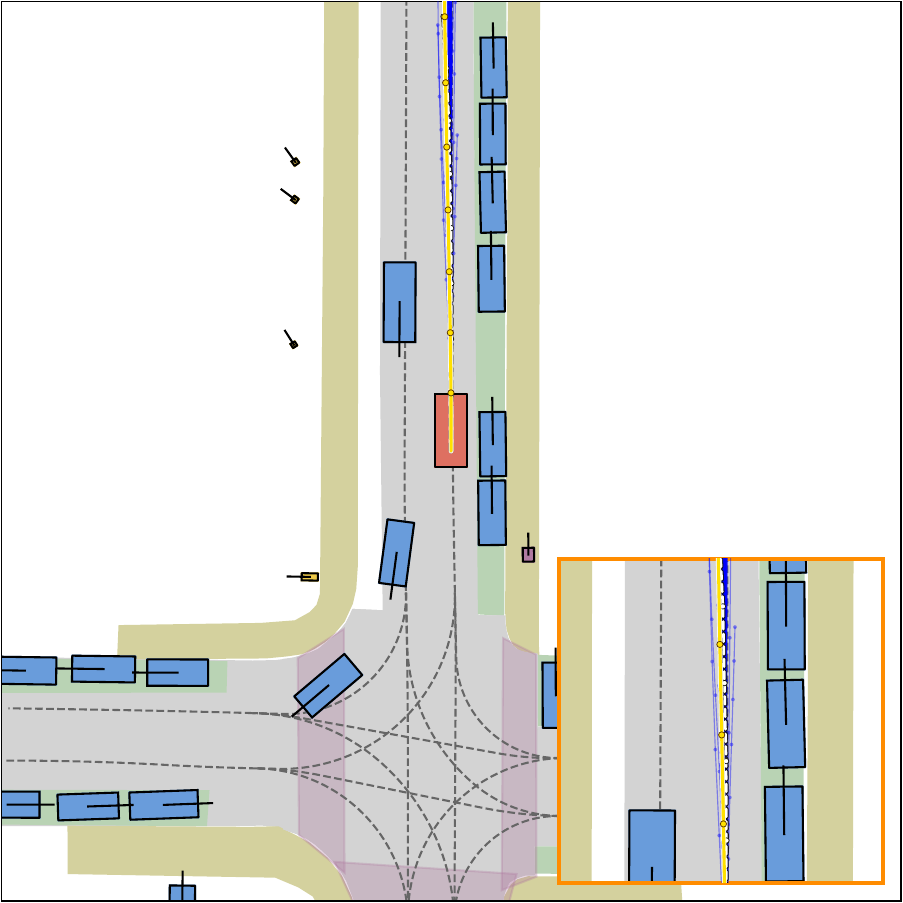}
\end{minipage}\hfill
\begin{minipage}[t]{.179\textwidth}\centering
{\fontsize{8}{9}\selectfont\bfseries Scene 32}\par\vspace{.5pt}
\includegraphics[width=\linewidth]{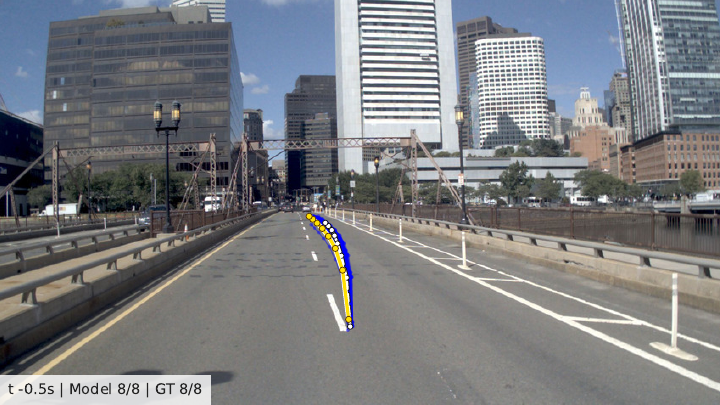}\par\vspace{.5pt}
\includegraphics[width=\linewidth]{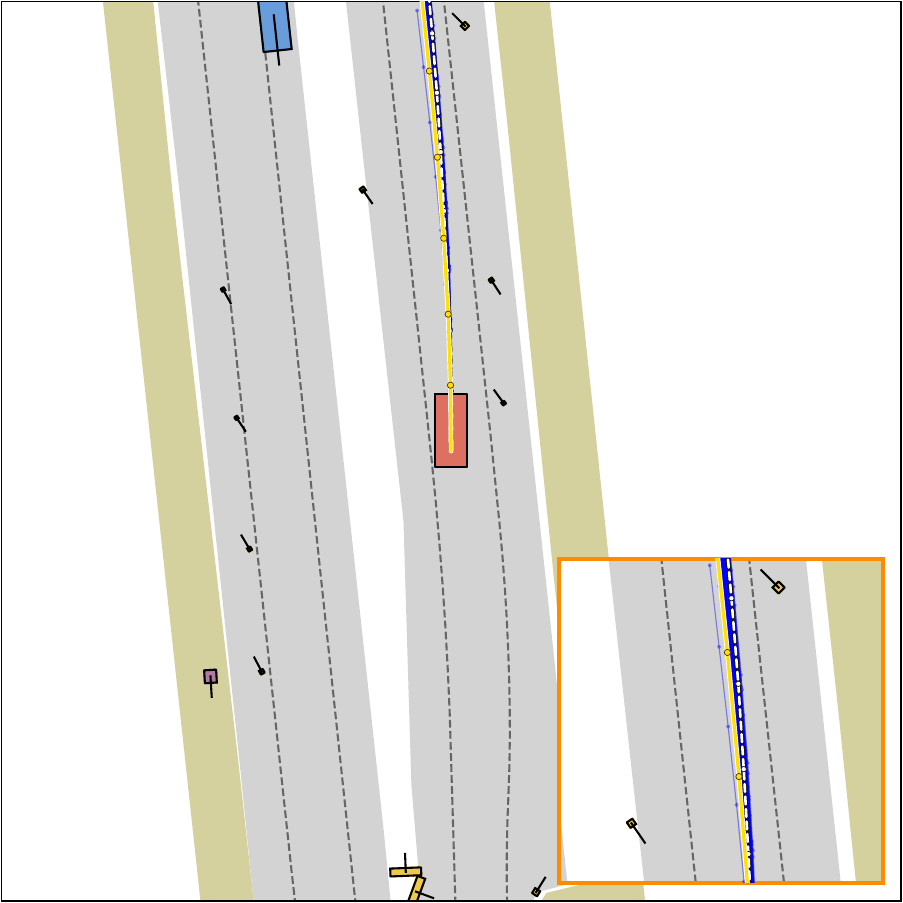}
\end{minipage}\hfill
\begin{minipage}[t]{.179\textwidth}\centering
{\fontsize{8}{9}\selectfont\bfseries Scene 33}\par\vspace{.5pt}
\includegraphics[width=\linewidth]{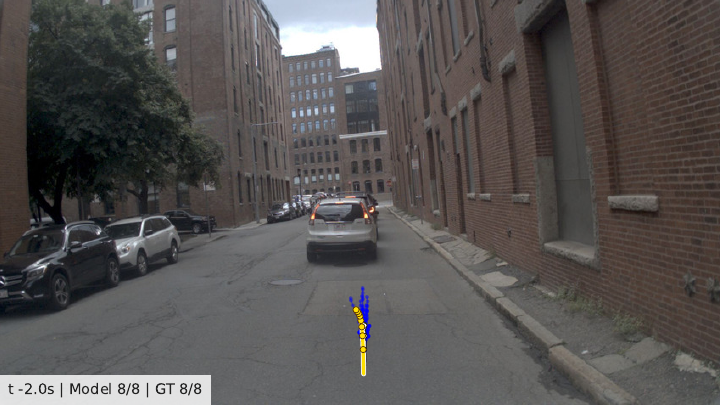}\par\vspace{.5pt}
\includegraphics[width=\linewidth]{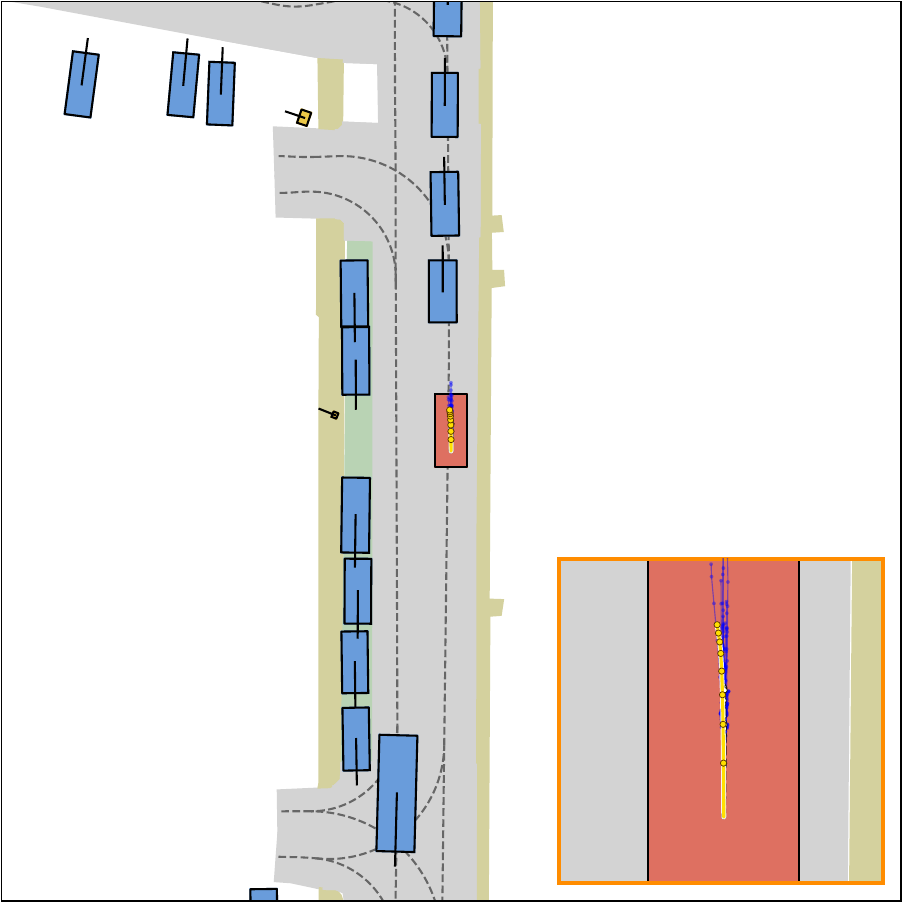}
\end{minipage}\hfill
\begin{minipage}[t]{.179\textwidth}\centering
{\fontsize{8}{9}\selectfont\bfseries Scene 34}\par\vspace{.5pt}
\includegraphics[width=\linewidth]{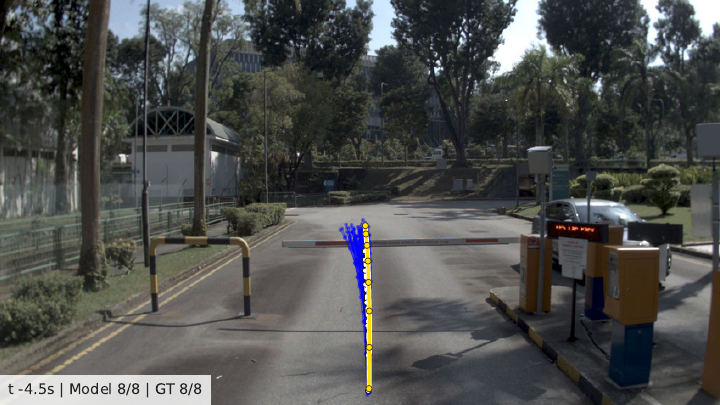}\par\vspace{.5pt}
\includegraphics[width=\linewidth]{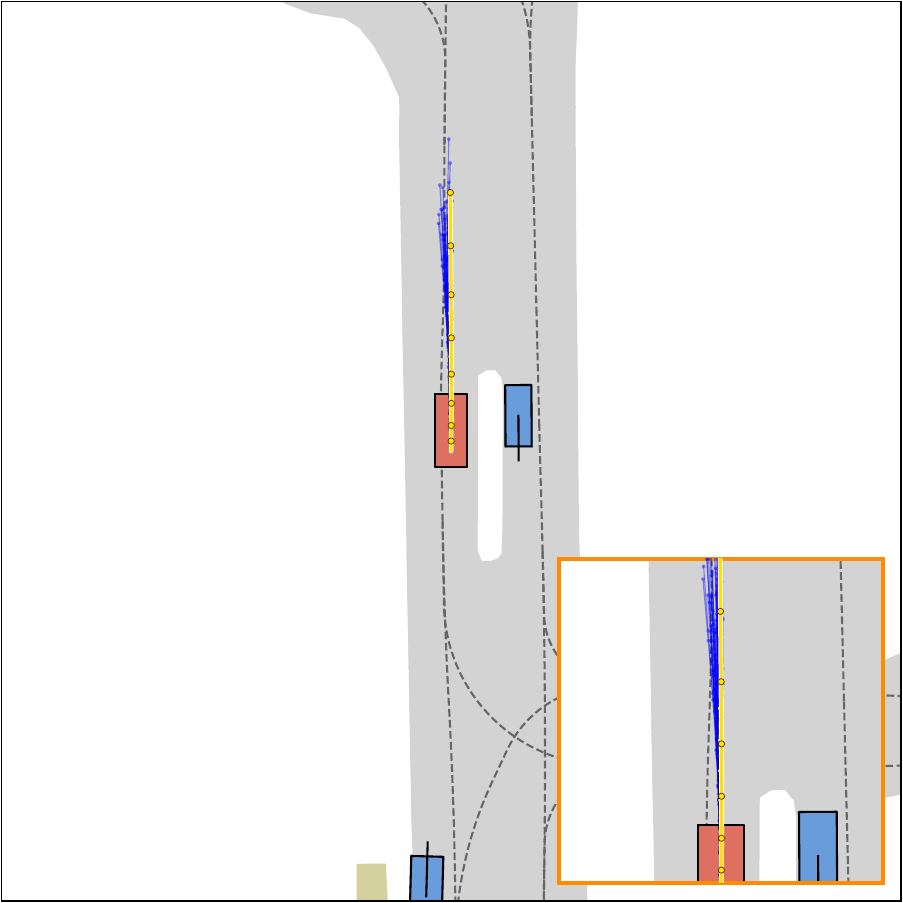}
\end{minipage}\hfill
\begin{minipage}[t]{.179\textwidth}\centering
{\fontsize{8}{9}\selectfont\bfseries Scene 35}\par\vspace{.5pt}
\includegraphics[width=\linewidth]{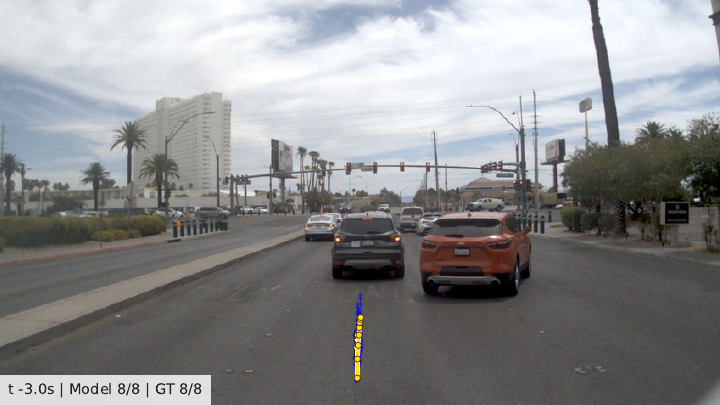}\par\vspace{.5pt}
\includegraphics[width=\linewidth]{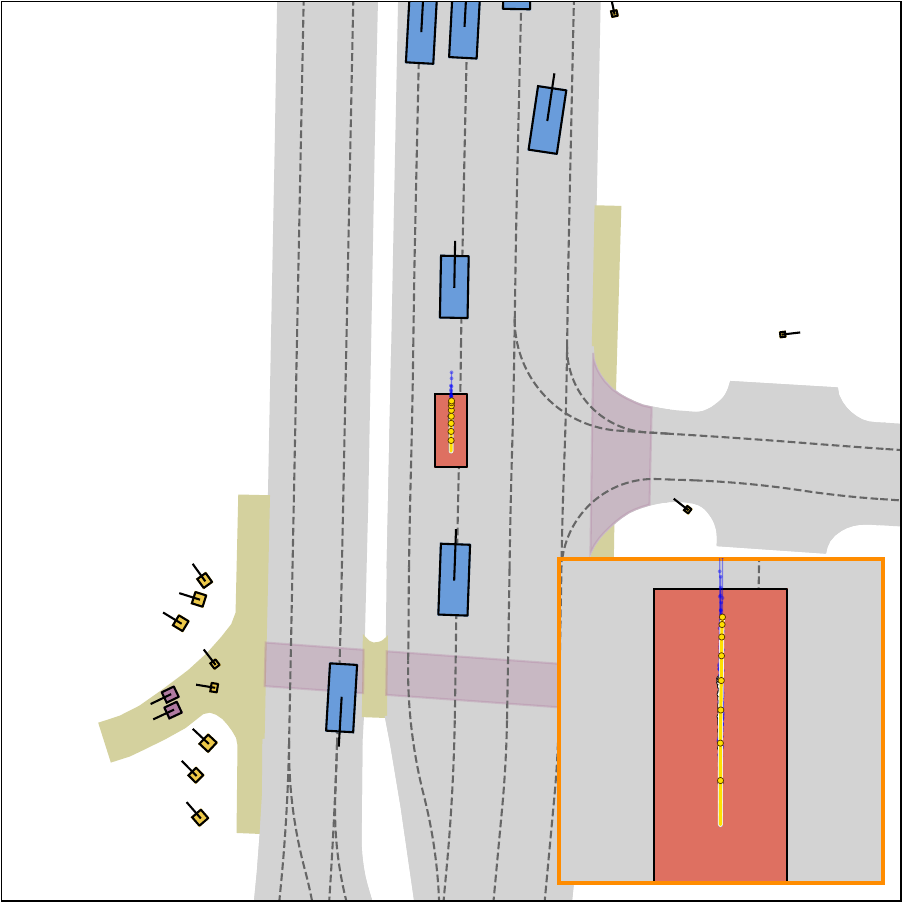}
\end{minipage}\par\vspace{1pt}
\begin{minipage}[t]{.179\textwidth}\centering
{\fontsize{8}{9}\selectfont\bfseries Scene 36}\par\vspace{.5pt}
\includegraphics[width=\linewidth]{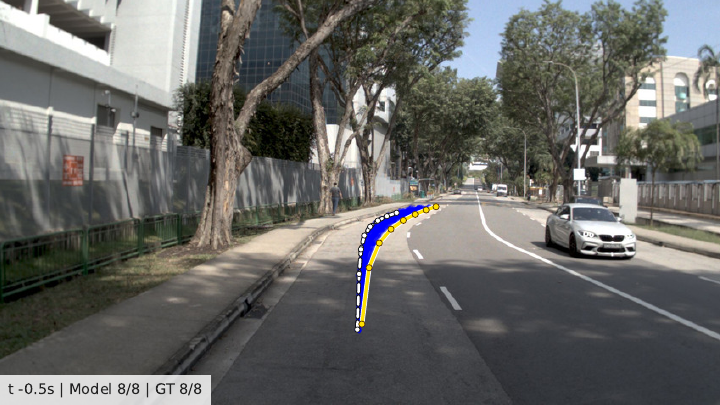}\par\vspace{.5pt}
\includegraphics[width=\linewidth]{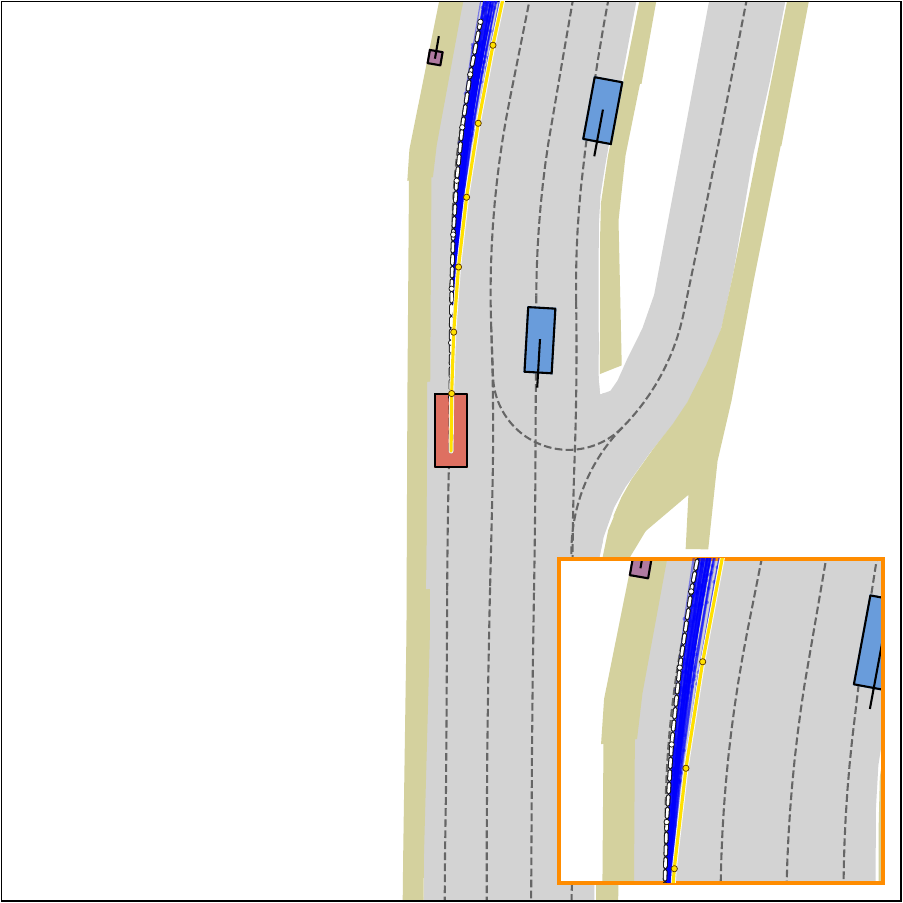}
\end{minipage}\hfill
\begin{minipage}[t]{.179\textwidth}\centering
{\fontsize{8}{9}\selectfont\bfseries Scene 37}\par\vspace{.5pt}
\includegraphics[width=\linewidth]{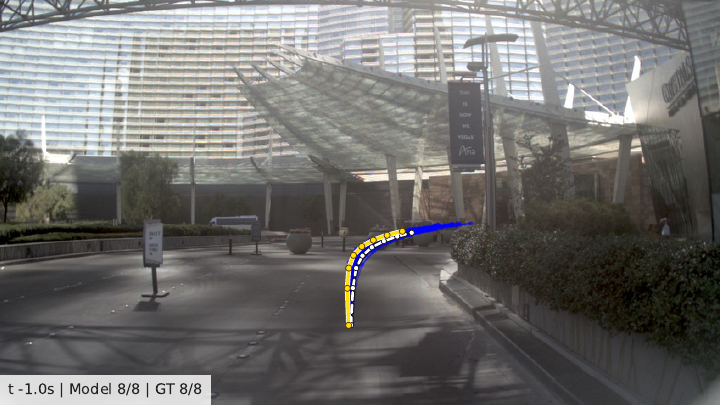}\par\vspace{.5pt}
\includegraphics[width=\linewidth]{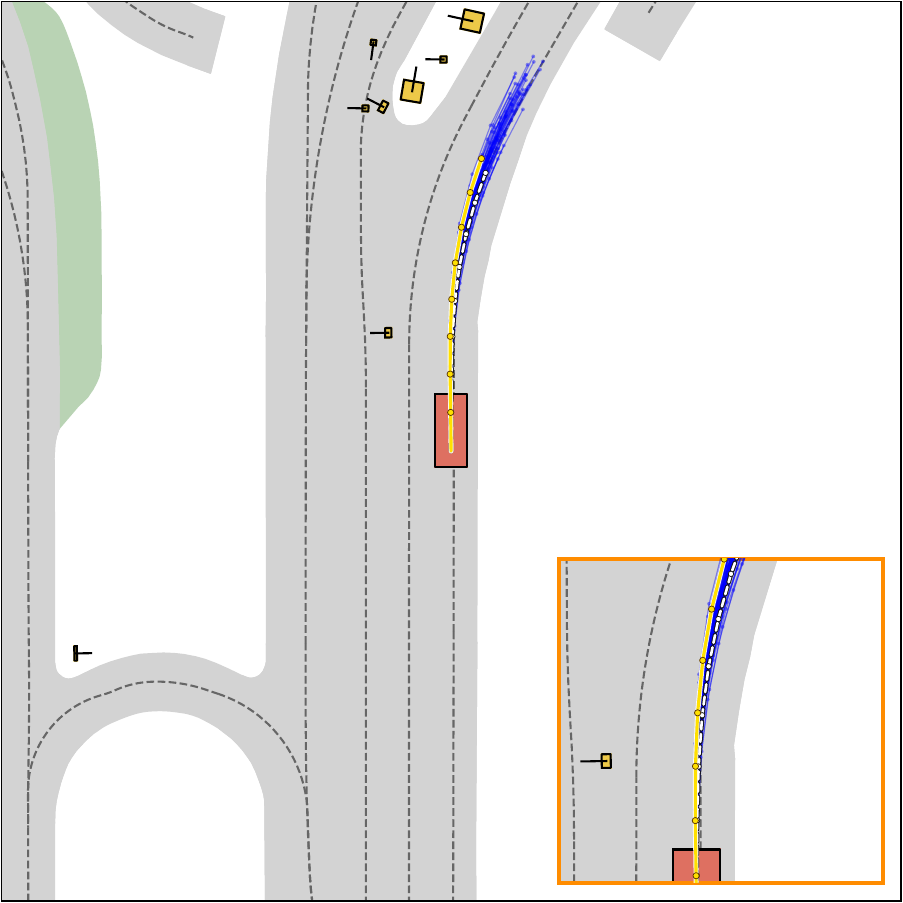}
\end{minipage}\hfill
\begin{minipage}[t]{.179\textwidth}\centering
{\fontsize{8}{9}\selectfont\bfseries Scene 38}\par\vspace{.5pt}
\includegraphics[width=\linewidth]{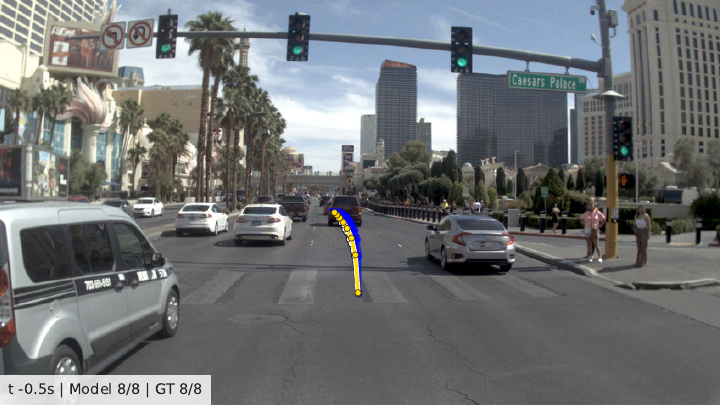}\par\vspace{.5pt}
\includegraphics[width=\linewidth]{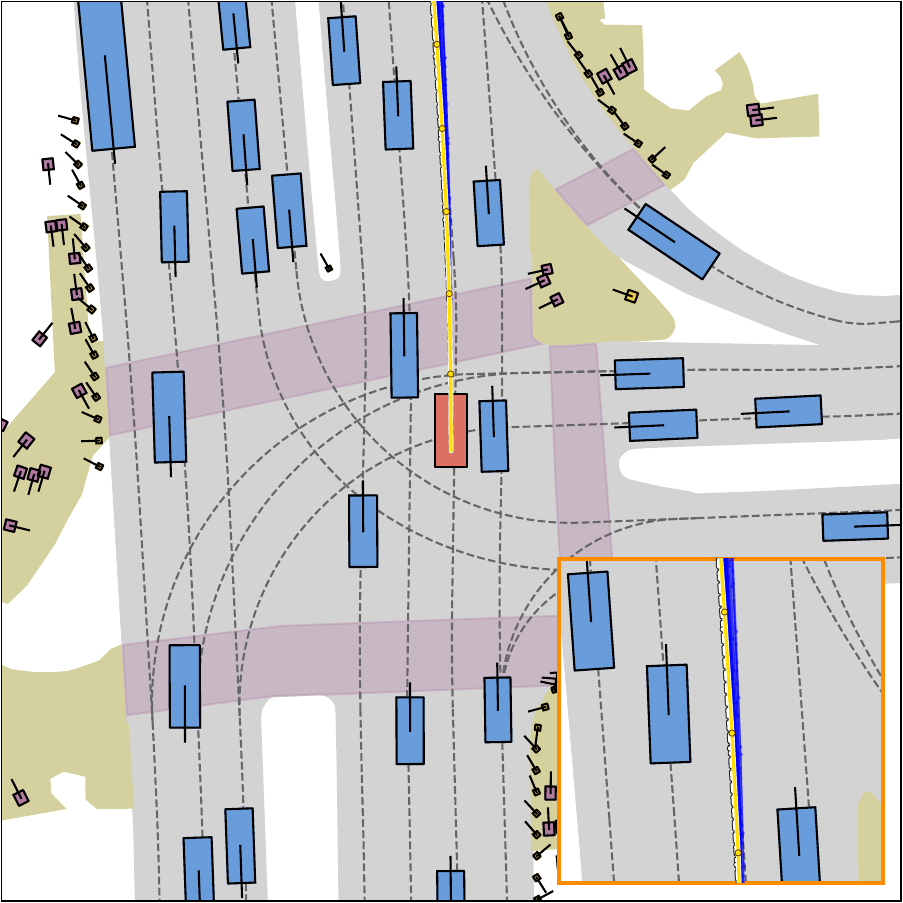}
\end{minipage}\hfill
\begin{minipage}[t]{.179\textwidth}\centering
{\fontsize{8}{9}\selectfont\bfseries Scene 39}\par\vspace{.5pt}
\includegraphics[width=\linewidth]{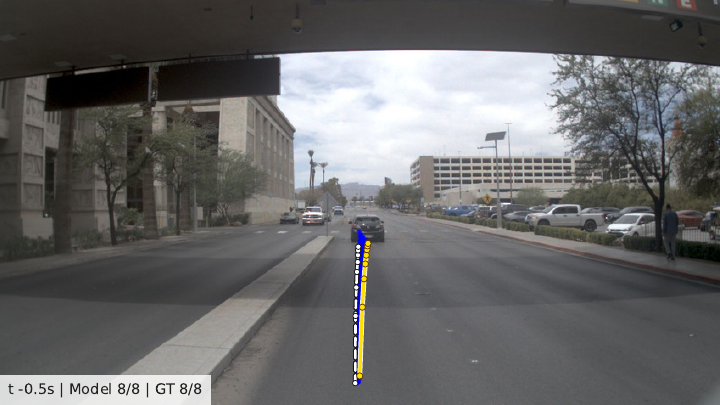}\par\vspace{.5pt}
\includegraphics[width=\linewidth]{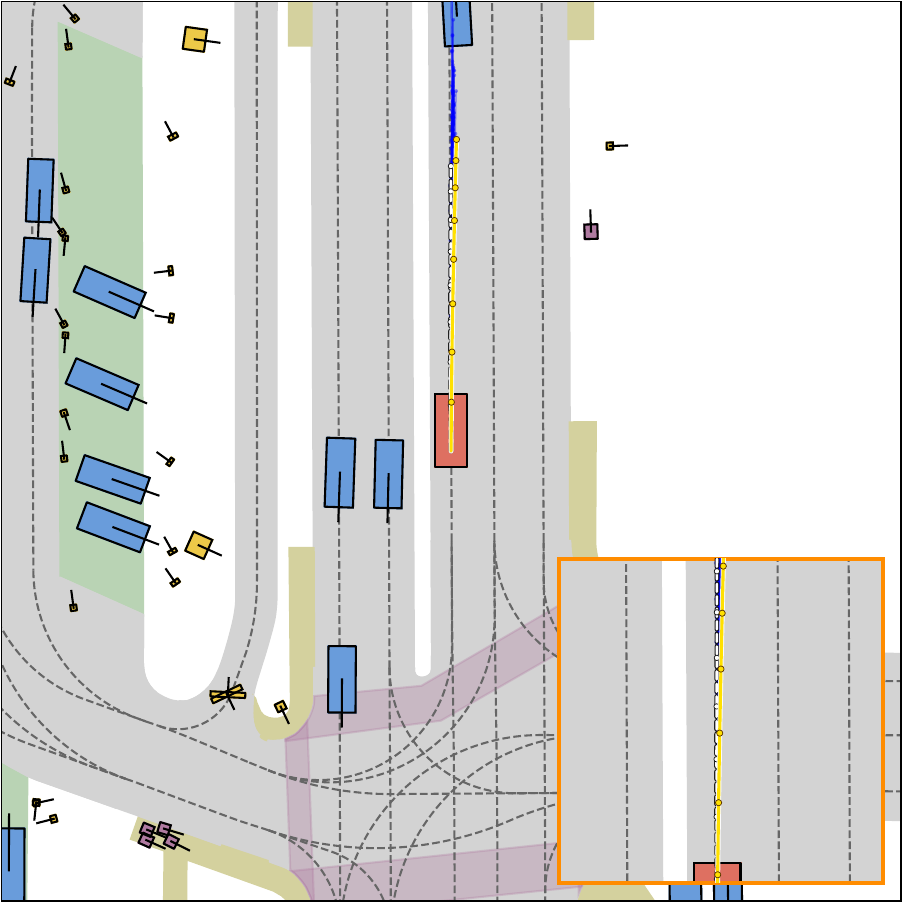}
\end{minipage}\hfill
\begin{minipage}[t]{.179\textwidth}\centering
{\fontsize{8}{9}\selectfont\bfseries Scene 40}\par\vspace{.5pt}
\includegraphics[width=\linewidth]{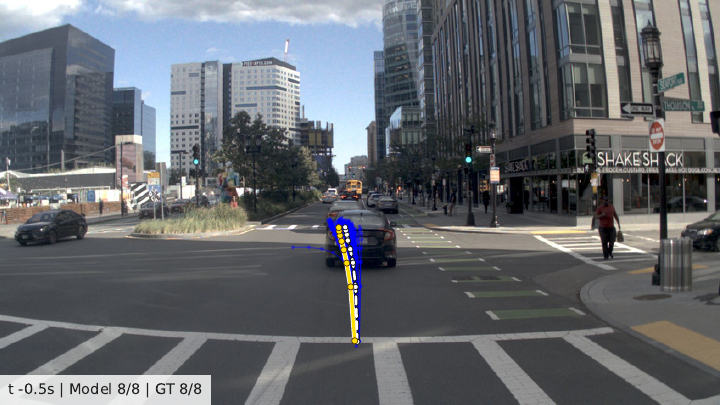}\par\vspace{.5pt}
\includegraphics[width=\linewidth]{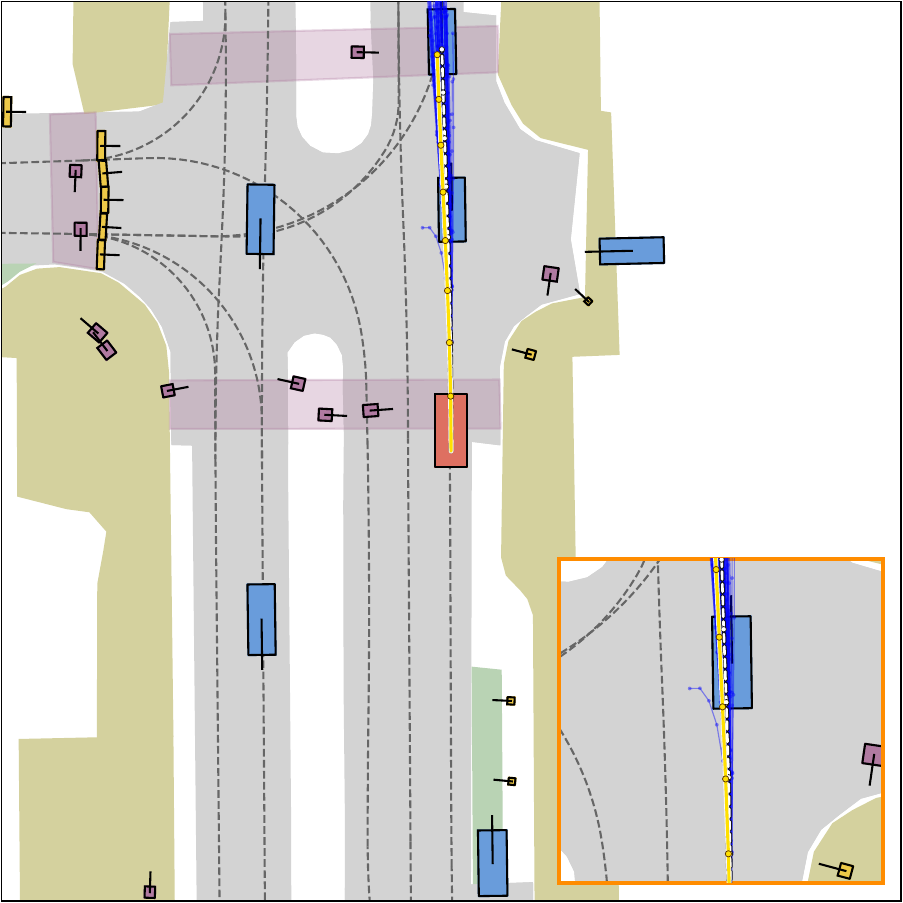}
\end{minipage}\par\vspace{.5pt}
\caption{\textbf{DriveTTO outputs, scenes 21--40.} Colors follow Figure~\ref{fig:qual_scenes}. Paired views cover streets with traffic, intersections, a bridge, a barrier-gate approach, and curved entrances. In the insets, the paths are nearly parallel on the straight road in scene 25 and diverge around the bend in scene 27. All 20 selected outputs score 100 PDMS.}
\label{fig:qual_scenes_more}
\end{figure*}

\begin{figure*}[p]
\centering
\makebox[\textwidth][l]{\footnotesize\bfseries Scene 1}\par\vspace{1pt}
\begin{minipage}[t]{.188\textwidth}\vspace{0pt}\includegraphics[width=\linewidth]{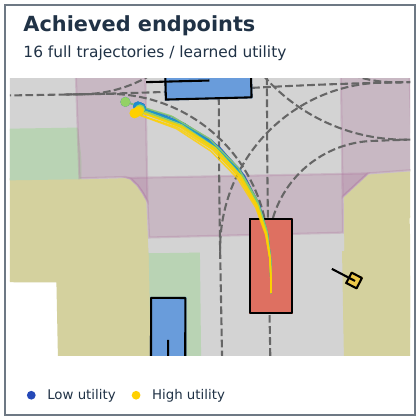}\end{minipage}\hfill
\begin{minipage}[t]{.188\textwidth}\vspace{0pt}\includegraphics[width=\linewidth]{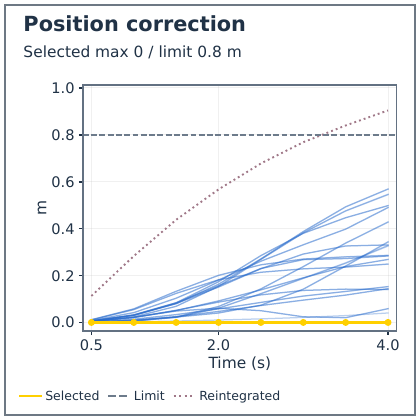}\end{minipage}\hfill
\begin{minipage}[t]{.188\textwidth}\vspace{0pt}\includegraphics[width=\linewidth]{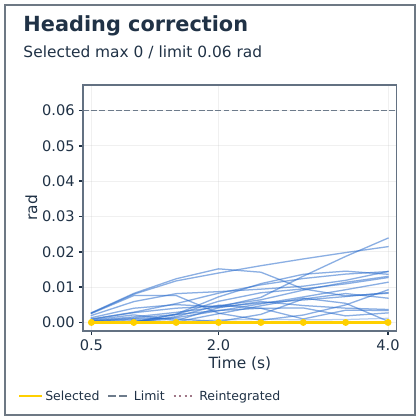}\end{minipage}\hfill
\begin{minipage}[t]{.188\textwidth}\vspace{0pt}\includegraphics[width=\linewidth]{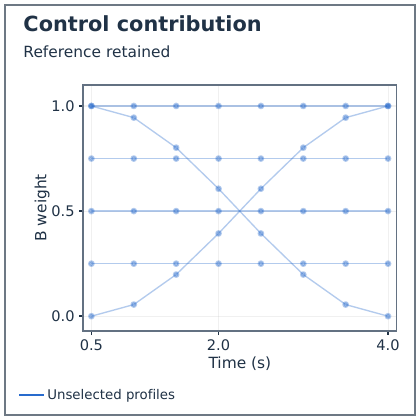}\end{minipage}\hfill
\begin{minipage}[t]{.188\textwidth}\vspace{0pt}\includegraphics[width=\linewidth]{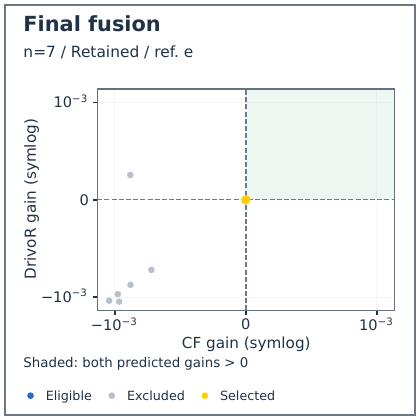}\end{minipage}\par\vspace{1pt}
\makebox[\textwidth][l]{\footnotesize\bfseries Scene 6}\par\vspace{1pt}
\begin{minipage}[t]{.188\textwidth}\vspace{0pt}\includegraphics[width=\linewidth]{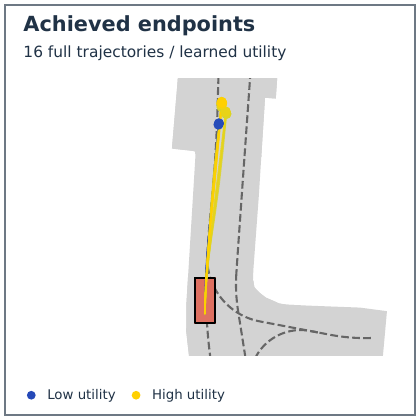}\end{minipage}\hfill
\begin{minipage}[t]{.188\textwidth}\vspace{0pt}\includegraphics[width=\linewidth]{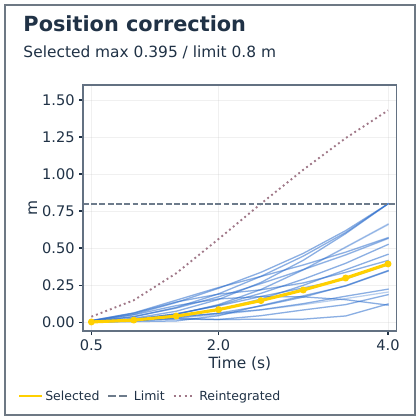}\end{minipage}\hfill
\begin{minipage}[t]{.188\textwidth}\vspace{0pt}\includegraphics[width=\linewidth]{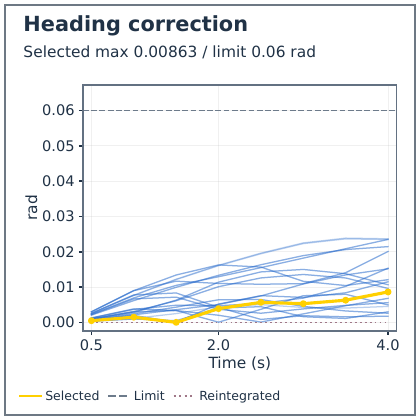}\end{minipage}\hfill
\begin{minipage}[t]{.188\textwidth}\vspace{0pt}\includegraphics[width=\linewidth]{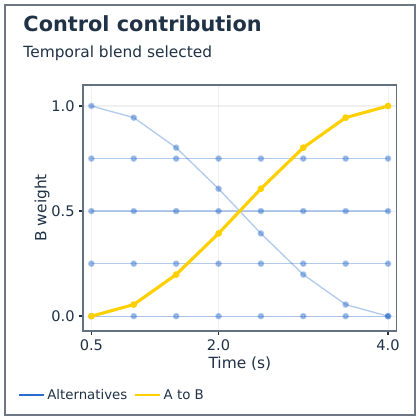}\end{minipage}\hfill
\begin{minipage}[t]{.188\textwidth}\vspace{0pt}\includegraphics[width=\linewidth]{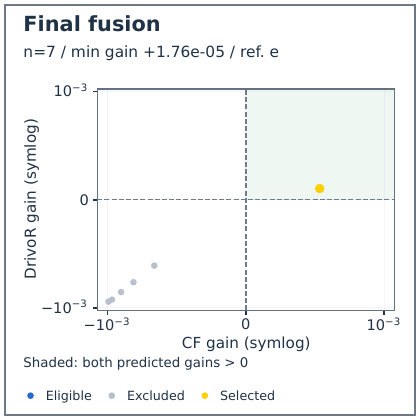}\end{minipage}\par\vspace{1pt}
\makebox[\textwidth][l]{\footnotesize\bfseries Scene 4}\par\vspace{1pt}
\begin{minipage}[t]{.188\textwidth}\vspace{0pt}\includegraphics[width=\linewidth]{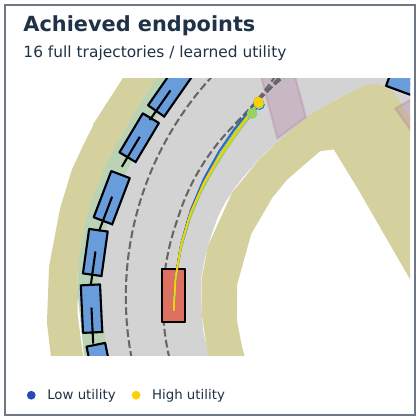}\end{minipage}\hfill
\begin{minipage}[t]{.188\textwidth}\vspace{0pt}\includegraphics[width=\linewidth]{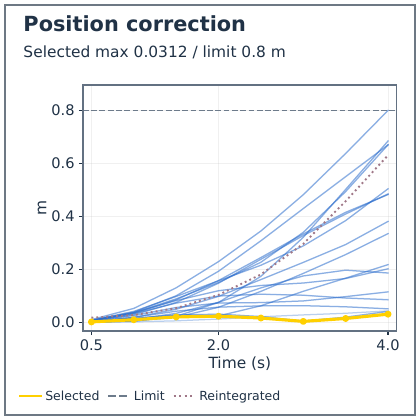}\end{minipage}\hfill
\begin{minipage}[t]{.188\textwidth}\vspace{0pt}\includegraphics[width=\linewidth]{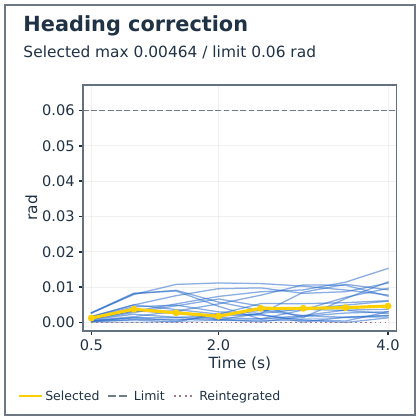}\end{minipage}\hfill
\begin{minipage}[t]{.188\textwidth}\vspace{0pt}\includegraphics[width=\linewidth]{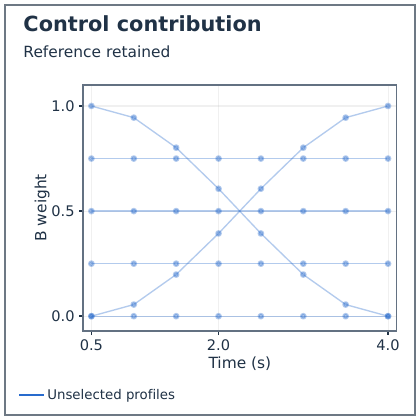}\end{minipage}\hfill
\begin{minipage}[t]{.188\textwidth}\vspace{0pt}\includegraphics[width=\linewidth]{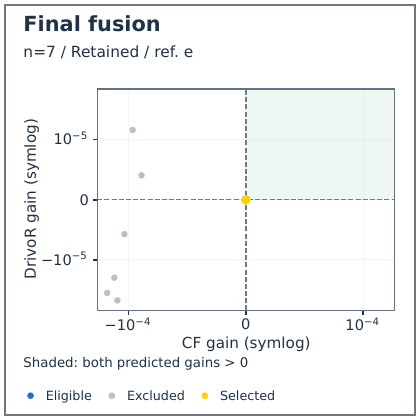}\end{minipage}\par\vspace{1pt}
\makebox[\textwidth][l]{\footnotesize\bfseries Scene 24}\par\vspace{1pt}
\begin{minipage}[t]{.188\textwidth}\vspace{0pt}\includegraphics[width=\linewidth]{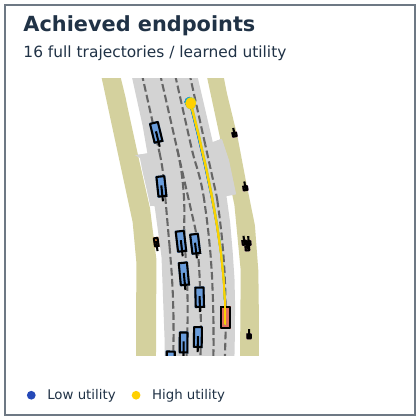}\end{minipage}\hfill
\begin{minipage}[t]{.188\textwidth}\vspace{0pt}\includegraphics[width=\linewidth]{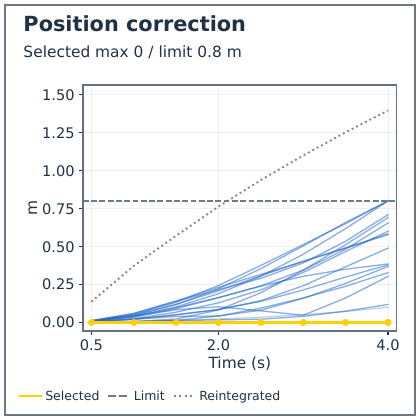}\end{minipage}\hfill
\begin{minipage}[t]{.188\textwidth}\vspace{0pt}\includegraphics[width=\linewidth]{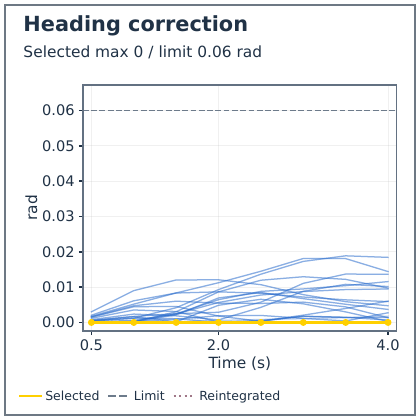}\end{minipage}\hfill
\begin{minipage}[t]{.188\textwidth}\vspace{0pt}\includegraphics[width=\linewidth]{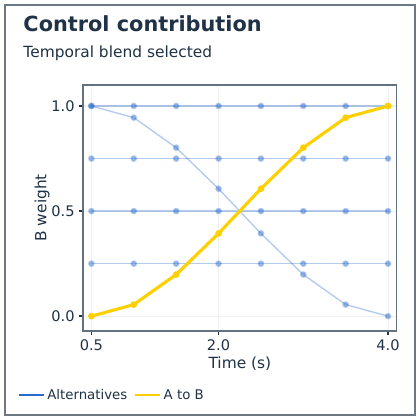}\end{minipage}\hfill
\begin{minipage}[t]{.188\textwidth}\vspace{0pt}\includegraphics[width=\linewidth]{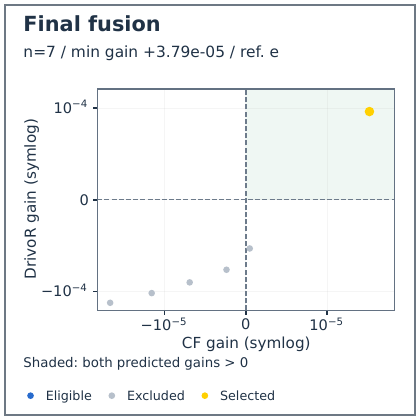}\end{minipage}\par\vspace{1pt}
\makebox[\textwidth][l]{\footnotesize\bfseries Scene 15}\par\vspace{1pt}
\begin{minipage}[t]{.188\textwidth}\vspace{0pt}\includegraphics[width=\linewidth]{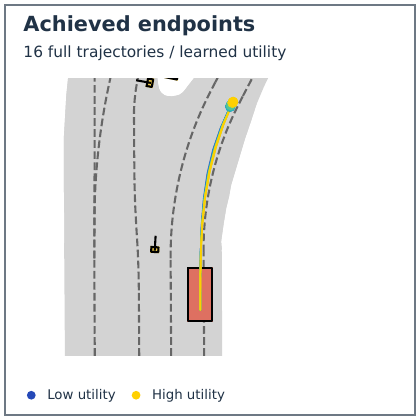}\end{minipage}\hfill
\begin{minipage}[t]{.188\textwidth}\vspace{0pt}\includegraphics[width=\linewidth]{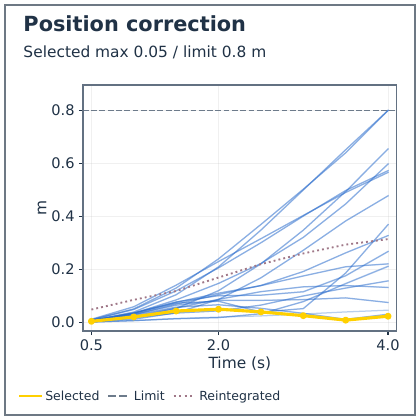}\end{minipage}\hfill
\begin{minipage}[t]{.188\textwidth}\vspace{0pt}\includegraphics[width=\linewidth]{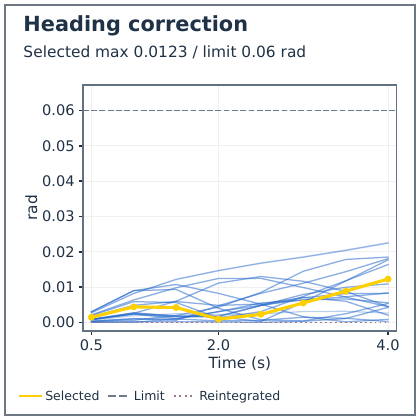}\end{minipage}\hfill
\begin{minipage}[t]{.188\textwidth}\vspace{0pt}\includegraphics[width=\linewidth]{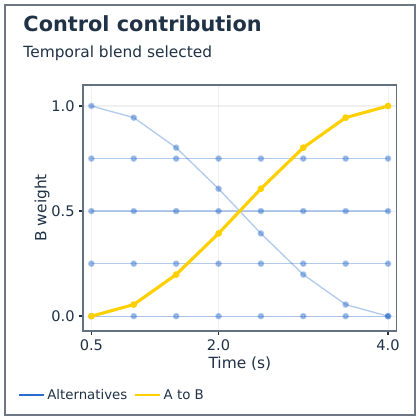}\end{minipage}\hfill
\begin{minipage}[t]{.188\textwidth}\vspace{0pt}\includegraphics[width=\linewidth]{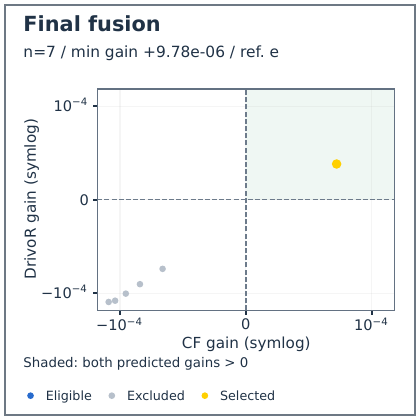}\end{minipage}
\caption{\textbf{Endpoint completions and bounded updates.} Columns show achieved endpoints, position corrections, heading corrections, fusion weights, and final selection. Endpoint colors encode predicted full-path utility. Yellow curves show returned corrections, dashed lines mark the 0.8\,m and 0.06\,rad caps, and purple dotted curves show nominal reintegration. Scenes 1 and 24 retain the control reference. Scenes 6, 4, and 15 use strength 0.75. In scene 6 the selected position and heading corrections peak at 0.395\,m and 0.00863\,rad. Fusion retains its reference in scenes 1 and 4 and selects A-to-B blends elsewhere. Final axes show predicted ChainFlow/DrivoR gains against the fusion reference.}
\label{fig:qual_stages}
\end{figure*}

\begin{figure*}[p]
\centering
\includegraphics[width=\textwidth]{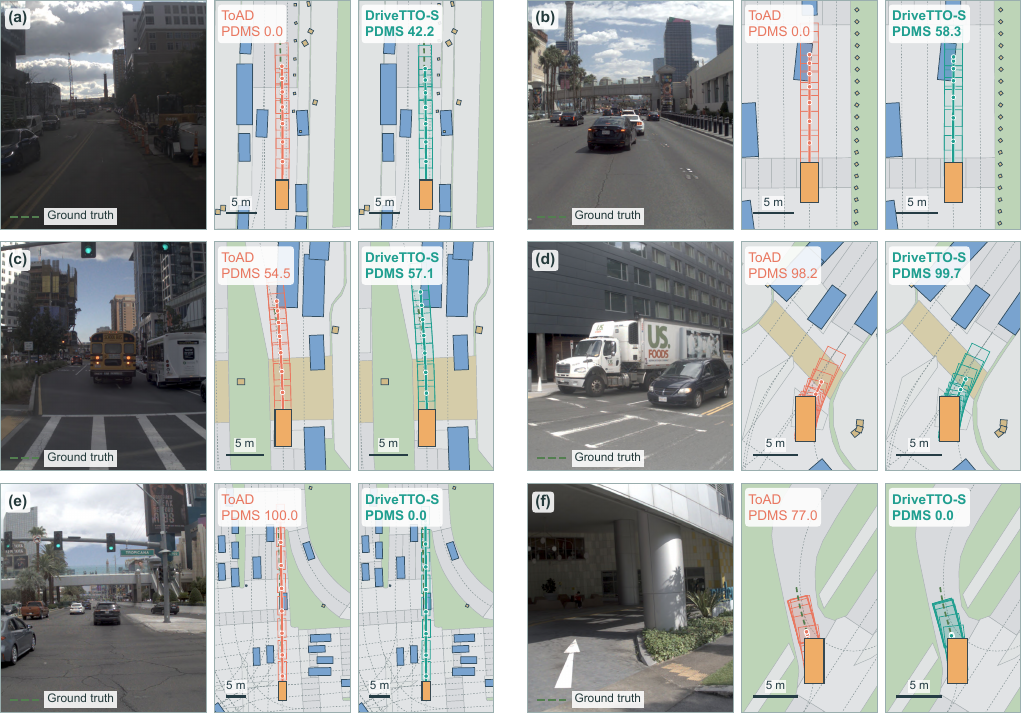}
\caption{\textbf{Matched ToAD and DriveTTO-S predictions on Navtest.} Examples use DrivoR predictions from Table~\ref{tab:transfer}. Each pair shares the map and scale. Blue boxes show observed vehicles, orange marks the ego vehicle, and thin outlines show planned footprints at 0.5\,s intervals. In (a,b), ToAD scores zero while DriveTTO-S scores above zero. Panels (c,d) improve on nonzero ToAD scores, and (e,f) show regressions relative to ToAD. These selected examples illustrate both outcomes of refinement.}
\label{fig:navtest_toad_tto_cases}
\end{figure*}
\clearpage
\endgroup
\end{document}